\documentclass[9pt]{article}
\usepackage[%
  top=2cm,
  bottom=2cm,
  left=2.5cm,
  right=2.5cm
]{geometry}

\usepackage{multirow}
\usepackage{xr}
\usepackage{xcolor}
\usepackage{bm}
\usepackage[numbers]{natbib}
\usepackage{graphicx, amsmath, amssymb}
\usepackage{float}

\makeatletter

\newcommand*{\addFileDependency}[1]{
\typeout{(#1)}
\@addtofilelist{#1}
\IfFileExists{#1}{}{\typeout{No file #1.}}
}\makeatother

\newcommand*{\myexternaldocument}[1]{%
\externaldocument{#1}%
\addFileDependency{#1.tex}%
\addFileDependency{#1.aux}%
}
\myexternaldocument{SI}

\title{Order parameters and phase transitions of continual learning in deep
neural networks}

\author{
  Haozhe Shan\textsuperscript{1,2,\dag}
  \quad
  Qianyi Li\textsuperscript{1,3,\dag}
  \quad
  Haim Sompolinsky\textsuperscript{1,4,\ddag}
}

\begin{document}

\maketitle

\begin{center}

\textsuperscript{1}Center for Brain Science, Harvard University,
Cambridge, Massachusetts, 02138, United States \\
\textsuperscript{2}Program in Neuroscience, Harvard Medical School,
Boston, Massachusetts, 02115, United States \\
\textsuperscript{3}Biophysics Graduate Program, Harvard University,
Cambridge, Massachusetts, 02138, United States \\
\textsuperscript{4}Edmond and Lily Safra Center for Brain Sciences,
Hebrew University, Jerusalem, 9190401, Israel \\
\vspace{1em}
\textsuperscript{\dag}These authors contributed equally to this work \\
\textsuperscript{\ddag}Corresponding author
\end{center}

\begin{abstract}
Continual learning (CL) enables animals to learn new tasks without
erasing prior knowledge. CL in artificial neural networks (NNs) is
challenging due to catastrophic forgetting, where new learning degrades
performance on older tasks. While various techniques exist to mitigate
forgetting, theoretical insights into when and why CL fails in NNs
are lacking. Here, we present a statistical-mechanics theory of CL
in deep, wide NNs, which characterizes the network's input-output
mapping as it learns a sequence of tasks. It gives rise to order parameters
(OPs) that capture how task relations and network architecture influence
forgetting and anterograde interference, as verified by numerical evaluations.
For networks with a shared readout for all tasks (single-head CL), the relevant-feature and rule similarity between tasks, respectively measured by two OPs, are sufficient to predict a wide range of CL behaviors. In addition, the theory predicts that increasing the network depth can effectively reduce interference between tasks, thereby lowering forgetting. 
For networks with task-specific readouts (multi-head CL), the theory identifies a phase transition where CL performance shifts dramatically as tasks become less similar, as measured by another task-similarity OP. While forgetting is relatively mild compared to single-head CL across all tasks, sufficiently low similarity leads to catastrophic anterograde interference, where the network retains old tasks perfectly but completely fails to generalize new learning. Our results delineate important factors affecting CL performance and suggest strategies for mitigating forgetting.
\end{abstract}

\section{Introduction}

Continual learning (CL), the capability to acquire and refine knowledge
and skills over time, is fundamental to how animals survive in a non-stationary world. As an animal learns and performs many tasks, CL allows it to leverage previous learning to help learn a new task while retaining the ability to perform old ones. In artificial neural networks
(NN), developing such abilities has been challenging. Artificial NNs especially struggle with catastrophic forgetting, where learning a new task overwrites
existing information and dramatically degrades performance on previously learned tasks \citep{mccloskey1989catastrophic,goodfellow2013empirical,parisi2019continual,van2019three}.
This problem is so prevalent and severe in machine learning (ML) that
it has become one of the biggest challenges for developing human-level
artificial general intelligence \citep{silver2011machine,kirkpatrick2017overcoming,hadsell2020embracing}.
Despite also relying on NNs for computation, the brain clearly does
not suffer from catastrophic forgetting to nearly the same extent \cite{hadsell2020embracing}.
Not only does this offer an ``existence proof'' of successful CL
in NNs \citep{parisi2019continual,hadsell2020embracing,hassabis2017neuroscience},
it also raises intriguing questions about mechanisms underlying CL
in the brain. A wide range of possible underpinnings, ranging from
memory reactivation \citep{rasch2007maintaining}, synaptic stabilization
\citep{yang2009stably,xu2009rapid}, to representational drift \citep{driscoll2022representational},
have been proposed. However, their specific contributions to CL in
the brain are not well understood.

Engineering CL in ML and understanding its mechanisms in the brain both suffer from a lack of theoretical understanding of the problem in NNs. The challenges stem from two interconnected issues: (1) developing an analytical understanding of the learning process of CL in NNs and (2) characterizing and quantifying the diverse types of task relations and their impacts on forgetting. Recently, progress on the first challenge has been made by analyzing simplified cases where the network is a shallow linear network \citep{asanuma2021statistical,dhifallah2021phase,evron2022catastrophic,saglietti2022analytical,li2023statistical}, or equivalently so via the neural tangent kernel (NTK) approximation \citep{doan2021theoretical,jacot2018neural}. However, these approaches fail to account for the realistic and common scenario of NNs having multiple readouts dedicated to different tasks. Other studies have addressed task-dedicated readouts but focused on networks with only a single hidden layer and a limited number of neurons \citep{lee2021continual,gerace2022probing,lee2022maslow}, limiting their relevance to real-world NNs with typically hundreds of hidden units in each layer. Moreover, many of these studies \citep{asanuma2021statistical,dhifallah2021phase,saglietti2022analytical,li2023statistical,lee2022maslow,lee2021continual,gerace2022probing,straat2022supervised} rely on specific assumptions about the task data, limiting the generalizability of their conclusions across diverse datasets and task relations. 

The second challenge involves systematically characterizing task relations, as illustrated by the hypothetical odor-mixture classification tasks in Fig. \ref{fig:Schematics}a. Task relations can vary depending on the overlap of the odors, whether overlapping odors are relevant to classification, and whether task rules are similar -- different aspects of relations may well have diverging impacts on CL performance. This underscores the need for systematic, theory-motivated quantification of CL-relevant aspects of task relations. While recent work has proposed metrics of task relations \cite{bennani2020generalisation,doan2021theoretical}, whether and how they may be used to predict CL performance remains unclear.

In this work, we utilize a Gibbs formulation of CL and use tools from statistical physics to develop a comprehensive theory of CL in deep NNs. Critically, our results do not rely on data assumptions, enabling analysis across a broad range of tasks. The theory connects the degree of forgetting and anterograde interference during CL to task relations, the NN's architecture, and hyper-parameters of the learning process. A key contribution is the identification of scalar order parameters (OPs) quantifying different aspects of task relations that are predictive of CL performance, as summarized in Table \ref{Table1:OPsummary}. For NNs without task-specific readouts, our theory identifies two critical OPs: one measuring relevant-feature similarity and another quantifying rule similarity. For networks with task-dedicated readouts, we introduce a third OP that captures overall task similarity. We also uncover three distinct CL regimes determined by this task-similarity OP and the ratio of training data to network width. While task-dedicated readouts generally reduce forgetting, sequentially learning dissimilar tasks can lead to “catastrophic anterograde interference”, where previous learning causes overfitting to the latest task. Our results offer a rigorous and predictive framework for understanding CL in deep NNs, highlighting measurable task and architecture factors that influence performance. Finally, we discuss the broader implications of our findings for understanding the neuroscience of CL.

\section{Gibbs Framework of  CL in Deep Neural Networks}
\label{sec:gibbsformulation}
\begin{figure}
\centering
\includegraphics[width=0.8\linewidth]{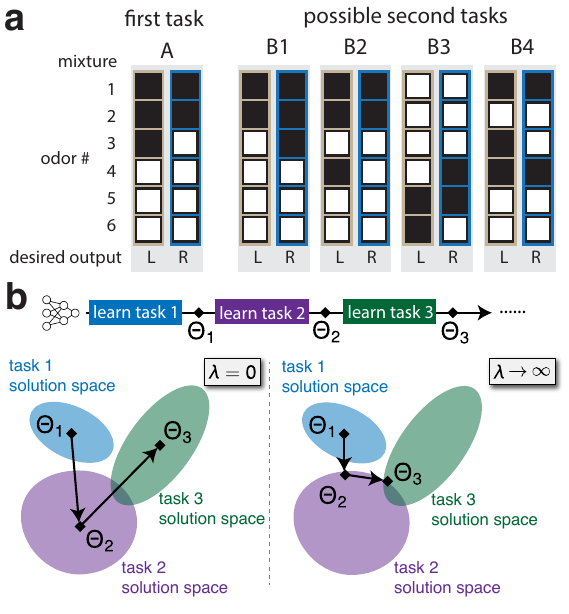}

\caption{\textbf{\protect\label{fig:Schematics}Different types of task relations in CL and the weight space schematics of the Gibbs formulation.}\protect \\
\textbf{a} Hypothetical odor-mixture classification tasks demonstrating various possible relations between two supervised-learning tasks. Each task requires classification of two odor mixtures (each column is a mixture; black/white squares indicate present/absent odors). The subject needs to respond "L" or "R" depending on the presented mixture. The first task is A; possible second tasks B1, B2, B3 and B4 exhibit different relations to A. Tasks can have highly overlapping odors (A vs. B1, A vs. B2), partially overlapping odors (A vs. B4) or entirely different odors (A vs. B3). Among these odors some are relevant for the task (e.g. odor 3 for task A) while others are not (e.g. odor 1, 2 for task A). Tasks can also share these relevant odors (A vs. B1, A vs. B4) or have different relevant odors (A vs. B2). Furthermore, with the same relevant odors, the task rules can be the same (A vs. B4) or reversed (A vs. B1). It is crucial to quantify these different aspects of relations and identify their impact on CL performance.  \\
\textbf{b} Weight-space schematics showing the Gibbs formulation of CL. Each dataset defines a space of solutions where the training loss is zero. The network learns the first task by sampling from its space of solutions. For subsequent tasks, the network assumes different solutions depending on the regularization strength ($\lambda$). At $\lambda=0$, learning of each task samples from its corresponding space of solutions independent of previous learning. At the other extreme of $\lambda\rightarrow\infty$, learning chooses the solution closest to the weights sampled while learning the previous task. These schematics assume $\beta\rightarrow\infty$.}

\end{figure}

We studied a task-based CL setting \citep{aljundi2019task} where
the network learns a sequence of $T$ tasks, respectively represented
by training datasets $D_{1},...,D_{T}$ of identical size. $D_{t}\equiv\left(\bm{X}_{t}\in\mathbb{R}^{P\times N_{0}},\bm{Y}_{t}\in\mathbb{R}^{P}\right)$,
where $P$ is the number of examples per task and $N_{0}$ is the
input dimensionality. Each row of $\bm{X}_{t}$, $\bm{ x}_{t}^{\mu}$,
is an input example with its corresponding label given by the $\mu$-th
element of $\bm{Y}_{t}$, $y_{t}^{\mu}$. While learning task $t$,
the network accesses $D_{t}$ but not the other datasets. We use ``at
time $t$'' to refer to the state of the network after sequentially
learning $D_{1}$ through $D_{t}$.

We first considered the simplest architecture: a multi-layer perceptron
where all weights are shared across tasks (``single-head'' CL \citep{farquhar2019unifying,el2019strategies}).
The network has $L$ fully-connected hidden layers, each containing
$N$ nonlinear neurons, assumed to be ReLU for concreteness. The network
load is denoted $\alpha\equiv P/N$. The input-output mapping of the
network at time $t$ is given by 
\begin{equation}
f_{t}\left(\bm{x}\right)\equiv\frac{1}{\sqrt{N}}\bm{ a}_{t}\cdot\Phi(\mathcal{W}_{t},\bm{x}),
\end{equation}
where $\bm{a}_{t}\in\mathbb{R}^{N}$ and $\mathcal{W}_{t}$ are the
readout and hidden-layer weights at time $t$, respectively. $\Phi\left(\mathcal{W}_{t},\bm{x}\right)\in\mathbb{R}^{N}$
is the activation vector in the last hidden layer for input $\bm{x}\in\mathbb{R}^{N_{0}}$.
The more complex ``multi-head'' CL scenario, where the network utilizes
task-specific readouts, is introduced and studied later.

We assume that learning $D_{t\geq2}$ involves selecting the weights
$\Theta_{t}\equiv\left\{ \bm{a}_{t},\mathcal{W}_{t}\right\} $ according
to a cost function 
\begin{align}
 & E\left(\Theta_{t}|\Theta_{t-1},D_{t}\right)\equiv\frac{1}{2}\sum_{\mu=1}^{P}\left(f_{t}\left(\bm{x}_{t}^{\mu}\right)-y_{t}^{\mu}\right)^{2}\label{eq:energy=000020fn}\nonumber \\
 & +\frac{1}{2}\beta^{-1}\sigma^{-2}\left\Vert \Theta_{t}\right\Vert ^{2}+\frac{1}{2}\beta^{-1}\lambda\left\Vert \Theta_{t}-\Theta_{t-1}\right\Vert ^{2}.
\end{align}
The first term measures the error of $f_{t}$ on $D_{t}$. The second
term acts as regularization that favors weights with small norm, which is known to encourage good generalization \citep{neyshabur2017exploring}. The
third term is a perturbation penalty that favors small weight changes
relative to $\Theta_{t-1}$, a natural strategy for mitigating forgetting \citep{xu2009rapid,yang2009stably,kirkpatrick2017overcoming,zenke2017continual}. $\beta>0$ denotes the inverse temperature,
controlling how well $f_{t}$ interpolates $D_{t}$. $\sigma^{-2}>0$
and $\lambda\geq0$ respectively scale the $L_{2}$ regularization
and the perturbation penalty. 
The cost function for learning $D_{1}$,
$E\left(\Theta_{1}|D_{1}\right)$, has the same form but without the
penalty.

Learning at each stage is modeled as a posterior distribution  of $\Theta_{t}$
conditioned on $\Theta_{t-1}$, $P\left(\Theta_{t}|\Theta_{t-1},D_{t}\right)\propto \exp\left(-\beta E\left(\Theta_{t}|\Theta_{t-1},D_{t}\right)\right)$.
This distribution defines a Markovian transition from $\Theta_{t-1}$
to $\Theta_{t}$, controlled by $D_{t}$. Multiplying all such transition matrices 
for $t\geq2$ and the posterior of learning the first task, $P\left(\Theta_{1}|D_{1}\right)\propto \exp\left(-\beta E\left(\Theta_{1}|D_{1}\right)\right)$,
yields a joint posterior over $\Theta_{1},...,\Theta_{T}$ \citep{franz1995recipes,saglietti2022analytical,li2023statistical}.
This posterior fully describes how the network evolves
during CL, from which various statistics can be calculated. We focus
on over-parameterized NNs \citep{poggio2020theoretical} in the
$\beta\rightarrow\infty$ limit. In this case, there is a large space of $\Theta_{t}$ that perfectly interpolates the dataset $D_{t}$ (Fig. $\text{\ref{fig:Schematics}}$b). At $\lambda=0$,
there is no coupling between weights from different times, and the
network has no memory of previous tasks. On the other hand, at $\lambda\rightarrow\infty$,
the network makes the minimum perturbation to weights that is required to
interpolate $D_{t}$ \citep{shan2022minimum}.  
Performance of the network on some dataset $D=\left(\bm{X},\bm{Y}\right)$
at time $t$ is measured by averaging the normalized mean-squared-error
(MSE) loss,
\begin{equation}
\mathcal{L}(f_{t},D)\equiv\sum_{\mu}\left(f_{t}\left(\bm{x}^{\mu}\right)-y^{\mu}\right)^{2}/\left\Vert \bm{Y}\right\Vert ^{2},\label{eq:loss}
\end{equation}
over the posterior of $\Theta_{t}$, denoted $\langle\mathcal{L}(f_{t},D)\rangle$.



\section{Networks Using a Readout Shared Across Tasks}
\label{sec:single-head}
Our theory of single-head CL, exact in the \textit{infinite-width limit} of $N\rightarrow\infty,\alpha\rightarrow0$,
allows analytical evaluations of $\langle\mathcal{L}(f_{t},D)\rangle$ for arbitrarily long task sequences (SI Eqs. \ref{SIeq:meanpredictor}, \ref{SIeq:v}). For intuition, we here discuss a naively simplified version of the full theory that (1) nevertheless reproduces the key qualitative behaviors of the full theory (Fig. \ref{fig:comparisonfull}) and (2) provides simple geometric intuitions of task relations. We also focus our discussion here to the case where forgetting is minimized by taking $\lambda \rightarrow \infty$ and $\sigma \ll 1$ (SI \ref{subsec:singleheadtheory}.\ref{paragraph:statisticssingle}).




\subsection{Task-Relation Order Parameters (OPs)}
\label{subsec:task-relation OPs}

\begin{figure*}[h]
\centering
\includegraphics[width=0.8\textwidth]{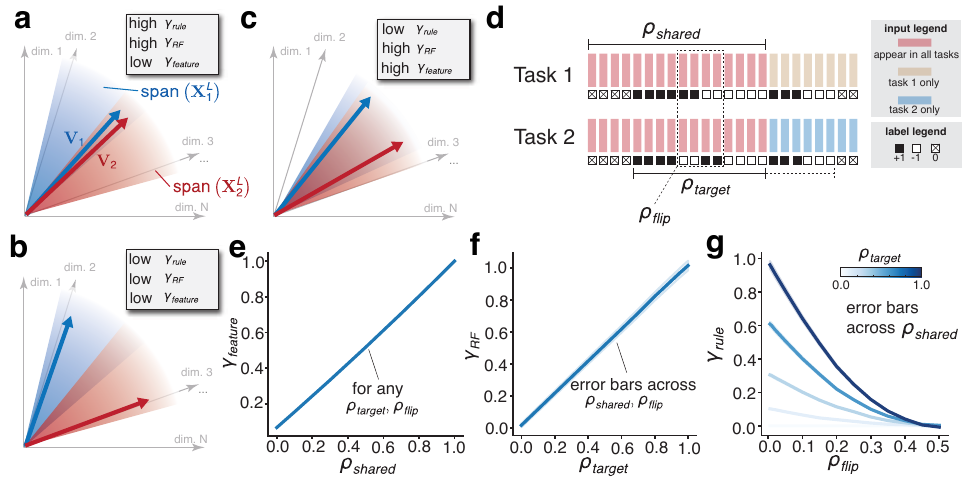}
\caption{\textbf{Schematics of the OPs and the target-distractor task.} \\
\textbf{a}-\textbf{c} Schematics of the OPs. The input features of each task span a P-dimensional subspace in the N dimensional feature space. $\text{span}(\bm{X}_{1}^{L})$ (shown in blue) denotes the space spanned by task 1 input features at the L-th layer, while $\text
{span}(\bm{X}_{2}^{L})$ (shown in red) denotes the space spanned by task 2 input features at the L-th layer. $\bm{V}_{1}$ and $\bm{V}_{2}$ are the rule vectors of task 1 and task 2, and by definition lie in $\text{span}(\bm{X}_{1}^{L})$ and $\text{span}(\bm{X}_{2}^{L})$ respectively. $\gamma_{\text{RF}}$ measures how much the rule vectors project onto the shared feature dimensions. In a, c, both rule vectors fully lie in the shared subspace and $\gamma_{\text{RF}}$ is high. In contrast, in b the rule vectors are away from the shared subspace, thus $\gamma_{\text{RF}}$ is small. $\gamma_{\text{rule}}$ measures the similarity between the projection of the rule vectors on to the shared feature dimensions. In a $\gamma_{\text{rule}}$ is high and in b, c $\gamma_{\text{rule}}$ is low. $\gamma_{\text{feature}}$ measures the degree of overlap between the shared feature dimensions, and is low in a, b but high in c. \\
\textbf{d} Schematics of the target-distractor task. Each task consists of a set of P images (rectangles) from CIFAR-100, assigned labels $\pm1$ or 0 (squares). $\rho_{\text{shared}}$ controls the ratio of shared images between two tasks. $\rho_{\text{target}}$ controls the ratio of images with $\pm 1$ labels that are shared between the tasks. For the shared images, some of the labels are flipped between the tasks, and $\rho_{\text{flip}}$ controls the ratio of the images with flipped labels. Varying these parameters allows us to explore the full range of the OPs.\\
\textbf{e}-\textbf{g} Controlling the 3 OPs with the target-distractor task. $\gamma_{\text{feature}}$ depends only on $\rho_{\text{shared}}$ (e), as $\rho_{\text{shared}}$ increases, the overlap between the shared feature subspaces increases, resulting in higher $\gamma_{\text{feature}}$. $\gamma_{\text{RF}}$ depends mainly on $\rho_{\text{target}}$ (f). As $\rho_{\text{target}}$ increases, the rule vectors project more onto the shared feature dimensions, thus $\gamma_{\text{RF}}$ increases. $\gamma_{\text{rule}}$ is tuned by both $\rho_{\text{target}}$ and $\rho_{\text{flip}}$ (g). $\rho_{\text{target}}$ sets an upper bound of $\gamma_{\text{rule}}$, for a fixed $\rho_{\text{target}}$, $\gamma_{\text{rule}}$ decreases with $\rho_{\text{flip}}$. At $\rho_{\text{flip}}=0.5$ about half of the labels are flipped, and $\gamma_{\text{rule}}$ goes to 0.}
\label{fig2:OPschematics}
\end{figure*}

The naive simplification is motivated by the well-known observation that learning in networks in the \textit{infinite-width limit} tends to induce only small modifications of the hidden weight matrices \cite{lee2017deep}. In the simplified theory, we neglect learning-induced changes to the hidden-layer weights and assume them to be fixed at $\mathcal{W}_0 \sim \mathcal{N}(0,\sigma^2)$ throughout learning. Thus, CL in the network can be viewed as solely learning the readout from a fixed feature layer. We denote the training data of the two tasks as $\{(\bm{X}_i, \bm{Y}_i)\}_{i \leq 2}$, and the features at the $L$-th hidden layer as $\bm{X}_i^L \in\mathbb{R}^{P\times N}$, where the $\mu$-th row is given by $\Phi({\mathcal W}_0, \bm{x}_i^{\mu})$. 

In this case, short-term forgetting, defined as the error on the training data of task 1 after learning two tasks ($F_{2,1}$), adopts a simple form:
\begin{equation}
\label{eq:f21}
    F_{2,1} = 2(\gamma_{\text{RF}} -\gamma_{\text{rule}})
\end{equation}
where
\begin{align}
\gamma_{\text{RF}}&\equiv \bm{V}_2^\top \mathcal{P}_{12} \bm{V}_2\\
\gamma_{\text{rule}}&\equiv \bm{V}_2^\top \mathcal{P}_{12} \bm{V}_1.
\end{align}
$ \mathcal{P}_{12}$ and $\bm{P}_{i=1,2}$ are $N \times N$ matrices,
\begin{align}
   \mathcal{P}_{12} &\equiv \frac{1}{2N}\left(\bm{P}_2 \left(\bm{X}_1^{L}\right)^\top \bm{X}_1^L \bm{P}_2 + \bm{P}_1 \left(\bm{X}_2^{L}\right)^\top\bm{X}_2^L\bm{ P}_1\right)  \\
\bm{P}_i &\equiv \left(\bm{X}_i^L\right)^\top \left[ \bm{X}_i^L \left(\bm{X}_i^L\right)^\top\right]^{-1}\bm{ X}_i^L \in \mathbb{R}^{N\times N} \label{eq:projection}
\end{align}
and $\bm{V}_{i=1,2}$ are $N$-dimensional vectors, 
\begin{equation}
 \bm{V}_i \equiv \sqrt{N}\left(\bm{X}_i^L\right)^\top \left[ \bm{X}_i^L \left(\bm{ X}_i^L\right)^\top\right]^{-1} {\bm{Y}}_i /\lVert \bm{Y}_i\rVert.\label{eq:rule}
 \end{equation}
These expressions allow us to describe task relations in terms of the geometry in the network's $N$-dimensional feature space. $\bm{P}_i$ projects vectors in this space onto the $P$-dimensional subspace spanned by the $P$ training examples of task $i$, $\bm{X}_i^L$. $\mathcal{P}_{12}$ represents the projection of the input features of one task onto those of the other task.  $\bm{V}_i$ is the normalized readout from the top hidden layer of the network, generated when learning  $\left({\bm{X}}_i, {\bm{Y}}_i\right)$ {\it alone}. We refer to it as the rule vector since it fully characterizes the learned input-output rule of each task.
The two terms in Eq. \ref{eq:f21} have interesting geometrical interpretations in the feature space, as illustrated in Fig. \ref{fig2:OPschematics}a-c. The first term, denoted as $\gamma_{\text{RF}}$, measures how much the rule vectors of \emph{individual tasks} project onto the shared input feature subspace of both tasks. Intuitively, it corresponds to the similarity of the input features which are \emph{relevant to the task rules}, and thus referred to as the relevant-feature (RF) similarity. The second term, $\gamma_{\text{rule}}$, measures the similarity between the two task rule vectors projected onto the shared input feature subspace, and thus referred to as the rule similarity. We note that both  $\gamma_{\text{RF}}$ and  $\gamma_{\text{rule}}$ depend on the target outputs \textit{and} the input features. It is also instructive to  define a third OP,  $\gamma_{\text{feature}} \equiv P^{-1}\mathrm{Tr}(\bm{P}_1\bm{P}_2)$, which measures the degree of overlap between the input feature subspaces. It is thus a similarity metric between the input feature vectors, independent of the target outputs. By definition, they are algebraically bounded: $\gamma_{RF} \geq 0$, $\gamma_{\text{rule}}\in [-\gamma_{\text{RF}}, \gamma_{\text{RF}}]$ and $\gamma_{\text{features}} \in [0, 1]$. Under reasonable assumptions (SI \ref{subsec:singleheadtheory}.\ref{paragraph:theops}) we further have $\gamma_{\text{RF}} \in [0,1]$ and $\gamma_{\text{rule}} \in [0,\gamma_{\text{RF}}]$, which empirically hold for our results throughout this paper. 

In summary, the quantity $\gamma_{\text{RF}} - \gamma_{\text{rule}}$ provides a direct measure of  short-term forgetting (Eq. \ref{eq:f21}), which we hereafter refer to as the ``conflict'' between two tasks. Conflict can be small under two scenarios. The first occurs when both $\gamma_{\text{RF}}$ and $\gamma_{\text{rule}}$ are large but close in value, indicating that the tasks have similar relevant features \textit{and} similar rules. The second is when both OPs are small, corresponding to the case where the tasks have dissimilar relevant features \emph{and} rules. In contrast, two tasks would have high conflict if they share a lot of the relevant features (high $\gamma_{\text{RF}}$) but use different rules (low $\gamma_{\text{rule}}$).

\subsection{Exploring the OPs with Target-Distractor Task Sequences}
\label{subsec:opdistractor}
We next sought to understand how the 3 OPs are tied to the severity of forgetting. We began by studying them in the setting of task sequences with parametrically controllable input distributions and task rules. We constructed ``target-distractor task sequences'' (full details in SI \ref{subsec:targetdistractor}), where each task consists of a set of $P$ randomly selected stimuli from a large pool of stimuli, e.g., $P$ images from CIFAR-100 \citep{krizhevsky2009learning},  with random labels $\pm 1$ or $0$ for each task (Fig. \ref{fig2:OPschematics}d). A subset of the inputs is shared across all tasks in the sequence, whereas other inputs are unique to each task. The parameter $\rho_{\text{shared}} \in [0,1]$ represents the proportion of shared inputs among all $P$ inputs across all tasks in the sequence and controls $\gamma_{\text{feature}}$ (Fig. \ref{fig2:OPschematics}e). Another parameter $\rho_{\text{target}} \in [0,1]$ represents the proportion of shared inputs \emph{among the inputs with $\pm 1$ labels} across all tasks; varying $\rho_{\text{target}}$ controls $\gamma_{RF}$ (Fig. \ref{fig2:OPschematics}f). 
For the shared images, part of the $\pm 1$ labels are flipped between tasks, and $\rho_{\text{flip}} \in [0, 0.5]$ controls the ratio of images with flipped labels between tasks on average. It thereby affects how consistent the rules for different tasks are, as reflected in $\gamma_{\text{rule}}$ (Fig. \ref{fig2:OPschematics}g). Therefore, through varying $(\rho_{\text{shared}}, \rho_{\text{target}}, \rho_{\text{flip}})$, we can explore the full range of the 3 OPs and elucidate their respective roles in forgetting.   

As expected from Eq. \ref{eq:f21}, $F_{2,1}$ is captured by the conflict (Fig. \ref{fig3:distractortask}a) and is independent of $\gamma_{\text{feature}}$ (Fig. \ref{fig3:distractortask}c, top panel). We also measured the effect of $\lambda$ by defining $\Delta F_{2,1} \equiv F_{2,1} - F_{2,1}(\lambda = 0)$, where $F_{2,1}(\lambda=0)$ denotes forgetting on the first task without regularization (Fig. \ref{fig3:distractortask}a, inset). For tasks with low conflict, the effect of regularization on mitigating forgetting decreases as the tasks become more similar, as measured by $\gamma_{\text{RF}}$ (since the conflict is small, $\gamma_{\text{rule}}$ is close to $\gamma_{\text{RF}}$ for these tasks).

\subsection{Long-Term Forgetting}
\begin{figure}[H]
\centering
\includegraphics[width=0.9\linewidth]{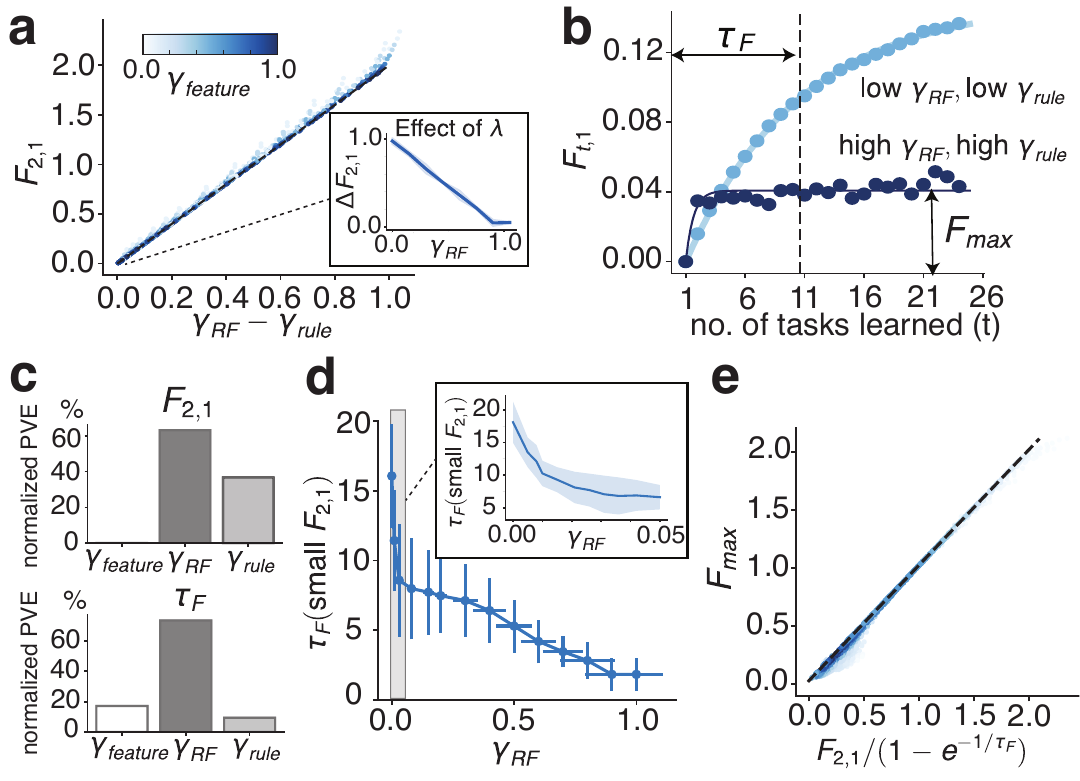}
\caption{\textbf{OPs predict short-term and long-term forgetting behaviors in target-distractor sequences.} \label{fig3:distractortask} \\
\textbf{a} Forgetting on the training data of the first task after learning two tasks ($F_{2,1}$) is accurately predicted by 2($\gamma_{\text{RF}}-\gamma_{\text{rule}})$,and does not depend on $\gamma_{\text{feature}}$ (represented by the color of the points). Each point represents a target-distractor task sequence with a different set of $(\rho_{\text{target}},\rho_{\text{shared}},\rho_{\text{flip}})$. Inset: $\Delta F_{2,1}$ measures the effect of the regularizer. When $F_{2,1}$ is small ($<0.05$), $\gamma_{\text{RF}}$ and $\gamma_{\text{rule}}$ are close, the effect of the regularizer decreases as $\gamma_{\text{RF}}$ (and $\gamma_{\text{rule}}$) increases, i.e., as the tasks become more similar.\\
\textbf{b} Long-term forgetting in a task sequence can be approximated by an exponential relaxation process, where $F_{\text{max}}$ denotes its asymptote as $t\rightarrow\infty$, and $\tau_{F}$ denotes its time constant (see SI \ref{sec:exponential}). We show two examples, one for dissimilar tasks (low $\gamma_{\text{RF}}$, low $\gamma_{\text{rule}}$) with relatively large $\tau_F$ and $F_{\text{max}}$, and the other for similar tasks (high $\gamma_{\text{RF}}$, high $\gamma_{\text{rule}}$), with relatively small $\tau_{F}$ and $F_{\text{max}}$.\\
\textbf{c} Normalized PVE (proportion of variance explained, \cite{o1982measures}) of $F_{2,1}$ and $\tau_{F}$ by the 3 OPs (see SI \ref{subsec:targetdistractor}). $F_{2,1}$ depends on $\gamma_{\text{rule}}$ and $\gamma_{\text{RF}}$ and is independent of $\gamma_{\text{feature}}$, consistent with a. $\tau_F$ mainly depends on $\gamma_{\text{RF}}$ and weakly depends on $\gamma_{\text{rule}}$ and $\gamma_{\text{feature}}$. \\\ 
\textbf{d} For tasks where $F_{2,1}$ is small ($<0.05$), $\tau_{F}$ decreases as $\gamma_{\text{RF}}$ increases. Inset: zoomed-in region of $\gamma_{\text{RF}}<0.05$ highlights the fast decrease of $\tau_{F}$ for small $\gamma_{\text{RF}}$. Data is binned by $\gamma_{\text{RF}}$, errorbars are standard deviations across data points within the same bin.\\
\textbf{e} $F_{\text{max}}$ can be accurately predicted given $F_{2,1}$ and $\tau_{F}$. Each point represents a target-distractor task sequence with a different set of $(\rho_{\text{target}},\rho_{\text{shared}},\rho_{\text{flip}})$. The color represents the density of points.\\
All $F_{t,1}$ and the corresponding OPs are averaged across 40 random seeds used for generating data. See SI \ref{subsec:targetdistractor} for detailed parameters.}
\end{figure}

We next studied forgetting after learning sequences of multiple tasks, which we refer to as long-term forgetting. Here we assume that all tasks in each sequence have identical pairwise relations. Empirically, we observed that forgetting of the first task ($F_{t,1}$) increases over time approximately as an exponential relaxation process, $F_{t,1} \approx F_{\text{max}}(1-e^{-(t-1)/\tau_F})$. We thus characterized long-term forgetting by its time constant $\tau_F$ and long-time asymptote $F_{max}$ (Fig. \ref{fig3:distractortask}b). Interestingly, task sequences can have similar short-term forgetting but very different long-term forgetting behaviors (Fig. \ref{fig3:distractortask}b). This suggests that $\tau_F$ and $F_{\text{max}}$ depend on the OPs in ways that differ from short-term forgetting ($F_{2,1}$). We analyzed the variance of $\tau_F$ explained by the 3 OPs (Fig. \ref{fig3:distractortask}c bottom panel). $\tau_F$ mainly depends on $\gamma_{\text{RF}}$, and weakly depends on the other two OPs.

Since $F_{t,1}$ can only be relatively mild when $F_{2,1}$ is small, we are particularly interested in long-term forgetting at small $F_{2,1}$. For task sequences with small $F_{2,1}$ ($F_{2,1} < 0.05$), $\tau_F$ decreases with $\gamma_{\text{RF}}$ (Fig. \ref{fig3:distractortask}d). The decrease is very fast around $\gamma_{\text{RF}}$ close to 0 (Fig. \ref{fig3:distractortask}d, inset), and slows down afterwards. Finally, since the exponential fit of $F_{t,1}$ is remarkably accurate across all parameters of the target-distractor task,  characterizing the relation between the OPs and $F_{2,1}$ and $\tau_F$ also characterizes $F_{\text{max}}$ (Fig. \ref{fig3:distractortask}e). 

In summary, our results suggest that while short-term forgetting is small as long as the tasks have low conflict (small $\gamma_{\text{RF}} - \gamma_{\text{rule}}$), long-term forgetting can still vary depending on the similarity (magnitude of $\gamma_{\text{RF}}$ and $ \gamma_{\text{rule}}$) between tasks. Forgetting tends to accumulate slowly for dissimilar tasks over time but quickly rises and plateaus for similar tasks (smaller $\tau_F$). On the other hand, neither short-term nor long-term forgetting depends on $\gamma_{\text{feature}}$, suggesting that a task-relation metric that includes only the input features is likely not informative of CL performance. Instead, it is crucial to consider the interaction between the task rules and the input features.

\label{subsec:orderparameters}

\subsection{The Effect of Depth on Forgetting in Benchmark Sequences}

Having varied the task relations to explore the entire OP space using the target-distractor task sequences, we next studied the effects of the network depth, $L$. $L$ affects CL performance by modifying $\bm{X}^L_i$. To study the effect of $L$ on general data, we analyzed the forgetting on several benchmark task sequences in NNs with different depths ($L=1,3,5,7,9$). Following standard practices, we created each sequence by applying a generation protocol (specified below) to a multi-way classification dataset (``source datasets''):
MNIST \citep{lecun1998gradient}, EMNIST \citep{cohen2017emnist}, Fashion-MNIST \citep{xiao2017fashion}, or CIFAR-100. To generate long task sequences ($T\gg2$) for the long-term forgetting analysis, we used the split protocol on EMNIST and CIFAR-100 and the permutation protocol on all source datasets, where a higher ``permutation ratio'' corresponds to less similar
inputs between tasks \citep{zenke2017continual} (see SI \ref{sec:benchmark} for further details). 

For permutation sequences, we aggregated over different source datasets as forgetting does not differ much between them. For split sequences, forgetting varies more between split EMNIST and split CIFAR-100, so we present their results separately. For all benchmark task sequences we explored, $F_{2,1}$ monotonically decreases with depth (Fig. \ref{fig4:OPsandbenchmark}a). $F_{2,1}$ is larger for split sequences and smaller for permutation sequences, and increases with a larger permutation ratio. As before, we are primarily interested in long-term forgetting for tasks with small $F_{2,1}$, we thus show $\tau_F$ and $F_{\text{max}}$ for permutation sequences with 5-15\% permutation ratio. As shown in Fig. \ref{fig4:OPsandbenchmark}b, $\tau_F$ increases with depth. Therefore, depth has opposing effects on $F_{2,1}$ and $\tau_F$, while deeper networks forget less in the short term, forgetting accumulates over longer periods of time. As a result, the dependence of $F_{\text{max}}$ on depth is more complex. $F_{\text{max}}$ does not change as significantly as $F_{2,1}$, and there may be an optimal depth where $F_{\text{max}}$ is at its lowest (Fig. \ref{fig4:OPsandbenchmark}c). 

\subsection{Forgetting in Benchmark Sequences are Explained by the OPs}
\label{subsec:benchmarkdatasets} 

In this section, we examine whether the dependence of forgetting on depth and across different benchmark sequences can be explained by the OPs as in the target-distractor sequences. To this end, we first computed the OPs for the task sequences, aggregated over the source datasets for the permutation sequences, and separately for split EMNIST and split CIFAR-100 sequences. As we showed in Section \ref{sec:single-head}.\ref{subsec:opdistractor}, forgetting does not exhibit significant dependence on $\gamma_{\text{feature}}$, so we focus on $\gamma_{\text{rule}}$ and $\gamma_{\text{RF}}$ in this section.

\begin{figure}[H]
\centering
\includegraphics[width=1\linewidth]{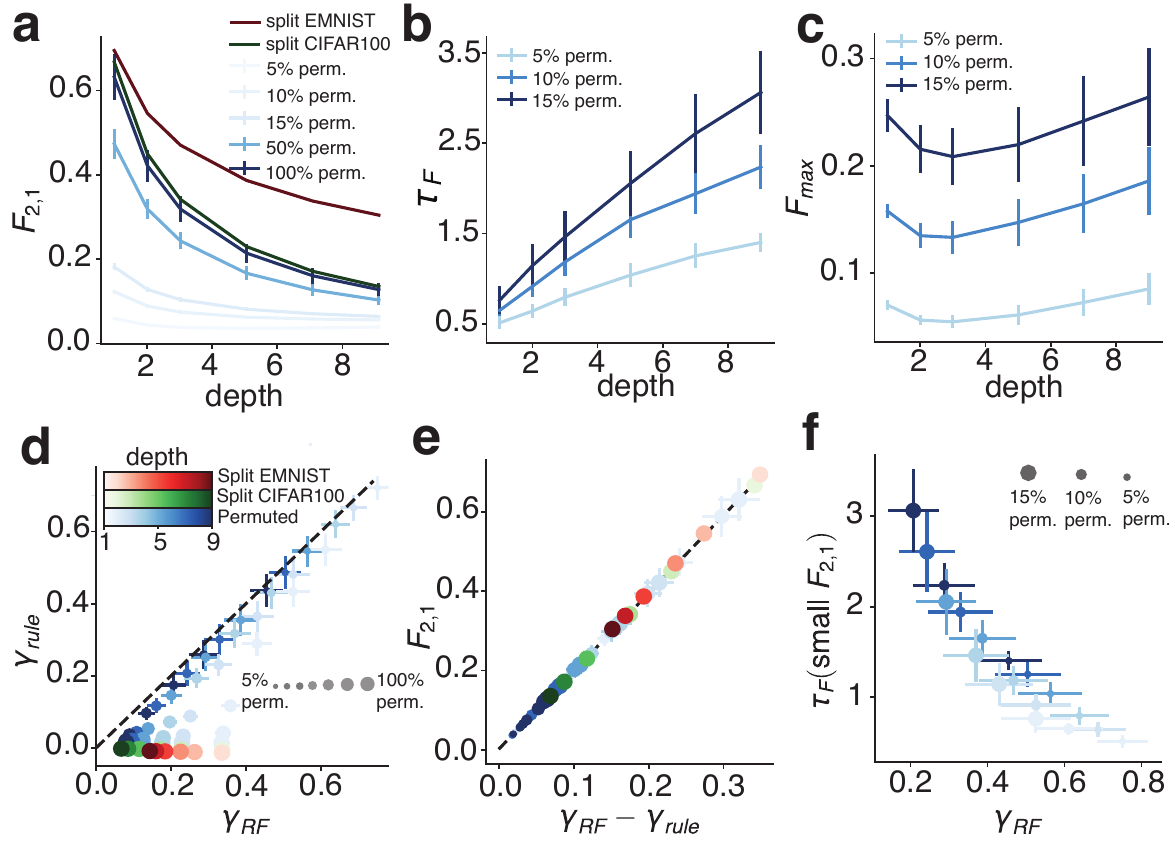}
\caption{\textbf{Forgetting in benchmark task sequences and the effect of depth.}\\
\textbf{a}-\textbf{c} The effect of depth on short-term ($F_{2,1}$) and long-term ($\tau_{F}$, $F_{\text{max}}$) forgetting on benchmark sequences. For permutation sequences, we averaged over source datasets including MNIST, EMNIST, Fashion-MNIST and CIFAR as their behaviors are similar, and errorbars are standard errors across the source datasets. For split sequences, we show separately split EMNIST and split CIFAR-100 sequences as their behaviors are more different. In all cases, $F_{2,1}$ decreases with depth (a). We look at long-term forgetting only in sequences with small $F_{2,1}$ (5, 10, 15\% permutation sequences), $\tau_{F}$ increases with depth (b). Due to the opposing behaviors of $F_{2,1}$ and $\tau_{F}$, $F_{\text{max}}$ does not vary strongly with depth, and there may be an optimal depth where $F_{\text{max}}$ is lowest (c). \\
\textbf{d} Task-relation OPs ($\gamma_{\text{rule}}$ and $\gamma_{\text{RF}}$) on the benchmark sequences ($\gamma_{\text{feature}}$ is not shown as it does not strongly affect either $F_{2,1}$ or $\tau_{F}$, as shown in Fig. \ref{fig3:distractortask}). Colors blue, red and green correspond to permutation sequences, split EMNIST and split CIFAR100 respectively. Colors from light to dark correspond to increasing depth of the network. In permutation sequences, larger sizes of the points correspond to larger permutation ratio. The benchmark sequences explore a more constrained region in the OP space compared to the target-distractor sequences. \\
\textbf{e} $F_{2,1}$ is accurately predicted by $2(\gamma_{\text{RF}}-\gamma_{\text{rule}})$, as in Fig. \ref{fig3:distractortask}a. Increasing depth or decreasing the permutation ratio results in smaller $\gamma_{\text{RF}}-\gamma_{\text{rule}}$ (as also shown in d), and thus leads to smaller $F_{2,1}$.\\
\textbf{f} For tasks with small $F_{2,1}$ (5,10,15\% permutation sequences), $\tau_{F}$ decreases with $\gamma_{\text{RF}}$, consistent with Fig. \ref{fig3:distractortask}d. For a fixed permutation ratio, increasing depth results in smaller $\gamma_{\text{RF}}$ (as also shown in d), and thus leads to larger $\tau_{F}$.\\
All $F_{t,1}$ and the corresponding OPs are averaged across 50 random seeds used for generating data. See SI \ref{sec:benchmark} for detailed parameters.} \label{fig4:OPsandbenchmark}
\end{figure}

As shown in Fig. \ref{fig4:OPsandbenchmark}d, the OPs on the benchmark sequences depend on both the sequence type and the network depth, and altogether partially fill the entire feasible OP space (below the dashed line in Fig. \ref{fig4:OPsandbenchmark}d). Split sequences including split EMNIST (red points) and split CIFAR-100 (green points) have close to zero $\gamma_{\text{rule}}$, and $\gamma_{\text{RF}}$ decreases with depth (darker colors represent increasing depths). Permutation sequences with large permutation ratios (large blue points) behave similarly to the split sequences. Permutation sequences with small permutation ratios (small blue points) have $\gamma_{\text{RF}}$ close to $\gamma_{\text{rule}}$ across all depths, and for networks with larger depths, $\gamma_{\text{rule}}$ and $\gamma_{\text{RF}}$ decrease simultaneously, such that the tasks become more dissimilar. 

To summarize, between-task conflict becomes lower in deeper networks, and in permutation sequences with lower permutation ratio. As verified in Fig. \ref{fig4:OPsandbenchmark}e, $F_{2,1}$ is still accurately predicted by the conflict. For task sequences with small conflict (permutation sequences with 5-15\% permutation ratio), both OPs decrease with depth, predicting a larger $\tau_F$, as verified in Fig. \ref{fig4:OPsandbenchmark}f. The relations between the OPs and forgetting in the target-distractor sequences in Section \ref{sec:single-head}.\ref{subsec:opdistractor} also hold for general benchmark task sequences, and can be used to explain the effect of network depth. 

\section{Networks with Task-Dedicated Readouts }
\label{sec:multi-head}
\subsection{Setup of Multi-Head CL}
\label{subsec:multi-head}
In many CL settings, both in ML applications and naturalistic environments for animals, the learner is aware (through external cues or inference) of the identity of the current task being learned or performed. A simple method of incorporating such information into the network \cite{farquhar2019unifying,el2019strategies,li2021statistical} is to use task-specific readouts (``multi-head'' CL). Learning a new task involves modifying the shared  hidden-layer weights while adding  a new task-specific
readout, leaving previous readouts untouched (Fig. \ref{fig5:ptdistractor}a). The network has $t$ different input-output mappings after learning
$t$ tasks, given by 
\begin{equation}
f_{t}^{\tau}({\bf x})\equiv\frac{1}{\sqrt{N}}\bm{a}_{\tau}\cdot\Phi\left(\mathcal{W}_{t},\bm{x}\right)\quad\tau=1,\cdots,t.\label{eq:predictor-statistics}
\end{equation}
At time $t$, the network selects the mapping $f_{t}^{\tau}(\bm{x})$ to perform the $\tau$-th task. Since the readout weights $\left\{ \bm{a}_{\tau}\right\} _{\tau=1,\cdots,t}$
are task-dedicated and the hidden-layer weights $\mathcal{W}$ are shared, only the changes in $\mathcal{W}$ need to be constrained
in order to mitigate forgetting. Due to these differences from single-head CL, the objective function of learning  is given by Eq. \ref{eq:energy=000020fn} but with $f_{t}(\bm{x})$ replaced with $f_{t}^{t}(\bm{x})$ and the
regularization term $\lVert\Theta_{t}-\Theta_{t-1}\rVert^{2}$ replaced with $\lVert\mathcal{W}_{t}-\mathcal{W}_{t-1}\rVert^{2}$.

The presence of task-specific parameters generally makes forgetting less severe than that in single-head CL \citep{farquhar2018towards}. Importantly, this architecture allows the network to perform tasks with high conflict, which single-head networks struggle with, as shown in the previous section. In fact, in the infinite-width
limit ($N\rightarrow\infty, \alpha\rightarrow 0$) studied above, forgetting and anterograde effects can be entirely avoided regardless of task relations, since the network can simply freeze its random hidden-layer weights and learn a separate readout for each task. However, this simple scheme breaks down in the more realistic cases where resources are limited
and the network may have to modify the hidden-layer weights to solve each task. 

To study interesting properties of CL in the task-dedicated multi-head architecture, we focus on  the \emph{thermodynamic}
limit, defined by $P,N\rightarrow\infty$ and $\alpha\equiv P/N \sim\mathcal{O}(1)$. We focused on the case of $T=2$ and $L=1$ (due to the complexity of the theory in this limit, but see Figs. \ref{fig:gdmulti}, \ref{fig:taskinterpolation} for results beyond these restrictions). 
Our theory analytically evaluates forgetting of task 1 and the anterograde effect on task 2 in \emph{multi-head} CL, respectively given by $F_{2,1}=\left\langle\mathcal{L}\left(f_{2}^{1},D_{1}\right)\right\rangle $ and $G_{2,2}=\left\langle\mathcal{L}\left(f_{2}^{2},D_{2}^{test}\right)\right\rangle/G_2^0$, where $G_2^0$ is the generalization error on task 2 when learning it alone. 

\subsection{Phase Transitions in CL Performance in the Target-Distractor Sequence}

\label{subsec:phase-transition-target-distractor}

\begin{figure}[H]
\centering
\includegraphics[width=1\linewidth]{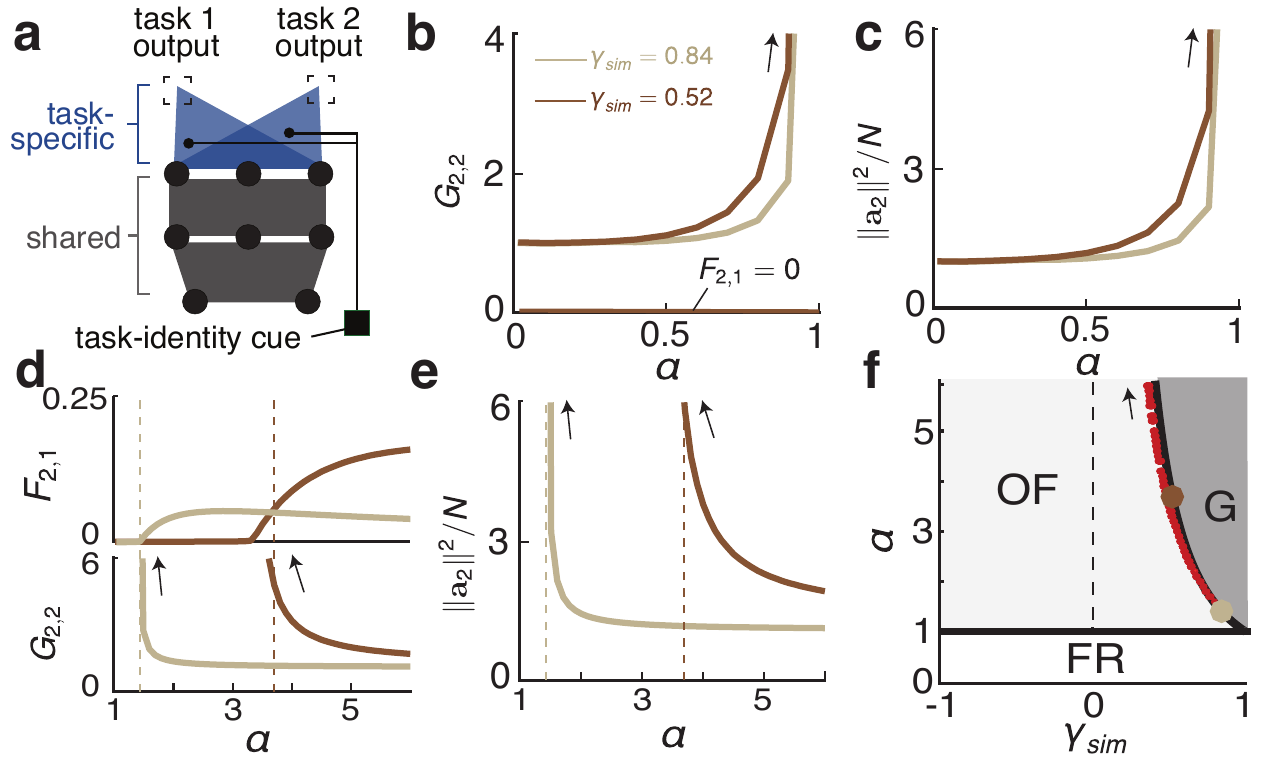}

\caption{\label{fig5:ptdistractor}\textbf{Multi-head CL exhibits phase transitions in the target-distractor sequence.}\\
\textbf{a} Schematics of multi-head CL. Different tasks utilize the same shared hidden-layer weights but different task-specific readouts. The weight-perturbation penalty is only applied to the hidden-layer weights. \\
\textbf{b} Forgetting of task 1 ($F_{2,1}$) and the normalized generalization error on task 2 ($G_{2,2}$) as a function of the network load ($\alpha$) for 2 different sets of $(\rho_{\text{target}},\rho_{\text{shared}},\rho_{\text{flip}})$ in the target-distractor task in the fixed-representation regime (FR, $\alpha<1$). Black arrows indicate divergence towards infinity as $\alpha$ approaches 1. Curves of different colors correspond to tasks with different parameters $(\rho_{\text{target}},\rho_{\text{shared}},\rho_{\text{flip}})$. light: $(1, 0.88, 0), \gamma_{\text{sim}}=0.84$; dark: $(1, 0.58, 0.005), \gamma_{\text{sim}}=0.52$. The generalization errors are calculated on the training data with small perturbations to the input (see SI \ref{subsec:targetdistractor}).\\
\textbf{c} The norm of ${\bf a}_{2}$, $\lVert{\bf a}_{2}\rVert^{2}/N$, as a function of $\alpha$ in the fixed-representation regime (FR). Since the hidden layer representations are fixed, learning the second task is equivalent to learning the linear weights ${\bf a}_{2}$ in linear regression, thus the divergence of $G_{2,2}$ results from the divergence of ${\bf a}_{2}$ when approaching the interpolation threshold in linear regression.\\
\textbf{d} Same as b, but for $\alpha>1$. For each combination of $(\rho_{\text{target}},\rho_{\text{shared}},\rho_{\text{flip}})$, $F_{2,1}$ and $G_{2,2}$ exhibit abrupt changes as $\alpha$ crosses a critical load ($\alpha_{c}$, vertical dashed line). In the overfitting regime (OF, $1<\alpha<\alpha_{c}$), $F_{2,1}$ is zero but $G_{2,2}$ diverges. In the generalization regime (G, $\alpha>\alpha_{c}$), both $F_{2,1}$ and $G_{2,2}$ are moderate and nonzero.\\
\textbf{e} Same as c, but for $\alpha>1$. The divergence of $G_{2,2}$ results from the divergence of ${\bf a}_{2}$ to compensate for minimal $\lVert\mathcal{W}_{2}-\mathcal{W}_{1}\rVert$ when learning task 2. \\ 
\textbf{f} The transition boundary between the fixed-representation regime (FR) and the overfitting regime (OF) is always at $\alpha=1$, and does not depend on the task. The transition boundary between the overfitting regime (OF) and the generalization regime (G), $\alpha_{c}$, can be theoretically predicted by the task similarity metric $\gamma_{\text{sim}}\in[-1,1]$ under reasonable assumptions (SI \ref{subsec:multiheadtheory}.\ref{subsubsec:summarymultihead}), as shown by the black line. Each red point shows the estimated transition boundary $\alpha_{c}$ from the shape of $F_{2,1}$ (SI \ref{subsec:targetdistractor}) for a different combination of $(\rho_{\text{target}},\rho_{\text{shared}},\rho_{\text{flip}})$, and thus a different value of $\gamma_{\text{sim}}$. The red points lie on top of the black curve, demonstrating the accuracy of the theoretical prediction. The light and dark brown points correspond to the lines shown in b and c.}
\end{figure}

We first used the target-distractor task sequences to probe how task relations affect CL performance in the limit of $\lambda\rightarrow\infty$.
In addition to varying $(\rho_{\text{shared}}, \rho_{\text{target}}, \rho_{\text{feature}})$ as in
the single-head analysis, we also varied the load $\alpha \equiv P/N$. In Fig. \ref{fig5:ptdistractor}b-e, we show two examples of task sequences generated under two combinations of $(\rho_{\text{shared}},\rho_{\text{target}}, \rho_{\text{feature}})$, and plot $F_{2,1}$ and $G_{2,2}$ as $\alpha$ increases. We found that, regardless of task relations, $F_{2,1}$ is zero as
long as $\alpha<1$, while $G_{2,2}$ diverges to infinity as $\alpha$ approaches $1$ (Fig. $\text{\ref{fig5:ptdistractor}}$b). Such behaviors can be explained by the fact that when $\alpha<1$, learning the task-2 readout ($\bm{a}_{2}$) alone is sufficient to interpolate $D_{2}$, requiring no change to the hidden-layer weights ($\mathcal{W}$). Due to the strong perturbation penalty ($\lambda\rightarrow\infty$), $\mathcal{W}$ do not change, maintaining the network representations
after learning task 1 (as derived in SI \ref{subsec:multiheadtheory}.\ref{paragraph:Hidden-layer-kernels}). This can also explain the divergence of $G_{2,2}$ as $\alpha\rightarrow1^-$: learning $D_{2}$ by modifying $\bm{a}_{2}$ on top of the $N$-dimensional fixed representations is effectively a linear regression, the generalization error of which is well known to diverge as $P$ approaches $N$ \citep{holzmuller2020universality}. This divergence is due to the divergence of the norm of $\bm{a}_2$ (Fig. \ref{fig5:ptdistractor}c). For smaller $\alpha$, the zero $F_{2,1}$ and $G_{2,2}$ remaining close to 1 demonstrate the advantage of using task-specific readouts. We term this regime of $\alpha<1$, where $F_{2,1}=0$ and $G_{2,2}$ is mostly finite, the ``fixed representations'' regime (FR).

As $\alpha$ increases past 1, interpolating $D_{2}$ requires changing $\mathcal{W}$. Consequently, we expected that such changes would induce forgetting of task 1. Surprisingly, we found that there exists
a critical load, $\alpha_{c}>1$, under which forgetting \emph{remains zero} (Fig. \ref{fig5:ptdistractor}d, top).
Further analysis showed that while changes in the network representations after learning task 1 no longer have zero norm, it is confined within the null space of $\bm{a}_{1}$ and thus does not alter the output on task 1 (SI \ref{subsec:multiheadtheory}.\ref{paragraph:Hidden-layer-kernels}). Although the absence of forgetting is desirable, this regime is accompanied by the network's inability to generalize on the second task, despite reaching zero training error. In fact, $G_{2,2}$ \emph{diverges} (Fig. \ref{fig5:ptdistractor}d, bottom), indicating the surprising phenomenon we term ``catastrophic anterograde interference'', where previous learning completely impedes generalization of new learning. We term this regime, where $F_{2,1}=0$, and $G_{2,2}\rightarrow\infty$, the ``overfitting'' regime. The divergence of $G_{2,2}$ in this regime is also due to the divergence of $\bm{a}_2$ (Fig. \ref{fig5:ptdistractor}e). In this regime, the minimal changes in the hidden layer weights $\mathcal{W}$ result in a hidden representation that does not learn the task rule of the second task (as we show later in Section 
\ref{sec:multi-head}.\ref{subsec:optimal-regularization}), causing $\bm{a}_2$ to diverge in order to compensate. As $\alpha$ further increases past $\alpha_{c}$, the network abruptly enters the ``generalization'' regime where $F_{2,1}>0$ and $G_{2,2}$ becomes finite. In this regime, the changes in the representation after learning task 1 are no longer confined to the null space of $\bm{a}_{1}$, inducing forgetting. The network partially forgets task 1, but learns to generalize on task 2.

Importantly, the boundary separating the two regimes ($\alpha_{c}$) depends on task relations. Different combinations of $(\rho_{\text{target}},\rho_{\text{shared}}, \rho_{\text{flip}})$ are associated with a different $\alpha_{c}$, separating the overfitting and generalization regimes. We found that under reasonable approximations, $\alpha_c$ is {\it fully} determined by a new OP measuring overall similarity of the two tasks, defined as 
\begin{equation} \label{eq:gammasim}
\gamma_{\text{sim}} = \gamma_{\text{feature}} + \cos(\bm{V}_1, \bm{V}_2) - \bm{V}_1^\top \bm{P}_2 \bm{V}_1/\lVert\bm{V}_1\rVert^2
\end{equation}
where $\bm{P}_i$ and $\bm{V}_i$ ($i\leq 2$) are defined as in Eqs. \ref{eq:projection}, \ref{eq:rule}. Although the precise definitions of the three terms in $\gamma_{\text{sim}}$ are different from the 3 OPs for single-head CL (introduced in Section \ref{sec:single-head}.\ref{subsec:task-relation OPs}) since we now focus on a different limit in a different CL architecture, they are actually closely related. The first term in $\gamma_{\text{sim}}$, $\gamma_{\text{feature}}$, is exactly the same as defined in Section \ref{sec:single-head}.\ref{subsec:task-relation OPs}. The second term, $\cos(\bm{V}_1, \bm{V}_2)$, measures the cosine similarity between the rule vectors of the two tasks; it has a similar interpretation as $\gamma_{\text{rule}}$. The third term, $\bm{V}_1^\top \bm{P}_2 \bm{V}_1/\lVert \bm{V}_1\rVert^2$, measures the projection of the rule vector onto the shared input feature subspace of the two tasks; it has a similar interpretation as $\gamma_{\text{RF}}$. $\gamma_{\text{sim}}$ is within the range of $[-1, 1]$. For conflicting tasks with the same input feature subspaces but opposite rule vectors, $\gamma_{\text{sim}} = -1$, for identical tasks $\gamma_{\text{sim}} = 1$. For dissimilar tasks with small overlap between input feature subspaces (small $\gamma_{\text{feature}}$), and rule vectors lying in the non-overlapping input feature subspaces of each task (small second and third terms in Eq. \ref{eq:gammasim}), $\gamma_{\text{sim}}$ is close to 0. As shown in Fig. \ref{fig5:ptdistractor}f, for $\alpha<1$, the network is in the ''fixed representation'' regime, independent of task relations. For $\alpha>1$, the phase transition $\alpha_c$ is accurately predicted by $\alpha_c = \gamma_{\text{sim}}^{-2}$ (black line) across different parameters of the target-distractor tasks (red points). $\alpha_c$ is larger for more dissimilar tasks with smaller $\gamma_{\text{sim}}$, resulting in a smaller generalization regime. For $\gamma_{\text{sim}}<0$, $\alpha_c=\infty$, and the network is always in the overfitting regime as long as $\alpha>1$.  

\subsection{Phase transitions in Benchmark Sequences}

\label{subsec:ptbenchmark}

\begin{figure}[H]
\centering
\includegraphics[width=0.8\linewidth]{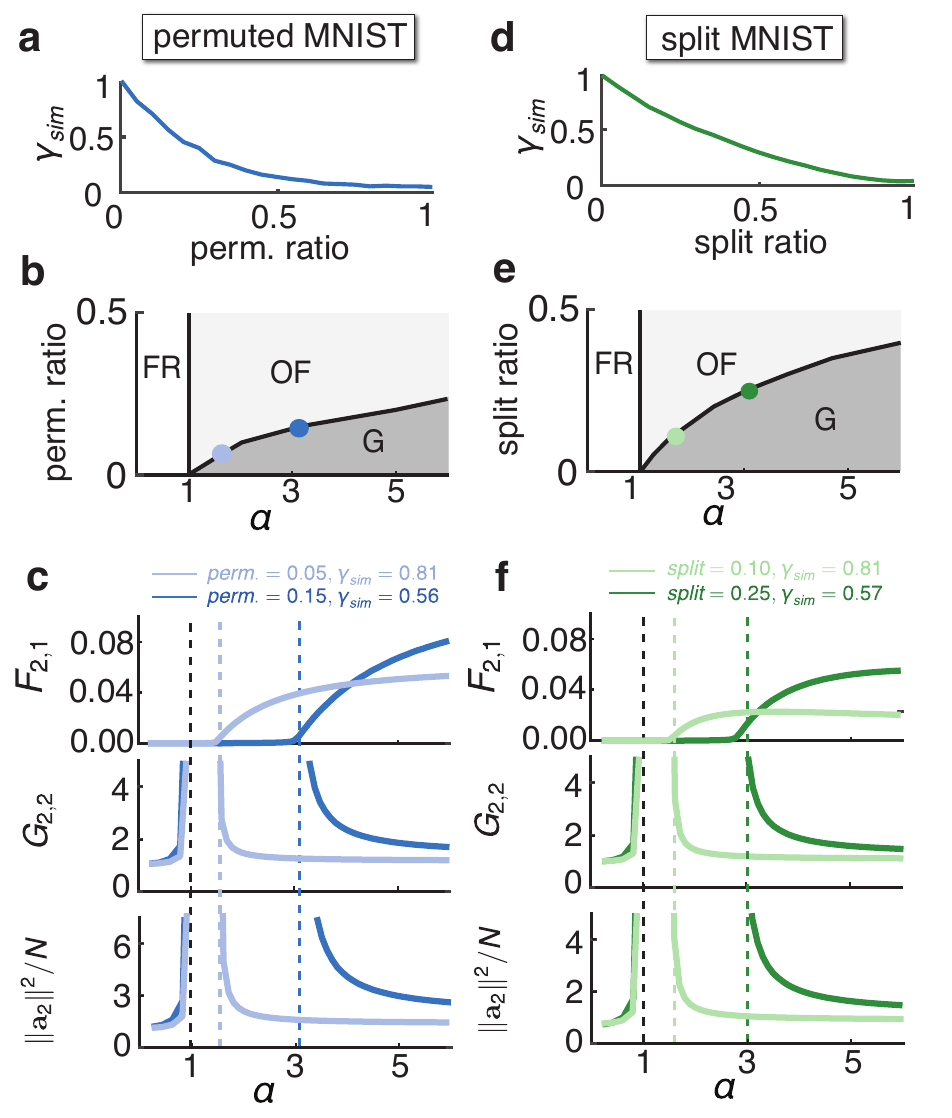}

\caption{\label{fig6:phase-transition-benchmark}\textbf{Similarity OP predicts phase transition in benchmark sequences.}\\
\textbf{a} For permuted MNIST with different permutation ratios, we can calculate the task similarity metric $\gamma_{\text{sim}}$, $\gamma_{\text{sim}}$ decreases with increasing permutation ratio. \\
\textbf{b} Using $\gamma_{\text{sim}}$ shown in a, the theory predicts a phase diagram in the permutation ratio-$\alpha$ space, showing the three regimes as in the target-distractor sequences: fixed representations (“FR”), overfitting (“OF”) and generalization (“G”). \\
\textbf{c} $F_{2,1}$, ${G}_{2,2}$ and $\lVert{\bf a}_{2}\rVert^{2}/N$ corresponding to the permutation ratios of the two points in b (light blue:perm. = $0.05$, $\gamma_{\text{sim}}=0.81$; dark blue: perm. = $0.15$, $\gamma_{\text{sim}}=0.56$), the transition from zero to positive $F_{2,1}$ and from diverging to finite ${G}_{2,2}$ or $\lVert{\bf a}_{2}\rVert^{2}/N$ is accurately predicted by the theoretical $\alpha_{c}$(light blue dashed line: $\alpha_c = 1.51$, dark blue dashed line: $\alpha_c=3.19$). \\
\textbf{d}-\textbf{f} Same as a-c, but for split MNIST, where we control the task similarity by changing the split ratio (see SI \ref{sec:benchmark}). $\gamma_{\text{sim}}$ decreases with increasing split ratio. The two examples shown in e,f correspond to split ratio $0.1$ (light green, $\gamma_{\text{sim}}=0.81$) and $0.25$ (dark green, $\gamma_{\text{sim}}=0.57$), and the theoretical prediction of their $\alpha_{c}$'s are $1.54$ and $3.03$ respectively.}
\end{figure}

Our results suggest that the three phases are in fact general phenomena, and the OP $\gamma_{\text{sim}}$ can be used to predict the phase transition boundary $\alpha_c$ across different task sequences. To verify, we computed the OP $\gamma_{\text{sim}}$ for two types of benchmark task sequences, permuted and split MNIST, where the task similarity is controlled by varying the permutation or split ratio (SI \ref{sec:benchmark}). Smaller permutation or split ratios correspond to intuitively more similar tasks and vice versa. As shown in Fig. \ref{fig6:phase-transition-benchmark}a, d, $\gamma_{\text{sim}}$ decreases with the permutation ratio or the split ratio, capturing the changes in the task similarity. When the permutation (split) ratio is 0, the tasks are identical and $\gamma_{\text{sim}}=1$, whereas when the permutation (split) ratio is 1, the tasks are dissimilar and $\gamma_{\text{sim}}=0$. 

Using $\gamma_{\text{sim}}$, our theory predicts an $\alpha_c$ for each permutation (split) ratio, producing a phase diagram in the permutation (split) ratio-$\alpha$ space (Fig. \ref{fig6:phase-transition-benchmark}b, e), with the same three regimes as we showed in the target-distractor sequences: the fixed representations regime (FR), the overfitting regime (OF), and the generalization regime (G). To verify the prediction of the phase diagram, we selected two examples for each type of task sequences with different permutation (split) ratios, and thus different $\gamma_{\text{sim}}$, and computed $F_{2,1}$, $G_{2,2}$ and $\lVert\bm{a}_2\rVert^2/N$ as a function of $\alpha$. As shown in Fig. \ref{fig6:phase-transition-benchmark}c, f, the theoretical prediction of $\alpha_c$ (dashed lines) accurately captures the abrupt changes in the performance of the network: for $\alpha<1$, $F_{2,1}=0$, $G_{2,2}$ and $\lVert\bm{a}_2\rVert^2/N$ are finite and start to diverge as $\alpha\rightarrow 1^-$; for $1<\alpha<\alpha_c$, $F_{2,1}$ remains 0, $G_{2,2}$ and $\lVert\bm{a}_2\rVert^2/N$ are diverging; for $\alpha>\alpha_c$, $F_{2,1}$, $G_{2,2}$ and $\lVert\bm{a}_2\rVert^2/N$ are all finite and nonzero. Finally, these qualitative behaviors of phase transitions were reproduced in gradient-descent trained networks (Fig. \ref{fig:gdmulti})
as well as in CL of longer task sequences (Fig. \ref{fig:taskinterpolation}).

\subsection{Balancing Memorization and New Learning with Finite $\lambda$}

\label{subsec:optimal-regularization}

\begin{figure}[H]
\centering
\includegraphics[width=1\linewidth]{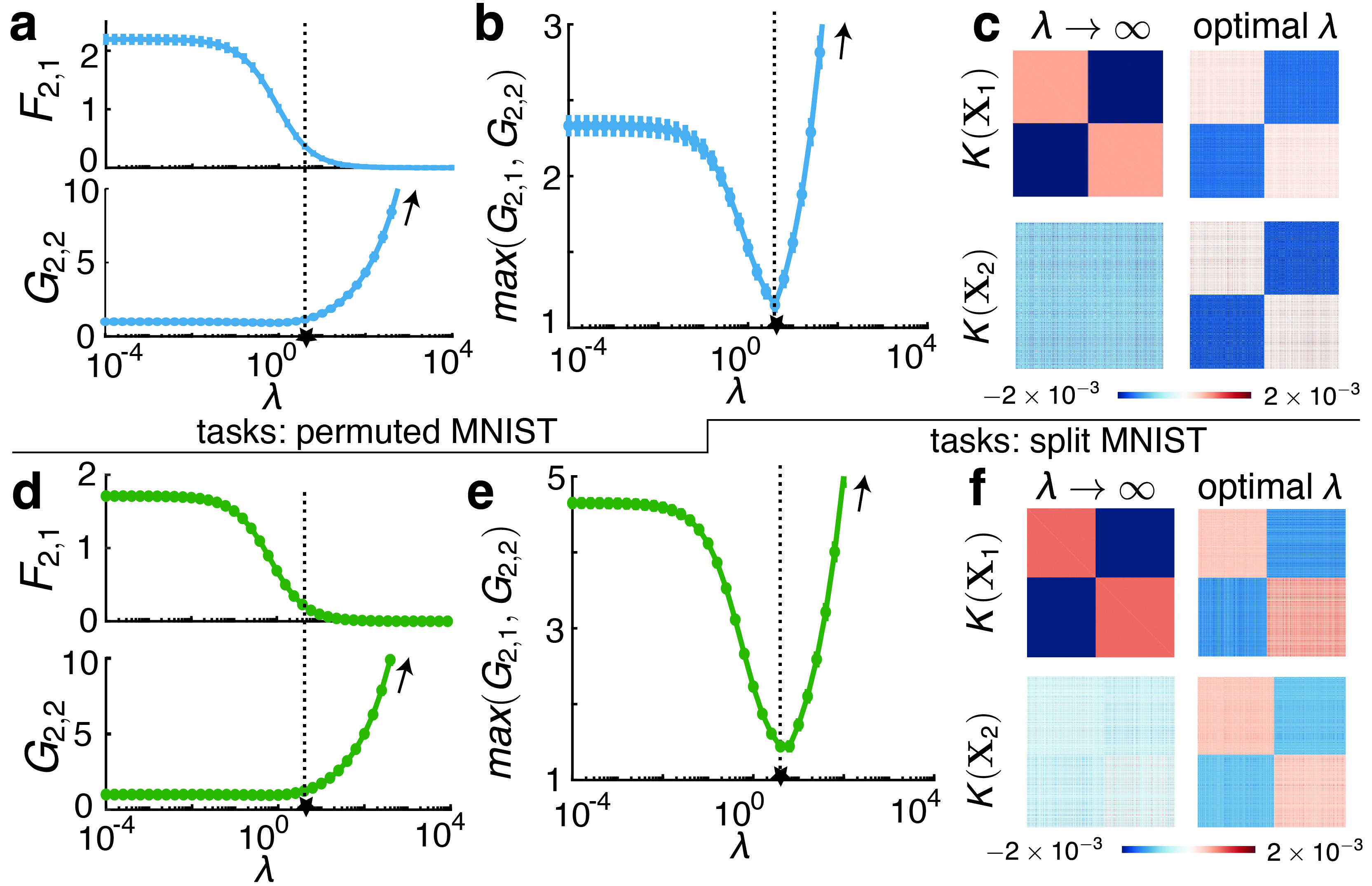}
\caption{\protect\label{fig7:optimal-lambda}\textbf{Optimal regularization strength balances memorization and learning new tasks.}\\
\textbf{a} Forgetting of task 1 ($F_{2,1}$) monotonically decreases with the regularization strength ($\lambda$) while the normalized generalization error on task 2 (${G}_{2,2}$) monotonically increases. The two tasks considered here are sufficiently dissimilar (permuted MNIST with 100\% permutation ratio) that they are in the overfitting regime and ${G}_{2,2}$ diverges at large $\lambda$.  \\
\textbf{b} Maximum of the normalized generalization error on task 1 and task 2 ($\max({G}_{2,1},{G}_{2,2})$) as a function of the regularization strength $\lambda$. There exists an intermediate optimal $\lambda$ which minimizes $\max({G}_{2,1},{G}_{2,2})$ by keeping both of them close to 1 such that there's minimal interference coming from the other task, indicated by the star. \\
\textbf{c} The learned component of the similarity matrix of the hidden layer activations on task 1 and task 2 training data ($\bm{X}_{1}$ and $\bm{X}_{2}$), after learning the two tasks, denoted $K(\bm{X}_{1})$ (top row) and $K(\bm{X}_{2})$ (bottom row) respectively. At $\lambda\rightarrow\infty$, only $K(\bm{X}_{1})$ but not $K(\bm{X}_{2})$ exhibits a task-relevant block structure. This indicates that the network fails to learn good representations for task 2, and over-memorizes task 1. In contrast, at the optimal $\lambda$ (corresponding to the star in b), both $K(\bm{X}_{1})$ and $K(\bm{X}_{2})$ show a block structure aligned with their corresponding tasks, exhibiting a shared representation beneficial for both tasks. \\
\textbf{d-f} Same as a-c, but for split MNIST with 100\% split. \\
In a, b, d, e, errorbars are standard deviations across 10 different random seeds of task sequence generation. $\alpha=3$ in both examples, which is below the corresponding $\alpha_{c}$ for these sequences.}
\end{figure}

The analysis so far has shown that, when tasks are sufficiently dissimilar,
$\lambda\rightarrow\infty$ can cause the network to memorize perfectly
(zero $F_{2,1}$) at the expense of catastrophic anterograde interference
(diverging $G_{2,2}$). We next characterized the trade-off for such
dissimilar tasks between improving $G_{2,2}$ and maintaining low $F_{2,1}$
by lowering $\lambda$. As expected, as $\lambda$ lowers, the network forgets the first task more (higher $F_{2,1}$, Fig. \ref{fig7:optimal-lambda}a, d, top), resulting in weaker interference of the second task (lower $G_{2,2}$, Fig. \ref{fig7:optimal-lambda}a, d bottom). To evaluate and compare the performance on both tasks and quantify the trade-off, we also computed the \emph{normalized test loss} on task 1 after learning task 2, denoted by $G_{2,1}$, and studied $\max(G_{2,1},G_{2,2})$ as a function of $\lambda$ (Fig. \ref{fig7:optimal-lambda}b, e). We found that there exists a finite optimal $\lambda$ that minimizes $\max(G_{2,1},G_{2,2})$ by keeping \textit{both} $G_{2,1}$ and $G_{2,2}$ close to 1. 

We next sought to understand how the representations of task 1 and task 2 inputs depend on $\lambda$ by studying the representation similarity matrix after learning \emph{both tasks}. Specifically, we analyzed the learned component in the representation similarity matrix on the
training data $\bm{X}_{1}$ and $\bm{X}_{2}$ (SI \ref{subsec:multiheadtheory}.\ref{paragraph:Hidden-layer-kernels}), denoted by
$K(\bm{X}_{1})\in\mathbb{R}^{P\times P}$ and $K(\bm{X}_{2})\in\mathbb{R}^{P\times P}$ respectively. Prior work has indicated that, for binary classification tasks that we considered, a $2\times2$ block structure in the similarity matrix suggests that the representations are clustered according to
the task labels, and is associated with good generalization performance \citep{baratin2021implicit,shan2021theory,bordelon2022self,atanasov2021neural}.
Indeed, we found that at large $\lambda$, the similarity matrix has such $2\times2$ block structure for $\bm{X}_{1}$ but not $\bm{X}_{2}$ (Fig. $\text{\ref{fig7:optimal-lambda}}$c, f), explaining our previous finding that the network fails to generalize on task 2 in the overfitting regime. However, when using the optimal $\lambda$ that minimizes $\max\left(G_{2,1},G_{2,2}\right)$, representations of inputs from both tasks have such block structure, consistent with the finding that both $G_{2,1}$ and $G_{2,2}$ are close to 1, highlighting the importance of representation learning on the generalization
capabilities in CL.

\section{Discussion }

\paragraph{Related Works}

Many mechanisms have been proposed for mitigating catastrophic forgetting (for a recent review, see \cite{wang2023comprehensive}). We used the simplest approach by adding an $L_2$ penalty on weight changes to facilitate theoretical analysis. While this work's aim is not to achieve state-of-the-art performance but to develop a theoretical understanding of CL, it is important that our theoretical results achieve reasonable performance compared to commonly used CL approaches. We confirmed this by comparing against networks trained using gradient descent with online EWC \cite{schwarz2018progress} or L2 regularizers (SI \ref{sec:gdsimulations}). 


Theoretical advances on CL have been made recently. For single-head CL, our theoretical results (SI Eqs.\ref{SIeq:meanpredictor}, \ref{SIeq:v} in SI \ref{subsec:singleheadtheory}.\ref{subsubsec:summarysinglehead}) are consistent with \citep{doan2021theoretical, bennani2020generalisation} in the $\lambda\rightarrow\infty$ limit. However, we stress that our theoretical framework and assumptions are different.  While \citep{doan2021theoretical,bennani2020generalisation} assume linearization of the dynamics around initialization in the NTK regime \citep{jacot2018neural}, our formulation does not rely on any assumptions on the learning dynamics. Interestingly, we recover an NTK-like theory of CL in the single-head scenario when $\lambda\rightarrow\infty$. Furthermore, while these previous results proposed an analytical expression for the mean predictor, they did not provide explicit predictions of the theory on CL performance. In this work, combining our theoretical solutions and extensive numerical evaluations, we elucidated the connections between forgetting in CL and task relations using the theoretically inspired OPs. Importantly, our Gibbs formulation also allows us to investigate the effect of $\lambda$ (Fig. \ref{fig:dependencelambda}), which is not amenable in the NTK formulation in \citep{doan2021theoretical, bennani2020generalisation}. For multi-head CL, most theoretical works make specific assumptions about the tasks \citep{lee2021continual, gerace2022probing, lee2022maslow}, limiting their applicability. We became aware of a very recent work during the preparation of our manuscript \citep{ingrosso2024statistical}, which adopts a Gibbs formulation for transfer learning similar to our work and does not make specific assumptions about the statistics of the tasks. However, this work focused on learning in the regime $\alpha<1$ with finite $\lambda$, and therefore did not uncover the intriguing phenomenon of phase transitions that are determined by task similarities as in our work.

\paragraph{Forgetting and Task Relation OPs}

We systematically investigated how task relations influence forgetting in $L_{2}$-regularized CL in wide DNNs in single-head
and multi-head scenarios, studying both short-term forgetting (sequential
learning of two tasks) and long-term forgetting (a long sequence of
tasks). In contrast to prior work, which mostly treated ``task similarity'' as a single variable \citep{goodfellow2013empirical,doan2021theoretical,ramasesh2020anatomy,saglietti2022analytical,lee2022maslow,gerace2022probing}
(but see \citep{lee2021continual,asanuma2021statistical}), our analysis emphasizes the importance of distinguishing different aspects of task relations on CL. Importantly, we identified several scalar OPs quantifying these task relations. These OPs can be evaluated given the training data of each task, and are highly predictive of CL performance. We summarize the definitions of these OPs in Table \ref{Table1:OPsummary}. 

\begin{table}
\caption{Definitions of the task relation OPs.}
\centering
\begin{tabular}{ |c|c|c| } 
\hline
Scenario & OP & Definition \\
\hline
\multirow{3}{6em}{Single-Head} 
& $\gamma_{\text{feature}}$ & $P^{-1}\mathrm{Tr}(\bm{P}_1\bm{P}_2)$ \\ 
& $\gamma_{\text{RF}}$ & $\bm{V}_2^\top \mathcal{P}_{12}\bm{V}_2$ \\ 
& $\gamma_{\text{rule}}$ &  $\bm{V}_2^\top \mathcal{P}_{12}\bm{V}_1$\\ 
\hline
Multi-Head & $\gamma_{\text{sim}}$ & $\gamma_{\text{feature}} + \cos(\bm{V}_1, \bm{V}_2) - \bm{V}_1^\top \bm{P}_2 \bm{V}_1/\lVert \bm{V}_1\rVert^2$ \\
\hline
\end{tabular}
\label{Table1:OPsummary}
\end{table}

For single-head CL, we studied $\gamma_{\text{feature}}$, $\gamma_{\text{RF}}$ and $\gamma_{\text{rule}}$. Interestingly, $\gamma_{\text{feature}}$ is only weakly related to forgetting, while the other two OPs play important roles. This suggests that the similarity between the input features, which are \emph{relevant for the task}, rather than the input features themselves, are the determining factors of forgetting. $\gamma_{\text{RF}}-\gamma_{\text{rule}}$, which we termed the conflict between tasks, is directly related to short-term forgetting. For tasks with low conflict and thus small short-term forgetting, long-term forgetting accumulates slowly for dissimilar tasks (small $\gamma_{\text{RF}}, \gamma_{\text{rule}}$) and quickly for similar tasks (high $\gamma_{\text{RF}}, \gamma_{\text{rule}}$). For multi-head CL, our theory identifies another task similarity OP $\gamma_{\text{sim}}$ composed of 3 terms, each bearing resemblance to the 3 OPs in single-head CL including $\gamma_{\text{feature}}$, which does not significantly affect single-head CL performance. The effect of this OP depends also on the load $\alpha$. For $\alpha<1$, task relations have no effect on forgetting as it vanishes at $\lambda\rightarrow\infty$. However, for $\alpha>1$ there exists a $\gamma_{\text{sim}}-\alpha$ phase diagram (Fig. \ref{fig5:ptdistractor}d).
For a fixed load ($\alpha$), when $\gamma_{\text{sim}}$ are high, CL is in the generalization regime where forgetting is non-zero but moderate. When the tasks become sufficiently dissimilar, CL abruptly enters the overfitting regime where forgetting
is zero but generalization on the new task fails despite reaching zero training error, a surprising phenomenon we termed ``catastrophic anterograde interference.'' For tasks in this regime, fine-tuning $\lambda$ of the learner can reach a reasonable compromise and allow
the network to perform both tasks (Fig. \ref{fig7:optimal-lambda}). 

\paragraph{Architecture}

Our analysis suggests that task relations are modulated by the architecture
of the learner. Increasing depth effectively mitigates single-head forgetting for short task sequences (decreased $F_{2,1}$, Fig. \ref{fig4:OPsandbenchmark}a)
by reducing the conflict between tasks, but has a more complicated effect on long-term forgetting (reflected in non-monotonic $F_{\text{max}}$, Fig. \ref{fig4:OPsandbenchmark}c) due to the opposing effects it has on $F_{2,1}$ and $\tau_F$. In addition, increasing the width ($N$), which we studied for multi-head
CL, can also mitigate forgetting. As $N$ increases for a fixed dataset
size ($P$), $\alpha$ decreases below $\alpha_{c}$. As a result,
CL transitions from the generalization regime, where forgetting is
finite, to the overfitting regime, where it is zero. Although the
specific value of $\alpha_{c}$ depends on task relations, our theory
indicates that the transition to zero forgetting is a general phenomenon.
Widening the network further eventually causes $\alpha$ to drop below
1 where network features are fixed and forgetting is zero for any
tasks. The observed beneficial effects of depth and width on \emph{mitigating
forgetting} are consistent with empirical reports of less forgetting
in larger networks \citep{ramasesh2021effect}.

\paragraph{Anterograde Effects}

In addition to forgetting (retrograde interference), we investigated
anterograde aspects of CL by studying how learning one task affects
the generalization performance on a subsequently learned one. For single-head CL, we omitted discussion of anterograde effects from the main text as they are generally weak in the \emph{infinite width limit} that we consider (Fig. \ref{fig:dependencep}), consistent with previous reports. However, results from multi-head (Figs. \ref{fig5:ptdistractor}-\ref{fig7:optimal-lambda}) CL indicate that anterograde interference can be severe and worsens as the tasks become less similar.
This suggests a parameter regime at $\alpha>1$ where, counter-intuitively,
single-head CL performs better than multi-head (Fig. \ref{fig:singlevsmulti}). It would be interesting to rigorously verify this in a future theory of single-head CL with finite $\alpha$. The existence of diverging test loss for $\alpha>1$ suggests that increasing
the width of the network (reducing $\alpha$ to a value below $1$) will have a very beneficial effect on sequential learning. While anterograde interference appears prevalent and severe in our analysis, this is
partially due to the specific settings we focused on. Assuming the second task to have substantially fewer training examples than the first or a compositional structure between tasks \citep{lee2024animals} could lead to stronger positive transfer effects. In addition, making transitions between dissimilar tasks ``smoother'' by inserting intermediate datasets can mitigate anterograde interference in multi-head CL (Fig. \ref{fig:taskinterpolation}).

\paragraph{Implications for CL in the Brain}

Recent neuroscience experiments indicate that neural representations of a learned task can ``drift'' after learning has concluded \citep{rule2019causes,driscoll2022representational}, raising the question of how the brain maintains stable task performance
despite such drifts \citep{driscoll2017dynamic}. While a multitude of mechanisms likely underlie this phenomenon, subsequent learning of other tasks by the same neural circuits likely contributes \citep{driscoll2022representational}. As shown by our analysis, this can indeed occur during multi-head CL at $\alpha>1$, where representations of task 1 inputs are altered
by learning the second task. Our analysis hints at how the brain may deal with this issue. Task 1 performance can be unperturbed as long as representational changes occur only in the null space of its readout, consistent with the notion that the brain orthogonalizes representations for different tasks to reduce interference \citep{duncker2020organizing,flesch2022orthogonal,driscoll2022representational}. The overfitting regime demonstrates that such orthogonality can occur
without storing task 1 inputs and explicitly confining new learning in their null space, as long as the penalty on weight perturbations is sufficiently strong. To avoid the failure to generalize on task
2 in this regime, the brain may weaken the penalty or enforce a hard constraint on the strength of the readout weights, such that representational changes are still mostly orthogonal to the task 1 readout but sufficient for good generalization of task 2. These results suggest the possibility of enforcing near-orthogonality between task subspaces by having a regularization-like mechanism (e.g., synaptic stabilization \citep{xu2009rapid,yang2009stably}) alone with appropriately tuned penalty or constrained readout weight strength.

Our results also highlight how architectural elements of the brain can confer CL benefits. Sensory expansion, a motif often seen in sensory cortices, projects a low-dimensional input signal into a much higher-dimensional code within a large population of neurons \citep{babadi2014sparseness}.
From the perspective of multi-head CL, this may effectively increase the NN width and reduce forgetting, as discussed above. Additionally, our finding that increasing depth can mitigate forgetting may indicate
an advantage of having a deep, multi-stage sensory processing system. This suggestion predicts that representations of different tasks are less similar in later stages of sensory processing \citep{ito2023multitask,johnston2023abstract}.
To assess such similarity in high-dimensional neural codes without
resorting to nonlinear dimensionality-reduction techniques (e.g.,
\citep{flesch2022orthogonal}), it may be promising to adapt our
OPs to experimental data.

Finally, it would be interesting to test whether the same connections between task relations and severity of forgetting hold in the brain. For instance, animals can be sequentially trained on a series of two-alternative forced choice tasks. In each task, the animal would need to distinguish
two classes of simple stimuli with a few attributes, much like the example shown in Fig.\ref{fig:Schematics}a. Different tasks would contain
different dichotomies on different attributes. The dichotomies should be consistent across tasks to ensure a small conflict $\gamma_{\text{RF}} - \gamma_{\text{rule}}$.  Assuming animals are using a single-head-like shared behavioral readout for these tasks \citep{ni2018learning} and a regularization-like mechanism for CL, our results predict forgetting will accumulate longer if attributes from different tasks are made more distinct (lower $\gamma_{\text{RF}}$ and $\gamma_{\text{rule}}$).

\paragraph{Extensions and Limitations}

The presented theory can be extended in several important directions. First, our Gibbs formulation assumes a uniform perturbation penalty across all weights, while popular regularization-based CL methods
typically use some metric to evaluate the importance of each individual weight for past performance and apply a stronger penalty to more important ones \citep{kirkpatrick2017overcoming,zenke2017continual}. Our theory may be extended to the case with weight-specific penalties and elaborate on how different importance metrics affect CL outcomes.

Second, we assumed that the tasks are symmetric, and have pairwise similar task relations in long task sequences, which prevents us from capturing how
different orderings of the same set of tasks elicit different CL performance.
While we have neglected ordering effects here because they are often
small in common task sequences \citep{bell2022effect,saglietti2022analytical}, it remains an interesting future direction to systematically probe into the effect of task ordering in more structured progressive task sequences. 

Finally, while we have focused on leveraging task-identity information during CL using the multi-head scheme, having multiple readouts is considered less realistic for CL in the brain. Conceptually, the task identity information can be incorporated through gating units that gate part of the readout weights on or off depending on the task, as in \citep{li2022globally}, making the multi-head scheme more biologically relevant. Task-identity information can be incorporated into single-head CL by appending a task-identity embedding vector to relevant inputs \citep{niv2019learning} or gating individual neurons in a task-dependent manner \citep{abati2020conditional,mirzadeh2020dropout,flesch2023modelling}. In Fig. \ref{fig:singlecontext}, preliminary results show that while appending task-identity embedding vectors to the inputs helps mitigate forgetting, its beneficial effect is still weaker compared to adding task-dependent readouts (multi-head CL). Extending our theory to other mechanisms of incorporating task-identity information such as gating and studying how they affect the OPs and CL performance is a promising future research direction.

\subsection*{Acknowledgement}
The authors would like to thank Alexander van Meegen and Daniel D.
Lee for helpful discussions. This research is supported by the Swartz
Foundation, the Gatsby Charitable Foundation, the Kempner Institute
for the Study of Natural and Artificial Intelligence at Harvard University, the grant NSF PHY-2309135 to the Kavli Institute for Theoretical Physics (KITP)
and Office of Naval Research (ONR) grant No. N0014-23-1-2051.


\newpage
\bibliographystyle{unsrtnat}  
\bibliography{pnas-sample}

\begin{center}
\section*{Order parameters and phase transitions of continual learning in deep
neural networks: Supplementary information}
\author{Haozhe Shan, Qianyi Li and Haim Sompolinsky}
\end{center}



%
%



\maketitle


\part{Theoretical Results}
\label{sec:theory}
In this section, we will present summary and derivations of our theoretical results. We begin by introducing more formally the network architecture (SI \ref{subsec:architecture}) and defining the important kernel functions which will show up repeatedly in our theoretical results (SI \ref{subsec:kernels}). We then present single-head CL and multi-head CL theories separately, first summarizing the main theoretical results (SI \ref{subsec:singleheadtheory}.\ref{subsubsec:summarysinglehead} and SI \ref{subsec:multiheadtheory}.\ref{subsubsec:summarymultihead}), and then presenting the detailed derivations (SI \ref{subsec:singleheadtheory}.\ref{subsubsec:derivationsingle} and SI \ref{subsec:multiheadtheory}.\ref{subsubsec:derivationmulti}). 

\section{Network Architecture}
\label{subsec:architecture}
All networks we studied have a fully-connected feedforward body. $\bm{x}^{l=0,...,L}$
denote the vector of activation in the $l$-th hidden layer in response
to input $\bm{x}$ ($\bm{x}^{0}\equiv\bm{x}$), given by
\begin{equation}
x_{i}^{l+1}\left(\bm{x}\right)=\phi\left(\frac{1}{\sqrt{\text{dim}\left(\bm{x}^{l}\right)}}\sum_{j}W_{ij}^{l}x_{j}^{l}\left(\bm{x}\right)\right),
\end{equation}
where $\text{dim}\left(\bm{x}^{0}\right)=N_{0}$ and $\dim\left(\bm{x}^{1,...,L}\right)=N$.
The representation of an input $\bm{x}$ denotes the last-layer activation,
$\Phi\left(\mathcal{W},\bm{x}\right)\equiv\bm{x}^{L}\left(\bm{x}\right)$.
$\phi:\mathbb{R}\rightarrow\mathbb{R}$ is the single-neuron activation
function, taken to be ReLU ($\phi(x)=\max\left\{ 0,x\right\} $).

\section{Generalized Kernel Functions}
\label{subsec:kernels}
Our theoretical results showed that the statistics of the network's input-output mappings depend on the input data through several generalized kernel functions, similar in spirit to the Neural Tangent Kernel (NTK) and Neural Network Gaussian Process (NNGP) theories of learning \citep{jacot2018neural,lee2017deep}. We will introduce these kernel functions and the notations we will use throughout in this section, and the main theoretical results in SI \ref{subsec:singleheadtheory} and SI \ref{subsec:multiheadtheory}. 

There are two crucial differences between the generalized kernel functions in our theory and the kernels in NTK/NNGP theories. First, our kernel functions are ``generalized'' in that they are
generally asymmetric with respect to the inputs and thus are not proper
kernels \citep{hofmann2008kernel}. Second, our kernel functions are
time-dependent, as opposed to the stationary NTK/NNGP kernels in classic
results. For brevity, we simply refer to the generalized kernel functions
as kernels or kernel functions hereafter.

The kernel functions are defined as the inner products between random
features averaged over correlated Gaussian weights. The statistics
of the Gaussian weights are given by the prior contribution 

\begin{equation}
P(\mathcal{W})\propto\exp\left(-\frac{1}{2}\beta^{-1}\sigma^{-2}\sum_{t=1}^{T}\left\Vert \mathcal{W}_{t}\right\Vert ^{2}+\frac{1}{2}\beta^{-1}\sum_{t=2}^{T}\lambda\left\Vert \mathcal{W}_{t}-\mathcal{W}_{t-1}\right\Vert ^{2}\right)\label{eq:priorw}
\end{equation}

\subsection{Kernel Functions for Arbitrary $\lambda$}

The important kernel functions are given by 
\begin{equation}
K_{t,t'}^{L,1}(\bm{x},\bm{x}')\equiv\frac{1}{N}\langle\Phi(\mathcal{W}_{t},\bm{x})\cdot\Phi(\mathcal{W}_{t'},\bm{x}')\rangle_{\mathcal{W}}\label{eq:kgp1-1}
\end{equation}
\begin{equation}
K_{t,t'}^{L,0}(\bm{x},\bm{x}')\equiv\frac{1}{N}\langle\langle\Phi(\mathcal{W}_{t},\bm{x})\rangle_{t,t'}\cdot\langle\Phi(\mathcal{W}_{t'},\bm{x}')\rangle_{t',t'}\rangle_{\mathcal{W}}\label{eq:kgp0-1}
\end{equation}
where $\langle\cdot\rangle_{\mathcal{W}}$ denotes the average over
the full prior distribution SI Eq. \ref{eq:priorw}, and $\langle\cdot\rangle_{t,t'}$
denotes the partial average over the conditional distribution $P(\mathcal{W}_{t,}\mathcal{W}_{t-1},\cdots,\mathcal{W}_{t'}|\mathcal{W}_{t'-1})$.
Furthermore, we introduced
\begin{equation}
\tilde{K}_{t,t'}^{L}(\bm{x},\bm{x}')\equiv m_{t,t'}^{1}K_{t,t'}^{L,1}(\bm{x},\bm{x}')-m_{t,t'}^{0}K_{t,t'}^{L,0}(\bm{x},\bm{x}')\label{eq:ktilde}
\end{equation}
For $t\geq t'\geq2$, $m_{t,t'}^{1}=\frac{\sigma^{2}}{1+\tilde{\lambda}}(\tilde{\lambda}^{t-t'}+\tilde{\lambda}^{t+t'-1})$,
and $m_{\tau,\tau^{\prime}}^{0}=\frac{\sigma^{2}}{1+\tilde{\lambda}}(\tilde{\lambda}^{t-t'+2}+\tilde{\lambda}^{t+t'-1})$,
where $\tilde{\lambda}\equiv\lambda(\lambda+\sigma^{-2})^{-1}$. Otherwise,
$m_{t,1}^{1}=\sigma^{2}\tilde{\lambda}^{t-1}$ and $m_{t,1}^{0}=0$.
Finally, we introduced the difference kernel
\begin{equation}
\Delta K_{t,t'}^{L}(\bm{x},\bm{x}')\equiv K_{t,t'}^{1}(\bm{x},\bm{x}')-K_{t,t'}^{0}(\bm{x},\bm{x}').
\end{equation}

\subsection{Kernel Functions in the $\lambda\rightarrow\infty$ Limit:}

In the limit $\lambda\rightarrow\infty$, these kernel functions become
stationary in time and can be simplified. In particular, $K_{t,t'}^{L,1}(\bm{x},\bm{x}')$
becomes the NNGP kernel, given by 
\begin{equation}
K_{t,t'}^{L,1}(\bm{x},\bm{x}')\overset{\lambda\rightarrow\infty}{\longrightarrow}K_{GP}^{L}(\bm{x},\bm{x}')\equiv\Phi(\mathcal{W}_{0},\bm{x})\cdot\Phi(\mathcal{W}_{0},\bm{x}'),\label{eq:gp}
\end{equation}
where $\Phi(\mathcal{W}_{0},\bm{x})\in\mathbb{R}^{N}$, and $\mathcal{W}_{0}\sim\mathcal{N}(0,\sigma^{2}\mathbb{I})$,
as introduced in the main text when we introduced the OPs. $\tilde{K}_{t,t'}^{L}$
($t,t'\geq2$) becomes the NTK, given by 
\begin{equation}
\tilde{K}_{t,t'}^{L}(\bm{x},\bm{x}')\overset{\lambda\rightarrow\infty}{\longrightarrow}\lambda^{-1}K_{NTK}^{L}(\bm{x},\bm{x}')\equiv\lambda^{-1}\partial_{\Theta_{random}}f(\Theta_{0},\bm{x})\cdot\partial_{\Theta_{random}}f(\Theta_{0},\bm{x}')\label{eq:ntk}
\end{equation}
where $\partial_{\Theta_{0}}f(\Theta_{0},\bm{x})\in\mathbb{R}^{N^{2}(L-1)+NN_{0}+N}$
is of the dimension of the total number of parameters in the network,
and $\Theta_{0}\sim\mathcal{N}(0,\sigma^{2}\mathbb{I}).$ $\Delta K_{t,t'}^{L}(\bm{x},\bm{x}')$
is given by
\begin{equation}
\Delta K_{t,t'}^{L}(\bm{x},\bm{x}')\overset{\lambda\rightarrow\infty}{\longrightarrow}\lambda^{-1}K_{NTK}^{L}(\bm{x},\bm{x}')\dot{K}^{L}(\bm{x},\bm{x}')\label{eq:deltak}
\end{equation}
where 
\begin{equation}
\dot{K}^{L}(\bm{x},\bm{x}')\equiv\phi'(h_{0}^{L}(\bm{x}))\cdot\phi'(h_{0}^{L}(\bm{x}'))\label{eq:dotk}
\end{equation}
 is the derivative kernel. $\left\{ h_{0}^{l}(\bm{x})\right\} _{l=1,\cdots,L}$
denote the pre-activation of each layer with random weights $\mathcal{W}_{0}=\left\{ W_{0}^{l}\right\} _{l=1,\cdots,L}$,
i.e.,
\begin{equation}
h_{0}^{l}(\bm{x})\equiv\frac{1}{\sqrt{\text{dim}\left(\bm{x}^{l-1}\right)}}\sum_{j}W_{0,ij}^{l-1}x_{0,j}^{l-1}\left(\bm{x}\right)
\end{equation}
\begin{equation}
x_{0}^{l}(\bm{x})=\phi(h_{0}^{l}(\bm{x}))
\end{equation}
In the infinite-width limit $N\rightarrow\infty$, SI Eqs. \ref{eq:gp}-\ref{eq:dotk}
are all self-averaging, namely, they do not depend on the specific
realization of $\mathcal{W}_{0}$ or $\Theta_{0}$, and is equivalent
to their averages across Gaussian $\mathcal{W}_{0}$ or $\Theta_{0}$.
See SI \ref{subsec:singleheadtheory}.\ref{paragraph:kernelfunctions} for detailed derivations of the kernel functions and SI \ref{subsec:singleheadtheory}.\ref{paragraph:kernelfunctionexpressions} for their analytical expressions in linear and ReLU networks.

Furthermore, to simplify the expression of the statistics of the network's
input-output mappings, for each kernel function, we introduce corresponding
notations for applying them to the training and testing data, respectively.
Specifically, for a kernel function $K_{t,t'}^{L}(\bm{x},\bm{x}')$,
we introduce 
\begin{equation}
\bm{k}_{t,t'}^{L}(\bm{x})\equiv K_{t,t'}^{L}(\bm{x},\bm{X}_{t'})\in\mathbb{R}^{P}
\end{equation}
\begin{equation}
\bm{K}_{t,t'}^{L}\equiv K_{t,t'}^{L}(\bm{X}_{t},\bm{X}_{t'})\in\mathbb{R}^{P\times P}
\end{equation}
\begin{equation}
k_{t,t'}^{L}\equiv K_{t,t'}^{L}(\bm{x},\bm{x})\in\mathbb{R}
\end{equation}
where $\bm{X}_{t}$ denotes the training data matrix of task $t$,
and $\bm{x}$ denotes an arbitrary test point. $K^{L}$ represents
the different kernel functions above, including $K^{L,1},K^{L,0},\tilde{K}^{L}$
and $\Delta K^{L}$.

\section{Single-Head Theory}
\label{subsec:singleheadtheory}

\subsection{Summary of Main Theoretical Results}
\label{subsubsec:summarysinglehead}

\subsubsection{Statistics of the Input-Output Mappings}
\label{paragraph:statisticssingle}
In single-head CL, the mean input-output mapping after learning $T$
tasks in a network with $L$ hidden layers in the infinite-width limit
($N\rightarrow\infty$, $\alpha=P/N\rightarrow0$) is given by 
\begin{equation}\label{SIeq:meanpredictor}
\langle f_{T}(\bm{x})\rangle=\sum_{t=1}^{T}\tilde{\bm{k}}_{T,t}^{L}(\bm{x})^{\top}\langle-i\bm{v}_{t}\rangle
\end{equation}
\begin{equation}\label{SIeq:v}
\langle-i\bm{v}_{t}\rangle=\left(\tilde{\bm{K}}_{t,t}^{L}\right)^{-1}(\bm{Y}_{t}-\sum_{t'=1}^{t-1}\tilde{\bm{K}}_{t,t'}^{L}\langle-i\bm{v}_{\tau'}\rangle)
\end{equation}

The equation is applied to evaluate $F_{t,t'}\approx\mathcal{L}(\langle f_{T}\rangle,D_{t'})$ and $G_{t,t'}\approx\mathcal{L}(\langle f_{t}\rangle,D_{t'}^{test})/G_{t'}^0$. For simplicity, we assumed small $\sigma$ such that the variance contribution to $\langle\mathcal{L}(f,D)\rangle$ is neglected, but the expression for the variance of $f_T(\bm{x})$ is given in SI \ref{subsec:singleheadtheory}.\ref{paragraph:predictor-statistics}. These results hold for general $\lambda$, in the $\lambda\rightarrow\infty$
limit, the kernels $\tilde{K}_{t,t'}^{L}(t,t'\geq2)$ are replaced with $\lambda^{-1}K_{NTK}^{L}$ (SI Eq. \ref{eq:ntk}) and the kernels $\tilde{K}_{t,1}^{L}$ are replaced with $K_{GP}^{L}$ (SI Eq. \ref{eq:gp}). 

Although $\mathcal{W}_t$ is still time-dependent in the $\lambda\rightarrow\infty$ limit, empirically, we found that the change in $\mathcal{W}_t$ over time is weak, and we can assume that $\mathcal{W}_t \approx \mathcal W_0 \sim \mathcal{N}(0,\sigma^2)$ for all $t$. This results in a simpler dynamical process where only the readout weights $\bm{a}_t$ change over time, learning on top of the random feature $\Phi(\mathcal{W}_0, {\bf x})$. In this case the predictor statistics are simplified, such that all $\tilde{K}_{t,t'}^L$ can be replaced by $K_{GP}^L$. We refer to this approximation as the random feature approximation. In Fig. \ref{fig3:distractortask} and Fig. \ref{fig4:OPsandbenchmark} in the main text and Figs. \ref{fig:dependencep}, \ref{fig:gdsingle}, \ref{fig:singlecontext} in SI \ref{subsec:singleheadnumerics}, we showed results with this simplified random feature approximation. In SI \ref{subsec:singleheadnumerics} Fig. \ref{fig:comparisonfull} we showed that results under the full Gibbs distribution exhibit qualitatively similar behaviors. In Fig. \ref{fig:dependencelambda} since we showed results with finite $\lambda$, we used $\langle f_T\rangle$ evaluated under the full Gibbs distribution, as given by SI Eqs. \ref{SIeq:meanpredictor}, \ref{SIeq:v}.

\subsubsection{The Order Parameters}
\label{paragraph:theops}
For two tasks ($T=2$), with the random feature approximation in the $\lambda\rightarrow\infty$ limit, we obtain a simple expression of $F_{2,1}\approx \mathcal{L}(\langle f_2\rangle, D_1)$ by plugging in $\langle f_2(\bm{x})\rangle$. 

\begin{equation}
F_{2,1} = \left(\bm{Y}_2 - \bm{K}_{2,1}^{L, 1} \left(\bm{K}_{1,1}^{L,1}\right)^{-1}\bm{Y}_1\right)^\top \left(\bm{K}_{2,2}^{L,1} \right)^{-1} \bm{K}_{2,1}^{L,1} \bm{K}_{1,2}^{L,1} \left(\bm{K}_{2,2}^{L,1}\right)^{-1}\left( \bm{Y}_2 - \bm{K}_{2,1}^{L,1}\left(\bm{K}_{1,1}^{L,1}\right)^{-1}\bm{Y}_1\right) /\lVert \bm{Y}_1\rVert^2 \label{eq:fullf21}
\end{equation}
Further assuming that the tasks are symmetric, i.e., the indices are interchangeable in each term, we have 

\begin{align}
F_{2,1} &= 2\Bigg( \underbrace{\frac{1}{2}\bm{Y}_2^\top \left(\bm{K}_{2,2}^{L,1}\right)^{-1}\bm{K}_{2,1}^{L,1} \left(\bm{I} + \left(\bm{K}_{1,1}^{L, 1}\right)^{-1}\bm{K}_{1,2}^{L, 1}\bm{K}_{2,1}^{L, 1} \left(\bm{K}_{1,1}^{L,1}\right)^{-1}\right) \bm{K}_{1,2}^{L,1}\left(\bm{K}_{2,2}^{L,1}\right)^{-1}\bm{Y}_2 /\lVert \bm{Y}_2\rVert^2}_{\gamma_{\text{RF}}} \nonumber\\
&  \underbrace{\bm{Y}_2^\top \left(\bm{K}_{2,2}^{L,1}\right)^{-1}\bm{K}_{2,1}^{L, 1} \bm{K}_{1,2}^{L,1}\left(\bm{K}_{2,2}^{L,1}\right)^{-1} \bm{K}_{2,1}^{L,1} \left(\bm{K}_{1,1}^{L,1}\right)^{-1}\bm{Y}_1/\left(\lVert \bm{Y}_1\rVert\lVert\bm{Y}_2\rVert\right)}_{\gamma_{\text{rule}}} \Bigg) \label{eq:fullop}
\end{align}

In SI Eqs. \ref{eq:fullf21}, \ref{eq:fullop}, $\{\bm{K}_{i,i'}^{L,1}\}_{i,i'\leq 2}$ are equivalent to $\{\bm{K}_{GP}^{L}(\bm{X}_i,\bm{X}_{i'})\}_{i,i'\leq 2}$ because $\lambda\rightarrow\infty$. SI Eq. \ref{eq:fullop} is equivalent to Eq. \ref{eq:f21} in the main text. In the main text, we expressed the OPs in terms of the rule vectors $\{\bm{V}_i\}_{i\leq 2} \in\mathbb{R}^N$ and the projection matrices $\{\bm{P}_i\}_{i\leq 2} \in \mathbb{R}^{N\times N}$ for better geometric interpretations (Eq. \ref{eq:f21}). However, as we take the $N\rightarrow\infty$ limit, $\{\bm{V}_i\}_{i\leq 2}$ and $\{\bm{P}_i\}_{i\leq 2}$ can not be directly evaluated. In fact, as shown in SI Eq. \ref{eq:fullop}, $\gamma_{\text{RF}}$ and $\gamma_{\text{rule}}$ can be expressed using $P$ dimensional quantities, including the labels $\{\bm{Y}_i\}_{i\leq 2} \in\mathbb{R}^P$ and the GP kernels $\{\bm{K}_{GP}^L({\bm{X}_i, \bm{X}_{i'}})\}_{i,i'\leq 2} \in\mathbb{R}^{P\times P}$, which allows us to compute them given any pair of training data $\{(\bm{X}_i, \bm{Y}_i)\}_{i\leq 2}$ even as $N\rightarrow \infty$. Similarly, the additional OP $\gamma_{\text{feature}}$ can be evaluated using $\gamma_{\text{feature}} = \frac{1}{P}\mathrm{Tr}\left(\left(\bm{K}_{2,2}^{L,1}\right)^{-1} \bm{K}_{2,1}^{L,1} \left(\bm{K}_{1,1}^{L,1}\right)^{-1} \bm{K}_{1,2}^{L,1}\right)$. 

Without any further assumptions, since the matrix $\mathcal{P}_{12}$ as defined in the main text is positive semi-definite, we have $\gamma_{\text{RF}} \geq 0$. It is straightforward that $\gamma_{\text{feature}} \in [0,1]$. Furthermore, given by the Cauchy-Schwarz inequality, we have 
\begin{align}
\lvert \gamma_{\text{rule}} \rvert &\leq \lVert \bm{Y}_2^\top \left(\bm{K}_{2,2}^{L,1}\right)^{-1}\bm{K}_{2,1}^{L,1}\rVert \cdot \lVert \bm{K}_{1,2}^{L,1} \left( \bm{K}_{2,2}^{L,1}\right)^{-1}\bm{K}_{2,1}^{L,1}\left(\bm{K}_{1,1}^{L,1}\right)^{-1} \bm{Y}_1\rVert/\left(\lVert\bm{Y}_1\rVert \lVert\bm{Y}_2\rVert\right) \nonumber \\
&= \lVert \bm{Y}_2^\top \left(\bm{K}_{2,2}^{L,1}\right)^{-1}\bm{K}_{2,1}^{L,1}\rVert \cdot \lVert \bm{K}_{2,1}^{L,1} \left( \bm{K}_{1,1}^{L,1}\right)^{-1}\bm{K}_{1,2}^{L,1}\left(\bm{K}_{2,2}^{L,1}\right)^{-1} \bm{Y}_2\rVert/\lVert \bm{Y}_2\rVert^2 \nonumber \\
&\leq \frac{1}{2}\left( \lVert \bm{Y}_2^\top \left(\bm{K}_{2,2}^{L,1}\right)^{-1}\bm{K}_{2,1}^{L,1}\rVert^2 + \lVert\bm{K}_{2,1}^{L,1} \left( \bm{K}_{1,1}^{L,1}\right)^{-1}\bm{K}_{1,2}^{L,1}\left(\bm{K}_{2,2}^{L,1}\right)^{-1} \bm{Y}_2\rVert^2\right)/\lVert\bm{Y}_2\rVert^2 = \gamma_{\text{RF}}
\end{align}
The equality follows from the assumption that the tasks are symmetric. 

For a long sequence of tasks, we assumed that any pair of tasks in a long sequence of length $T$ have the same relation, i.e., 
\begin{equation}
\mathcal{P}_{tt'} \equiv \frac{1}{2} \left( \bm{P}_t\left(\bm{X}_{t'}^L \right)^\top \bm{X}_{t'}^L \bm{P}_t + \bm{P}_{t'}\left(\bm{X}_{t}^L \right)^\top \bm{X}_{t}^L \bm{P}_{t'}\right)= \mathcal{P}, t, t' \leq T, t\neq t'
\end{equation}
$\gamma_{\text{rule}}$ can thus be seen as the inner product of two vectors $\tilde{\bm{V}}_t = \bm{V}_t^\top \mathcal{P}^{1/2}$ and $\tilde{\bm{V}}_{t'} = \bm{V}_{t'}^\top\mathcal{P}^{1/2}$, $t,t'\leq T, t\neq t'$. Since any two tasks have the same relation, the vectors $\tilde{\bm{V}}_t$ and $\tilde{\bm{V}}_{t'}$ have the same inner product for all $t,t'\leq T, t\neq t'$. We also assumed that $\lVert \tilde{\bm{V}}_{t} \rVert=\lVert \tilde{\bm{V}}\rVert$ for all $t$. The $T\times T$ covariance matrix of the vectors $\{\tilde{\bm{V}}_t\}_{t\leq T}$ is thus given by 
\begin{equation}
\lVert \tilde{\bm{V}}\rVert^2 \bm{I} + \gamma_{\text{rule}} \mathbf{1}\mathbf{1}^\top
\end{equation}
where $\mathbf{1}$ denotes an $N$-dimensional all-one vector. It follows from the positive semi-definiteness of this matrix, that $\gamma_{\text{rule}} \geq \lVert \bm{V}\rVert^2/T$. When $T\rightarrow\infty$, $\gamma_{\text{rule}} \in [0, \gamma_{\text{RF}}]$, as indicated in Section \ref{sec:single-head} of the main text, and shown in almost all examples in this paper. 

Further assuming that the random features of each task are orthogonal, i.e., $\bm{K}_{GP}^L(\bm{X}_i, \bm{X}_i) \propto \bm{I}$. We have $\gamma_{\text{RF}} = \frac{1}{2}\bm{V}_2^\top \bm{P}_1 (I+\bm{P}_2) \bm{P}_1 \bm{V}_2/\lVert\bm{V}_2\rVert^2$. Since the spectral norm of an orthogonal projection matrix is 1, we have $\gamma_{RF} \in [0, 1]$. 

\subsection{Detailed Derivations}
\label{subsubsec:derivationsingle}
In this section, we present the detailed derivation for the statistics of input-output mappings in single-head CL in the infinite-width limit.
\subsubsection{Moment Generating Function}
\label{paragraph:mgf}
We start from the MGF for multi-head CL, given by 
\begin{equation}
\mathcal{M}(\ell_{T})\equiv\left[Z(D_{1})\prod_{t=2}^{T}Z\left(\Theta_{t-1},D_{t}\right)\right]^{-1}\exp\left(-\beta E(\Theta_{1}|D_{1})-\sum_{t=2}^{T}\beta E\left(\Theta_{t}|\Theta_{t-1},D_{t}\right)+i\sum_{t=1}^{T}\ell_{T}f_{T}({\bf x})\right)\label{eq:mgf}
\end{equation}
where 
\begin{equation}
E\left(\Theta_{t}|\Theta_{t-1},D_{t}\right)=\frac{1}{2}\sum_{\mu=1}^{P}\left(f_{t}\left(\bm{x}_{t}^{\mu}\right)-y_{t}^{\mu}\right)^{2}+\frac{1}{2}\beta^{-1}\sigma^{-2}\left\Vert \Theta_{t}\right\Vert ^{2}+\frac{1}{2}\beta^{-1}\lambda\left\Vert \Theta_{t}-\Theta_{t-1}\right\Vert ^{2}.\label{eq:energy=000020fn-1-1}
\end{equation}
\begin{equation}
E(\Theta_{1},D_{1})=\frac{1}{2}\sum_{\mu=1}^{P}\left(f_{2}^{1}\left(\bm{x}_{1}^{\mu}\right)-y_{1}^{\mu}\right)^{2}+\frac{1}{2}\beta^{-1}\sigma^{-2}\left\Vert \Theta_{1}\right\Vert ^{2}
\end{equation}
and 
\begin{equation}
Z\left(\Theta_{t-1},D_{t}\right)\equiv\int d\Theta_{t}\exp(-\beta E(\Theta_{t}|\Theta_{t-1},D_{t}))
\end{equation}
\begin{equation}
Z(D_{1})\equiv\int d\Theta_{1}\exp(-\beta E(\Theta_{1},D_{1})
\end{equation}
Here we introduced field $\ell_{T}$ coupled to the mapping after learning all $T$ tasks, the statistics of $f_{T}({\bf x})$ can therefore be calculated by 
\begin{equation}
\langle f_{T}({\bf x})\rangle=-i\frac{\partial\mathcal{M}(\ell_{T})}{\partial\ell_{T}}\lvert_{\ell_{T}=0}
\end{equation}
\begin{equation}
\langle\delta^{2}f_{T}({\bf x})\rangle=-\frac{\partial^{2}\mathcal{M}(\ell_{T})}{\partial(\ell_{T})^{2}}\lvert_{\ell_{T}=0}
\end{equation}
We use the replica method for the denominator in SI Eq. $\text{\ref{eq:mgf}}$, and denote the physical copy of $\Theta_{t}$
as $\Theta_{t}^{n}$,we have 
\begin{equation}
Z(\Theta_{t-1},D_{t})^{-1}=Z(\Theta_{t-1}^{n},D_{t})^{-1}=\lim_{n\rightarrow0}\int\prod_{\alpha=1}^{n-1}d\Theta_{t}^{\alpha}\exp(-\beta\sum_{\alpha=1}^{n-1}E_{t}(\Theta_{t}^{\alpha}|\Theta_{t-1}^{n},D_{t})).
\end{equation}
We then introduce auxiliary integration variable $\left\{ \bm{v}_{t}^{\alpha}\right\} _{t=2,\cdots,T;\alpha=1,\cdots n},\bm{v}_{1}^{n}\in\mathbb{R}^{P}$
using the H-S transform, and arrive at
\begin{align}
\mathcal{M}\left(\ell_{T}\right) & =\lim_{n\rightarrow0}\intop\prod_{\alpha=1}^{n}\prod_{t=2}^{T}d\bm{v}_{t}^{\alpha}\int d\bm{v}_{1}^{n}\int\prod_{\alpha=1}^{n}\prod_{t=2}^{T}d\Theta_{t}^{\alpha}\int d\Theta_{1}^{n}\exp(-i\sum_{t,\alpha}\sum_{\mu=1}^{P}v_{t}^{\alpha,\mu}\left(f_{t}\left(\Theta_{t}^{\alpha},\bm{x}_{t}^{\mu}\right)-y_{t}^{\mu}\right)\nonumber \\
 & -\frac{1}{2}\sigma^{-2}\sum_{t,\alpha}\left\Vert \Theta_{t}^{\alpha}\right\Vert ^{2}-\frac{1}{2}\lambda\sum_{t=2}^{T}\sum_{\alpha=1}^{n}\left\Vert \Theta_{t}^{\alpha}-\Theta_{t-1}^{n}\right\Vert ^{2}-\frac{1}{2}\beta^{-1}\sum_{t,\alpha}\bm{v}_{t}^{\alpha\top}\bm{v}_{t}^{\alpha}+\sum_{t=1}^{T}\ell_{T}f_{T}(\Theta_{T}^{n},\bm{x}))
\end{align}
where we use $f_{t}\left(\Theta_{t}^{\alpha},\bm{x}_{t}^{\mu}\right)$
to denote the mapping with the replicated $\Theta_{t}^{\alpha},$$f_{t}(\Theta_{t}^{\alpha},{\bf x}_{t}^{\mu})\equiv\frac{1}{\sqrt{N}}\bm{a}_{t}^{\alpha}\Phi(\mathcal{W}_{t}^{\alpha},{\bf x}_{t}^{\mu})$;
and we use $f_{T}(\Theta_{T}^{n},\bm{x})$ to denote the mapping after
learning all $T$ tasks on arbitrary test input $\bm{x}$. Note that
$t=1$ is different from other time indices as we do not introduce
replica for $Z_{1}(D_{1})$. Therefore only $\Theta_{1}^{n}$ (and
thus $\bm{v}_{1}^{n}$) appears in the MGF. For notational convenience
we define $\sum_{t,\alpha}(\cdot)_{t,\alpha}\equiv(\cdot)_{1,n}+\sum_{t=2}^{T}\sum_{\alpha=1}^{n}(\cdot)_{t,\alpha}.$

We then integrate the readout weights $\left\{ \bm{a}_{t}^{\alpha}\right\} _{t=1,\cdots T;\alpha=1,\cdots n}$,
and obtain in the $\beta\rightarrow\infty$ limit
\begin{equation}
\mathcal{M}\left(\ell_{T}\right)=\lim_{n\rightarrow0}\intop\prod_{\alpha=1}^{n}\prod_{t=1}^{T}dv_{t}^{\alpha}\exp\left(-\frac{1}{2}\beta^{-1}\sum_{t,\alpha}\bm{v}_{t}^{\alpha\top}\bm{v}_{t}^{\alpha}+i\sum_{t,\alpha}\bm{v}_{t}^{\alpha\top}\bm{Y}_{t}+G(\left\{ \tilde{\bm{v}}_{t}^{\alpha}\right\} _{\alpha=1,\cdots,n;t=2,\cdots,T},\tilde{\bm{v}}_{1}^{n})\right)
\end{equation}
where 
\begin{equation}
G(\left\{ \tilde{\bm{v}}_{t}^{\alpha}\right\} _{\alpha=1,\cdots,n;t=2,\cdots,T},\tilde{\bm{v}}_{1}^{n})\equiv\log\left[\intop\prod_{\alpha=1}^{n}\prod_{t=1}^{T}d\mathcal{W}_{t}^{\alpha}\exp\left(S_{0}\left(\left\{ \mathcal{W}^{\alpha}\right\} _{\alpha=1,\cdots,n}\right)-\frac{1}{2}\sum_{t,\alpha}\sum_{t',\beta}m_{t,t'}^{\alpha\beta}\tilde{\bm{v}}_{t}^{\alpha\top}\bm{M}_{t,t'}^{\alpha\beta}\tilde{\bm{v}}_{t'}^{\beta}\right)\right]
\end{equation}
\begin{equation}
S_{0}(\left\{ \mathcal{W}^{\alpha}\right\} _{\alpha=1,\cdots,n})\equiv-\frac{1}{2}\sigma^{-2}\sum_{t,\alpha}\left\Vert \mathcal{W}_{t}^{\alpha}\right\Vert ^{2}-\frac{1}{2}\lambda\sum_{t=2}^{T}\sum_{\alpha=1}^{n}\left\Vert \mathcal{W}_{t}^{\alpha}-\mathcal{W}_{t-1}^{n}\right\Vert ^{2}\label{eq:prior}
\end{equation}
and 
\begin{equation}
\bm{M}_{t,t'}^{\alpha\beta}\equiv\left[\begin{array}{cc}
\frac{1}{N}\Phi(\mathcal{W}_{t}^{\alpha},\bm{X}_{t})\cdot\Phi(\mathcal{W}_{t'}^{\beta},\bm{X}_{t'})\in\mathbb{R}^{P\times P} & \frac{1}{N}\Phi(\mathcal{W}_{t}^{\alpha},\bm{X}_{t})\cdot\Phi(\mathcal{W}_{T}^{\beta},{\bf x})\in\mathbb{R}^{P\times1}\\
\frac{1}{N}\Phi(\mathcal{W}_{T}^{\alpha},{\bf x})\cdot\Phi(\mathcal{W}_{t'}^{\beta},\bm{X}_{t'})\in\mathbb{R}^{1\times P} & \frac{1}{N}\Phi(\mathcal{W}_{T}^{\alpha},{\bf x})\cdot\Phi(\mathcal{W}_{T}^{\beta},{\bf x})\in\mathbb{R}
\end{array}\right]\in\mathbb{R}^{(P+1)\times(P+1)}\label{eq:M}
\end{equation}
For simplicity, here we denote $\tilde{\bm{v}}_{t}^{\alpha}\equiv[\bm{v}_{t}^{\alpha},\delta_{\alpha n}\delta_{tT}\ell_{T}]\in\mathbb{R}^{P+1}$, absorbing the field coupled to the mapping on arbitrary ${\bf x}$ into $\bm{v}_{t}^{\alpha}$.  Since $m_{t,t'}^{\alpha\beta}$ is symmetric in $t,t'$, w.l.o.g., we assume $t\geq t'$. For $t,t'\geq2$, we have
\begin{equation}
m_{t,t'}^{\alpha\beta}=\begin{cases}
m_{t,t^{\prime}}^{1}=\frac{\sigma^{2}}{1+\tilde{\lambda}}(\tilde{\lambda}^{t-t'}+\tilde{\lambda}^{t+t'-1}) & \left\{ \alpha=\beta,t=t^{\prime}\right\} \cup\left\{ \beta=n,t>t^{\prime}\right\} \\
m_{\tau,\tau^{\prime}}^{0}=\frac{\sigma^{2}}{1+\tilde{\lambda}}(\tilde{\lambda}^{t-t'+2}+\tilde{\lambda}^{t+t'-1}) & otherwise
\end{cases}
\end{equation}
where $\tilde{\lambda}\equiv\frac{\lambda}{\lambda+\sigma^{-2}}.$
Otherwise we denote
\begin{equation}
m_{t,1}^{\alpha n}=m_{t,1}^{1}=\sigma^{2}\tilde{\lambda}^{t-1};m_{t,1}^{0}\equiv0
\end{equation}

While it is in general highly nontrivial to evaluate $G$, in the
infinite-width limit, the distribution of $\mathcal{W}$ is dominated
by the Gaussian prior determined by $S_{0}(\mathcal{W})$, and the
weights become self-averaging. $G$ is thus given by 
\begin{equation}
G(\left\{ \tilde{\bm{v}}_{t}^{\alpha}\right\} _{\alpha=1,\cdots,n;t=2,\cdots,T},\tilde{\bm{v}}_{1}^{n})=-\frac{1}{2}\sum_{t,\alpha}\sum_{t,\beta}m_{t,t'}^{\alpha\beta}\tilde{\bm{v}}_{t}^{\alpha\top}\langle\bm{M}_{t,t'}^{\alpha\beta}\rangle_{\mathcal{W}}\tilde{\bm{v}}_{t'}^{\beta}\label{eq:Gsinglehead}
\end{equation}
where $\langle\cdot\rangle_{\mathcal{W}}$ denotes averaging over
the prior Gaussian distribution proportional to $\exp(-S_{0}(\left\{ \mathcal{W}^{\alpha}\right\} _{\alpha=1,\cdots,n}))$. 

\subsubsection{Definition of Kernel Functions}
\label{paragraph:kernelfunctions}
We observe that due to the structure of the prior distribution, $\langle\bm{M}_{t,t'}^{\alpha\beta}\rangle_{\mathcal{W}}$
can be expressed by two different kernel functions, defined on arbitrary
inputs $\bm{x}$ and $\bm{x}'$.The kernel functions are symmetric
in $t$ and $t'$, so w.l.o.g. we define them with $t\geq t'$. 
\begin{equation}
K_{t,t'}^{L,1}(\bm{x},\bm{x}')\equiv\frac{1}{N}\langle\Phi(\mathcal{W}_{t},\bm{x})\cdot\Phi(\mathcal{W}_{t'},\bm{x}')\rangle_{\mathcal{W}}\label{eq:kgp1}
\end{equation}
\begin{equation}
K_{t,t'}^{L,0}(\bm{x},\bm{x}')\equiv\frac{1}{N}\langle\langle\Phi(\mathcal{W}_{t},\bm{x})\rangle_{t,t'}\cdot\langle\Phi(\mathcal{W}_{t'},\bm{x}')\rangle_{t,t'}\rangle_{\mathcal{W}}\label{eq:kgp0}
\end{equation}
where $\langle\cdot\rangle_{\mathcal{W}}$ denotes the average over
the full prior distribution SI Eq. \ref{eq:priorw}, and $\langle\cdot\rangle_{t,t'}$
denotes the partial average over the conditional distribution $P(\mathcal{W}_{t,}\mathcal{W}_{t-1},\cdots,\mathcal{W}_{t'}|\mathcal{W}_{t'-1})$. 

$\langle\bm{M}_{t,t'}^{\alpha\beta}\rangle_{\mathcal{W}}$ can be
expressed as 
\begin{equation}
\langle\bm{M}_{t,t'}^{\alpha\beta}\rangle_{\mathcal{W}}=\left[\begin{array}{cc}
\bm{K}_{t,t'}^{L,\alpha\beta}\in\mathbb{R}^{P\times P} & \bm{k}_{t,T}^{L,\alpha n}(\bm{x})\in\mathbb{R}^{P\times1}\\
\bm{k}_{T,t'}^{L,n\beta}(\bm{x})\in\mathbb{R}^{1\times P} & k_{T,T}^{L,nn}(\bm{x},\bm{x})\in\mathbb{R}
\end{array}\right]\label{eq:averagedM}
\end{equation}
with $\bm{K}_{t,t'}^{L,\alpha\beta}$,$\bm{k}_{t,T}^{L,\alpha n}(\bm{x})$,$\bm{k}_{T,t}^{L,n\beta}(\bm{x})$
and $k_{T,T}^{L,nn}(\bm{x},\bm{x})$ denoting the 4 blocks corresponding
to SI Eq. \ref{eq:M}. They are given by applying
the kernel functions on the training and testing data, respectively.Again since $\langle\bm{M}_{t,t'}^{\alpha\beta}\rangle_{\mathcal{W}}$ is symmetric in $t$ and $t',$ w.l.o.g. for $t\geq t'$, we have

\begin{equation}
\bm{K}_{t,t'}^{L,\alpha\beta}=\begin{cases}
\bm{K}_{t,t'}^{L,1}\equiv K_{t,t'}^{L,1}(\bm{X}_{t},\bm{X}_{t'}) & \left\{ \alpha=\beta,t=t^{\prime}\right\} \cup\left\{ \beta=n,t>t^{\prime}\right\} \\
\bm{K}_{t,t'}^{L,0}\equiv K_{t,t'}^{L,0}(\bm{X}_{t},\bm{X}_{t'}) & otherwise
\end{cases}\label{eq:datamatrix}
\end{equation}
\begin{equation}
\bm{k}_{t,T}^{L,\alpha n}(\bm{x})=\begin{cases}
\bm{k}_{t,T}^{L,1}(\bm{x})\equiv K_{t,T}^{L,1}(\bm{X}_{t},\bm{x}) & \alpha=n\\
\bm{k}_{t,T}^{L,0}(\bm{x})\equiv K_{t,T}^{L,0}(\bm{X}_{t},\bm{x}) & otherwise
\end{cases}\label{eq:testdata}
\end{equation}
\begin{equation}
k_{T,T}^{L,nn}(\bm{x},\bm{x})=k_{T,T}^{L,1}(\bm{x},\bm{x})\equiv K_{T,T}^{L,1}(\bm{x},\bm{x})\label{eq:testdatascalar}
\end{equation}
We introduced notations $\bm{K}_{t,t'}^{L,1}$,$\bm{K}_{t,t'}^{L,0}$,$\bm{k}_{t,T}^{L,1}(\bm{x})$,
$\bm{k}_{t,T}^{L,0}(\bm{x})$ and $k_{T,T}^{L,1}(\bm{x},\bm{x})$,
for applying the kernel functions (SI Eqs. \ref{eq:kgp1}, \ref{eq:kgp0}) 
on the training and testing data.

For notational convenience, we introduce another kernel function as
it will appear frequently in the statistics of input-output mappings
\begin{equation}
\tilde{K}_{t,t'}^{L}(\bm{x},\bm{x}')\equiv m_{t,t'}^{1}K_{t,t'}^{L,1}(\bm{x},\bm{x}')-m_{t,t'}^{0}K_{t,t'}^{L,0}(\bm{x},\bm{x}')\label{eq:generalized-ntk}
\end{equation}
Applying this kernel function on the training and testing data, we
have $\tilde{\bm{K}}_{t,t'}^{L}=m_{t,t'}^{1}\bm{K}_{t,t'}^{L,1}-m_{t,t'}^{0}\bm{K}_{t,t'}^{L,0}$
, $\tilde{\bm{k}}_{t,t'}^{L}(\bm{x})=m_{t,t'}^{1}\bm{k}_{t,t'}^{L,1}(\bm{x})-m_{t,t'}^{0}\bm{k}_{t,t'}^{L,0}(\bm{x})$.
We will use the same notations for these kernel functions and for
applying them on training and testing data throughout the supplementary.
Interestingly, in the limit $\lambda\rightarrow\infty$, $\tilde{K}_{t,t'}^{L}(\bm{x},\bm{x}')$
corresponds to a generalized two-times Neural Tangent Kernel, as we
will show in SI \ref{subsec:singleheadtheory}.\ref{paragraph:ktilde-and-ntk}. 

\subsubsection{Derivation of the Statistics of Input-Output Mappings}

\label{paragraph:predictor-statistics}

With the above definition of the kernel functions, we can replace
$\langle\bm{M}_{t,t'}^{\alpha\beta}\rangle_{\mathcal{W}}$ with the
corresponding kernels, and thus rewriting $\mathcal{M}(\ell_{T})$
as
\begin{align}
\mathcal{M}\left(\ell_{T}\right) & =\lim_{n\rightarrow0}\intop\prod_{\alpha=1}^{n}\prod_{t=1}^{T}d\bm{v}_{t}^{\alpha}\exp(-\frac{1}{2}\beta^{-1}\sum_{t,\alpha}\bm{v}_{t}^{\alpha\top}\bm{v}_{t}^{\alpha}+i\sum_{t,\alpha}\bm{v}_{t}^{\alpha\top}\bm{Y}_{t} -\frac{1}{2}\sum_{t,\alpha}\bm{v}_{t}^{\alpha}\tilde{\bm{K}}_{t,t'}^{L}\bm{v}_{t}^{\alpha}-\frac{1}{2}\sum_{t=2}^{T}\sum_{\alpha,\beta=1}^{n}m_{t,t}^{0}\bm{v}_{t}^{\alpha\top}\bm{K}_{t,t}^{L,0}\bm{v}_{t}^{\beta}\nonumber \\
 & -\sum_{t=t'+1}^{T}\sum_{t'=1}^{T}\sum_{\alpha=1}^{n}\bm{v}_{t}^{\alpha\top}\tilde{\bm{K}}_{t,t'}^{L}\bm{v}_{t'}^{n}-\sum_{t=t'+1}^{T}\sum_{t'=2}^{T}\sum_{\alpha,\beta=1}^{n}m_{t,t'}^{0}\bm{v}_{t}^{\alpha}\bm{K}_{t,t'}^{L,0}\bm{v}_{t'}^{\beta}\nonumber \\
 &-\ell_{T}\sum_{t=1}^{T}\tilde{\bm{k}}_{T,t}^{L}(\bm{x})\bm{v}_{t}^{n}-\sum_{t=2}^{T}\sum_{\beta=1}^{n}m_{T,t}^{0}\ell_{T}\bm{k}_{T,t}^{L,0}(\bm{x})\bm{v}_{t}^{\beta}-\frac{1}{2}\ell_{T}^{2}m_{T,T}^{1}k_{T,T}^{L,1}(\bm{x},\bm{x}))
\end{align}
The remaining calculation is to integrate over $\left\{ \bm{v}_{t}^{\alpha}\right\} _{t=1,\cdots,T;\alpha=1,\cdots,n}$.
To decouple the replica indices, we introduce $\bm{p}_{t}=\sum_{\alpha=1}^{n}\bm{v}_{t}^{\alpha}$,
and its corresponding conjugate variable $\bm{q}_{t}$. Using Fourier
representation of the Dirac delta function $\delta(\bm{p}_{t}-\sum_{\alpha=1}^{n}\bm{v}_{t}^{\alpha})=\int d\bm{q}_{t}\exp\left(i\bm{q}_{t}\left(\bm{p}_{t}-\sum_{\alpha=1}^{n}\bm{v}_{t}^{\alpha}\right)\right)$,
we rewrite $\mathcal{M}(\ell_{T})$ as 
\begin{align}
\mathcal{M}\left(\ell_{T}\right)= & \lim_{n\rightarrow0}\int d\bm{v}_{1}\intop\prod_{\alpha=1}^{n}\prod_{t=2}^{T}d\bm{v}_{t}^{\alpha}\int\prod_{t=1}^{T}d\bm{p}_{t}\int\prod_{t=1}^{T}d\bm{q}_{t}\exp\Bigg[-\frac{1}{2}\beta^{-1}\sum_{t,\alpha}\bm{v}_{t}^{\alpha\top}\bm{v}_{t}^{\alpha}+i\sum_{t=2}^{T}\bm{p}_{t}^{\top}\bm{Y}_{t}+i\bm{v}_{1}^{\top}\bm{Y}_{1}\\
 & +i\sum_{t=2}^{T}\bm{q}_{t}(\bm{p}_{t}-\sum_{\alpha=1}^{n}\bm{v}_{t}^{\alpha})-\frac{1}{2}\sum_{t,\alpha}\bm{v}_{t}^{\alpha}\tilde{\bm{K}}_{t,t}^{L}\bm{v}_{t}^{\alpha}-\frac{1}{2}\sum_{t=2}^{T}m_{t,t}^{0}\bm{p}_{t}^{\top}\bm{K}_{t,t}^{L,0}\bm{p}_{t} \nonumber \\ 
 &-\sum_{t=t'+1}^{T}\sum_{t'=1}^{T}\bm{p}_{t}^{\top}\tilde{\bm{K}}_{t,t'}^{L}\bm{v}_{t'}-\sum_{t=t'+1}^{T}\sum_{t'=2}^{T}m_{t,t'}^{0}\bm{p}_{t}^{\top}\bm{K}_{t,t'}^{L,0}\bm{p}_{t'}\nonumber \\
 & -\ell_{T}\sum_{t=1}^{T}\tilde{\bm{k}}_{T,t}^{L}(\bm{x})\bm{v}_{t}^{n}-\sum_{t=2}^{T}m_{T,t}^{0}\ell_{T}\bm{k}_{T,t}^{L,0}(\bm{x})\bm{p}_{t}-\frac{1}{2}\ell_{T}^{2}m_{T,T}^{1}k_{T,T}^{L,1}(\bm{x},\bm{x})\Bigg]
\end{align}
We note that the different replica indices have been decoupled, which
allows us to integrate over $\left\{ \bm{v}_{t}^{\alpha}\right\} _{\alpha=1,\cdots,n}$
independently. Let $\bm{v}_{t}\equiv\bm{v}_{t}^{n}$, integrate over
$\left\{ \bm{v}_{t}^{\alpha}\right\} _{\alpha=1}^{n-1}$, and keep
only the $\mathcal{O}(1)$ terms (neglecting $\mathcal{O}(n)$ contributions),
we have 
\begin{align}
\mathcal{M}\left(\ell_{T}\right)&=\int\prod_{t=2}^{T}d\bm{p}_{t}\int\prod_{t=1}^{T}d\bm{v}_{t}\exp\Bigg[-\frac{1}{2}\sum_{t=2}^{T}\bm{p}_{t}^{\top}(m_{t,t}^{0}\bm{K}_{t,t}^{L,0}-\tilde{\bm{K}}_{t,t}^{L})\bm{p}_{t}-\frac{1}{2}m_{1,1}^{1}\bm{v}_{1}^{\top}\bm{K}_{1,1}^{L,1}\bm{v}_{1}\nonumber\\
&-\sum_{t=t'}^{T}\sum_{t'=2}^{T}\bm{p}_{t}^{\top}\tilde{\bm{K}}_{t,t'}^{L}\bm{v}_{t'}-\sum_{t=t'+1}^{T}\sum_{t'=2}^{T}m_{t,t'}^{0}\bm{p}_{t}^{\top}\bm{K}_{t,t'}^{L,0}\bm{p}_{t'}-\sum_{t=2}^{T}m_{t,1}^{1}\bm{v}_{1}^{\top}\bm{K}_{1,t}^{L,1}\bm{p}_{t}+i\sum_{t=2}^{T}\bm{p}_{t}^{\top}\bm{Y}_{t}+i\bm{v}_{1}^{\top}\bm{Y}_{1}\nonumber\\
&-\frac{1}{2}\ell_{T}^{2}m_{T,T}^{1}k_{T,T}^{L,1}(\bm{x},\bm{x})-\ell_{T}\sum_{t=2}^{T}m_{t,T}^{0}\bm{k}_{T,t}^{L,0}(\bm{x})\bm{p}_{t}-\ell_{T}\sum_{t=1}^{T}\tilde{\bm{k}}_{T,t}^{L}(\bm{x})\bm{v}_{t}\Bigg]\label{eq:mgfsingle}
\end{align}
We have eliminated all the replica indices, allowing us to proceed
to computing the mapping statistics.

\paragraph{The Mean Input-Output Mappings} 
The average mapping can be obtained by taking derivative of $\mathcal{M}(\ell_{T})$
w.r.t. $\ell_{T}$, resulting in 
\begin{equation}
\langle f_{T}(\bm{x})\rangle=\sum_{t=2}^{T}m_{T,t}^{0}\bm{k}_{T,t}^{L,0}(\bm{x})\langle-i\bm{p}_{t}\rangle+\sum_{t=1}^{T}\tilde{\bm{k}}_{T,t}^{L}(\bm{x})\langle-i\bm{v}_{t}\rangle
\end{equation}
with the statistics of $\bm{p}_{t}$ and $\bm{v}_{t}$ determined
by $\mathcal{M}(\ell_{T}=0)$, resulting in $\langle\bm{p}_{t}\rangle=0$
and 
\begin{equation}
\langle-i\bm{v}_{t}\rangle=\left(\tilde{\bm{K}}_{t,t}^{L}\right)^{-1}(\bm{Y}_{t}-\sum_{t'=1}^{t-1}\tilde{\bm{K}}_{t,t'}^{L}\langle-i\bm{v}_{\tau'}\rangle)\label{eq:meanv}
\end{equation}
The mean mapping thus simplifies to 
\begin{equation}
\langle f_{T}(\bm{x})\rangle=\sum_{t=1}^{T}\tilde{\bm{k}}_{T,t}^{L}(\bm{x})\langle-i\bm{v}_{t}\rangle.
\end{equation}

\paragraph{The Variance of the Input-Output Mappings}

The variance of the mapping can be evaluated by taking the second
derivative of $\mathcal{M}(\ell_{T})$ w.r.t. $\ell_{T}$, resulting
in 
\begin{align}
\langle\delta^{2}f_{T}(\bm{x})\rangle & =m_{T,T}^{1}k_{T,T}^{L,1}(\bm{x},\bm{x})-\sum_{t,t'=2}^{T}\left(m_{T,t}^{0}\right)^{2}\bm{k}_{T,t}^{L,0}(\bm{x})\langle\delta\bm{p}_{t}\delta\bm{p}_{t'}^{\top}\rangle\bm{k}_{T,t}^{L,0}(\bm{x})^{\top}\nonumber \\
 & -\sum_{t,t'=1}^{T}\tilde{\bm{k}}_{T,t}^{L}(\bm{x})\langle\delta\bm{v}_{t}\delta\bm{v}_{t'}^{\top}\rangle\tilde{\bm{k}}_{T,t'}^{L}(\bm{x})^{\top}-2\sum_{t=2}^{T}\sum_{t'=1}^{T}m_{t,T}^{0}\bm{k}_{T,t}^{L,0}(\bm{x})\langle\delta\bm{p}_{t}\delta\bm{v}_{t'}^{\top}\rangle\tilde{\bm{k}}_{T,t'}^{L}(\bm{x})^{\top}
\end{align}
The statistics of $\bm{p}_{t}$ and $\bm{v}_{t}$ is again determined
by $\mathcal{M}(\ell_{T}=0)$, and we have $\langle\delta\bm{p}_{t}\delta\bm{p}_{t'}^{\top}\rangle=0$
\begin{equation}
\langle\delta\bm{p}_{t}\delta\bm{v}_{t'}^{\top}\rangle=\begin{cases}
0 & t'<t\\
\left(\tilde{\bm{K}}_{t,t}^{L}\right)^{-1} & t=t'\\
-\left(\tilde{\bm{K}}_{t,t}^{L}\right)^{-1}\left(\sum_{\tau=t+1}^{t'}\tilde{\bm{K}}_{t,\tau}^{L}\langle\delta\bm{p}_{\tau}\delta\bm{v}_{t'}^{\top}\rangle\right) & t'>t
\end{cases}\label{eq:covpv}
\end{equation}
$\langle\delta\bm{v}_{t}\delta\bm{v}_{t'}^{\top}\rangle$ is symmetric
in $t$,$t'$, so we show $\langle\delta\bm{v}_{t}\delta\bm{v}_{t'}^{\top}\rangle$
only for $t\leq t'$
\begin{align}
\langle\delta\bm{v}_{t}\delta\bm{v}_{t'}^{\top}\rangle & =-\left(\tilde{\bm{K}}_{t,t}^{L}\right)^{-1}(\sum_{\tau=2}^{t-1}\tilde{\bm{K}}_{t,\tau}^{L}\langle\delta\bm{v}_{\tau}\delta\bm{v}_{t'}^{\top}\rangle+(m_{t,t}^{0}\bm{K}_{t,t}^{L,0}-\tilde{\bm{K}}_{t,t}^{L})\langle\delta\bm{p}_{t}\delta\bm{v}_{t'}^{\top}\rangle\nonumber \\
 & +\sum_{\tau=2}^{t'}(1-\delta_{\tau,t})m_{t,\tau}^{0}\bm{K}_{t,\tau}^{L,0}\langle\delta\bm{p}_{\tau}\delta\bm{v}_{t'}^{\top}\rangle+(1-\delta_{1,t})m_{1,t}^{1}\bm{K}_{t,1}^{L,1}\langle\delta\bm{p}_{1}\delta\bm{v}_{t'}^{\top}\rangle)\label{eq:covv}
\end{align}
Thus the variance can be simplified as 
\begin{align}
\langle\delta^{2}f_{T}(\bm{x})\rangle & =m_{T,T}^{1}k_{T,T}^{L,1}(\bm{x},\bm{x})-\sum_{t,t'=1}^{T}\tilde{\bm{k}}_{T,t}^{L}(\bm{x})\langle\delta\bm{v}_{t}\delta\bm{v}_{t'}^{\top}\rangle\tilde{\bm{k}}_{T,t'}^{L}(\bm{x})^{\top}\nonumber \\
 & -2\sum_{t'=1}^{T}\sum_{t=2}^{t'}m_{t,T}^{0}\bm{k}_{T,t}^{L,0}(\bm{x})\langle\delta\bm{p}_{t}\delta\bm{v}_{t'}^{\top}\rangle\tilde{\bm{k}}_{T,t'}^{L}(\bm{x})^{\top}
\end{align}
The variance can therefore be calculated iteratively. Since $\left\{ m_{t,t'}^{1}\right\} _{t,t'=1,\cdots,T},\left\{ m_{t,t'}^{0}\right\} _{t,t'=1,\cdots,T}$
scales as $\sigma^{2}$ , and the GP kernels scale as $\sigma^{2L}$,
the variance scales with $\sigma^{2(L+1)}$, therefore when $\sigma$
is small, the variance contribution can be neglected. For simplicity,
we focus on the contribution of the bias term to the performance,
namely $\langle\mathcal{L}(f_{T},D)\rangle\approx\mathcal{L}(\langle f_{T}\rangle,D)$.

\subsubsection{Analytical Forms of Kernel Functions in Linear and ReLU Neurons}
\label{paragraph:kernelfunctionexpressions}
The kernel functions in SI \ref{subsec:singleheadtheory}.\ref{paragraph:kernelfunctions}
can be evaluated iteratively across layers, using 
\begin{equation}
K_{t,t'}^{L,1}(\bm{x},\bm{x}')=F(m_{t,t}^{1}K_{t,t}^{L-1,1}(\bm{x},\bm{x}),m_{t',t'}^{1}K_{t',t'}^{L-1,1}(\bm{x}',\bm{x}'),m_{t,t'}^{L-1,1}K_{t,t'}^{L-1,1}(\bm{x},\bm{x}'))\label{eq:iterativek1}
\end{equation}
\begin{equation}
K_{t,t'}^{L,0}(\bm{x},\bm{x}')=F(m_{t,t}^{1}K_{t,t}^{L-1,1}(\bm{x},\bm{x}),m_{t',t'}^{1}K_{t',t'}^{L-1,1}(\bm{x}',\bm{x}'),m_{t,t'}^{L-1,0}K_{t,t'}^{L-1,0}(\bm{x},\bm{x}'))\label{eq:iterativek0}
\end{equation}
with the initial conditions 
\begin{equation}
K_{t,t'}^{L=0,1}(\bm{x},\bm{x}')=K_{t,t'}^{L=0,0}(\bm{x},\bm{x}')=N_{0}^{-1}\bm{x}\cdot\bm{x}'
\end{equation}
The function $F(\mathbb{E}[z^{2}],\mathbb{E}[z^{'2}],\mathbb{E}[zz'])$
is a function of the variances of two Gaussian variables $z$ and
$z'$ and their covariance. The form of $F$ depends on the nonlinearity
of the network \citep{cho2009kernel}. $F$ has analytical forms
for certain types of nonlinearities $\phi$. In this paper we show
results for networks with linear or ReLU nonlinearities. We present
the analytical forms of the kernels in this section. 

For linear networks, 
\begin{equation}
K_{t,t'}^{L,0}(\bm{x},\bm{x}')=N_{0}^{-1}\left(m_{t,t'}^{0}\right)^{L}\bm{x}\cdot\bm{x}'
\end{equation}
\begin{equation}
K_{t,t'}^{L,1}(\bm{x},\bm{x}')=N_{0}^{-1}\left(m_{t,t'}^{1}\right)^{L}\bm{x}\cdot\bm{x}'
\end{equation}
\begin{equation}
\tilde{K}_{t,t'}^{L}(\bm{x},\bm{x}')=N_{0}^{-1}\left(\left(m_{t,t'}^{1}\right)^{L+1}-\left(m_{t,t'}^{0}\right)^{L+1}\right)\bm{x}\cdot\bm{x}'
\end{equation}
In the $\lambda\rightarrow\infty$ limit, $\tilde{K}_{t,t'}^{L}(\bm{x},\bm{x}')$
scales with $\lambda^{-1}$, and can be given by
\begin{equation}
\tilde{K}_{t,t'}^{L}(\bm{x},\bm{x}')=N_{0}^{-1}(L+1)\lambda^{-1}\bm{x}\cdot\bm{x}'
\end{equation}

For ReLU nonlinearity, we first define the function 
\begin{equation}
J(\theta)=(\pi-\theta)\cos(\theta)+\sin(\theta)
\end{equation}
Then we have 
\begin{equation}
K_{t,t'}^{L,0}(\bm{x},\bm{x}')=\frac{\sqrt{K_{t,t}^{L-1,1}(\bm{x},\bm{x})K_{t',t'}^{L-1,1}(\bm{x}',\bm{x}')}}{2\pi}J(\theta_{t,t'}^{L-1,0}(\bm{x},\bm{x}'))
\end{equation}
\begin{equation}
K_{t,t'}^{L,1}(\bm{x},\bm{x}')=\frac{\sqrt{K_{t,t}^{L-1,1}(\bm{x},\bm{x})K_{t',t'}^{L-1,1}(\bm{x}',\bm{x}')}}{2\pi}J(\theta_{t,t'}^{L-1,1}(\bm{x},\bm{x}'))
\end{equation}
where
\begin{equation}
\theta_{t,t'}^{L,0}(\bm{x},\bm{x}')=\cos^{-1}\left(\frac{m_{t,t'}^{0}K_{t,t'}^{L,0}(\bm{x},\bm{x}')}{\sqrt{m_{t,t}^{1}K_{t,t}^{L,1}(\bm{x},\bm{x})}\sqrt{m_{t',t'}^{1}K_{t',t'}^{L,1}(\bm{x'},\bm{x}')}}\right)
\end{equation}
and 
\begin{equation}
\theta_{t,t'}^{L,1}(\bm{x},\bm{x}')=\cos^{-1}\left(\frac{m_{t,t'}^{1}K_{t,t'}^{L,1}(\bm{x},\bm{x}')}{\sqrt{m_{t,t}^{1}K_{t,t}^{L,1}(\bm{x},\bm{x})}\sqrt{m_{t',t'}^{1}K_{t',t'}^{L,1}(\bm{x'},\bm{x}')}}\right)
\end{equation}
with the initial condition that 
\begin{equation}
K_{t,t'}^{L=0,0}(\bm{x},\bm{x}')=K_{t,t'}^{L=0,1}(\bm{x},\bm{x}')=N_{0}^{-1}\bm{x}\cdot\bm{x}'
\end{equation}
As usual we have $\tilde{K}_{t,t'}^{L}(\bm{x},\bm{x}')=m_{t,t'}^{1}K_{t,t'}^{L,1}(\bm{x},\bm{x}')-m_{t,t'}^{0}K_{t,t'}^{L,0}(\bm{x},\bm{x}')$.In
the $\lambda\rightarrow\infty$ limit, $\tilde{K}_{t,t'}^{L}(\bm{x},\bm{x}')$
scales with $\lambda^{-1}$, and is given iteratively by 
\begin{equation}
\tilde{K}_{t,t'}^{L}(\bm{x},\bm{x}')=\lambda^{-1}K_{t,t'}^{L,1}(\bm{x},\bm{x}')+\sigma^{2}\frac{1}{2\pi}(\pi-\theta_{t,t'}^{L-1,1}(\bm{x},\bm{x}'))\tilde{K}_{t,t'}^{L-1}(\bm{x},\bm{x}')
\end{equation}
with initial condition 
\begin{equation}
\tilde{K}_{t,t'}^{L=0}(\bm{x},\bm{x}')=\lambda^{-1}N_{0}^{-1}\bm{x}\cdot\bm{x}'
\end{equation}

\subsubsection{$\tilde{K}_{t,t'}^{L}$ and the Neural Tangent Kernel}
\label{paragraph:ktilde-and-ntk}
In this section, we show that the kernel function $\tilde{K}_{t,t'}^{L}(\bm{x},\bm{x}')$
defined in SI Eq. \ref{eq:generalized-ntk} and appearing in the input-output mapping statistics in SI \ref{subsec:singleheadtheory}.\ref{paragraph:predictor-statistics}
is closely related the the neural tangent kernel \citep{jacot2018neural},
in the limit $\lambda\rightarrow\infty$.

\paragraph{Iterative Expression of $\tilde{K}_{t,t'}^{L}$}

First, we derive an iterative expression of $\tilde{K}_{t,t'}^{L}(\bm{x},\bm{x}')$
in the $\lambda\rightarrow\infty$ limit. By expanding in $\mathcal{O}(\lambda^{-1})$,
we can rewrite the $\tilde{K}_{t,t'}^{L}(\bm{x},\bm{x}')$ as 
\begin{equation}
\tilde{K}_{t,t'}^{L}(\bm{x},\bm{x}')=\lambda^{-1}K_{t,t'}^{L,1}(\bm{x},\bm{x}')+\sigma^{2}\Delta K_{t,t'}^{L}(\bm{x},\bm{x}')
\end{equation}
where $\Delta K_{t,t'}^{L}(\bm{x},\bm{x}')$ is defined as
\begin{equation}
\Delta K_{t,t'}^{L}(\bm{x},\bm{x}')\equiv K_{t,t'}^{L,1}(\bm{x},\bm{x}')-K_{t,t'}^{L,0}(\bm{x},\bm{x}')
\end{equation}
Applying and expanding SI Eqs. \ref{eq:iterativek1}, \ref{eq:iterativek0},
we have 
\begin{equation}
\Delta K_{t,t'}^{L}(\bm{x},\bm{x}')=\dot{K}_{t,t'}^{L}(\bm{x},\bm{x}')\tilde{K}_{t,t'}^{L-1}(\bm{x},\bm{x}')
\end{equation}
\begin{equation}
\dot{K}_{t,t'}^{L}(\bm{x},\bm{x}')\equiv\langle\phi'(h_{t}^{L}(\bm{x}))\cdot\phi'(h_{t'}^{L}(\bm{x}'))\rangle
\end{equation}
where $h_{t}^{L}(\bm{x})\equiv N^{-1/2}W_{t}^{L}\cdot\bm{x}$ is the
pre-activation at the $L$-th layer.Thus we have an iterative relation
\begin{equation}
\tilde{K}_{t,t'}^{L}(\bm{x},\bm{x}')=\lambda^{-1}K_{t,t'}^{L,1}(\bm{x},\bm{x}')+\sigma^{2}\dot{K}_{t,t'}^{L}(\bm{x},\bm{x}')\tilde{K}_{t,t'}^{L-1}(\bm{x},\bm{x}')\label{eq:iterativektilde1}
\end{equation}
with initial condition 
\begin{equation}
\tilde{K}_{t,t'}^{L=0}(\bm{x},\bm{x}')=\lambda^{-1}N_{0}^{-1}\bm{x}\cdot\bm{x}'\label{eq:initialcond}
\end{equation}

Note that in the $\lambda\rightarrow\infty$ limit both $K_{t,t'}^{L,1}$
and $\dot{K}_{t,t'}^{L}$ are independent of time. Therefore $\tilde{K}_{t,t'}^{L}(\bm{x},\bm{x}')$
is also independent of time.Thus we have
\begin{equation}
\tilde{K}^{L}(\bm{x},\bm{x}')=\lambda^{-1}K_{GP}^{L}(\bm{x},\bm{x}')+\sigma^{2}\dot{K}^{L}(\bm{x},\bm{x}')\tilde{K}^{L-1}(\bm{x},\bm{x}')\label{eq:iterativektilde}
\end{equation}

\paragraph{Relation to the Neural Tangent Kernel:}
Next, we note that the neural tangent kernel (NTK), is given by
\begin{equation}
K^{L,NTK}(\bm{x},\bm{x}')=\langle\partial_{\Theta_{random}}f(\Theta_{random},\bm{x})\cdot\partial_{\Theta_{random}}f(\Theta_{random},\bm{x}')\rangle\label{eq:ntkdef}
\end{equation}
where the average is w.r.t. Gaussian random $\Theta_{random}\sim\mathcal{N}(0,\sigma^{2}\mathbb{I})$.
We aim to show that $K^{L,NTK}(\bm{x},\bm{x}')$ obeys the same relation
as $\lambda\tilde{K}^{L}(\bm{x},\bm{x}')$, given by SI Eq. \ref{eq:iterativektilde}. To this end, we separate SI Eq. \ref{eq:ntkdef} into two parts, derivative w.r.t. the readout weights, and derivative w.r.t. the hidden-layer weights. 
\begin{itemize}
\item Derivative w.r.t. the readout weights:
\begin{equation}
\langle\partial_{\bm{a}_{random}}f(\Theta_{random},\bm{x})\cdot\partial_{\bm{a}_{random}}f(\Theta_{random},\bm{x}')\rangle=K_{GP}^{L}(\bm{x},\bm{x}')
\end{equation}
\item Derivative w.r.t. the hidden-layer weights:

Using $\left\{ h_{m}^{l}(\bm{x})\right\} _{m=1,\cdots N;l=1,\cdots,L}$
and $\left\{ x_{n}^{l}(\bm{x})\right\} _{n=1,\cdots,N;l=1,\cdots,L}$
to denote the hidden-layer pre- and post-activations with random weights
$\Theta_{random}$. By chain rule, we have 
\begin{equation}
\partial_{W_{random,ij}^{l}}x_{m}^{L}(\bm{x})=\begin{cases}
\left(N_{L-1}\right)^{-1/2}\phi'(h_{m}^{L}(\bm{x}))\sum_{n}W_{random,mn}^{L}\frac{\partial x_{n}^{L-1}(\bm{x})}{\partial W_{random,ij}^{l}} & l\leq L-1\\
\left(N_{L-1}\right)^{-1/2}\phi'(h_{m}^{L}(\bm{x}))\delta_{im}x_{j}^{L-1}(\bm{x}) & l=L
\end{cases}
\end{equation}
To the leading order
\begin{align}
 & \sum_{l=1}^{L}\langle\partial_{W_{random}^{l}}f(\Theta_{random},\bm{x})\cdot\partial_{W_{random}^{l}}f(\Theta_{random},\bm{x}')\rangle \nonumber \\
= & \langle N_{L}^{-1}\bm{a}_{random}\cdot\bm{a}_{random}\rangle\cdot N_{L}^{-1}\sum_{l=1}^{L}\langle\sum_{m,ij}\left(\partial_{W_{random,ij}^{l}}x_{m}^{L}(\bm{x})\cdot\partial_{W_{random,ij}^{l}}x_{m}^{L}(\bm{x}')\right)\rangle \nonumber \\
= & N_{L}^{-1}\sigma^{2}\sum_{l=1}^{L}\langle\sum_{m,ij}\left(\partial_{W_{random,ij}^{l}}x_{m}^{L}(\bm{x})\cdot\partial_{W_{random,ij}^{l}}x_{m}^{L}(\bm{x}')\right)\rangle\label{eq:wderivative}
\end{align}
and by plugging in SI Eq. $\text{\ref{eq:wderivative}}$
and keeping only the leading order terms
\begin{align}
 & N_{L}^{-1}\sum_{l=1}^{L}\langle\sum_{m,ij}\left(\partial_{W_{random,ij}^{l}}x_{m}^{L}(\bm{x})\cdot\partial_{W_{random,ij}^{l}}x_{m}^{L}(\bm{x}')\right)\rangle \nonumber \\
= & \sigma^{2}N_{L}^{-1}\langle\left(\phi'(h_{m}^{L}(\bm{x}))\cdot\phi'(h_{m}^{L}(\bm{x}'))\right)\rangle N_{L-1}^{-1}\sum_{l=1}^{L-1}\langle\sum_{n,ij}\left(\partial_{W_{random,ij}^{l}}x_{n}^{L-1}(\bm{x})\cdot\partial_{W_{random,ij}^{l}}x_{n}^{L-1}(\bm{x}')\right)\rangle \nonumber \\
 & +N_{L}^{-1}\langle\left(\phi'(h_{m}^{L}(\bm{x}))\cdot\phi'(h_{m}^{L}(\bm{x}'))\right)\rangle N_{L-1}^{-1}\sum_{j}x_{j}^{L-1}(\bm{x})\cdot x_{j}^{L-1}(\bm{x}')\nonumber \\
= & \dot{K}^{L}(\bm{x},\bm{x}')\left(\sigma^{2}N_{L-1}^{-1}\sum_{l=1}^{L-1}\langle\sum_{n,ij}\left(\partial_{W_{random,ij}^{l}}x_{n}^{L-1}(\bm{x})\cdot\partial_{W_{random,ij}^{l}}x_{n}^{L-1}(\bm{x}')\right)\rangle+K_{GP}^{L-1}(\bm{x},\bm{x}')\right)\label{eq:iterativederivative}
\end{align}
Denote 
\begin{equation}
\mathcal{Q}^{L-1}(\bm{x},\bm{x}')=\sigma^{2}N_{L-1}^{-1}\sum_{l=1}^{L-1}\langle\sum_{n,ij}\left(\partial_{W_{random,ij}^{l}}x_{n}^{L-1}(\bm{x})\cdot\partial_{W_{random,ij}^{l}}x_{n}^{L-1}(\bm{x}')\right)\rangle +K_{GP}^{L-1}(\bm{x},\bm{x}')
\end{equation}
and we have 
\begin{equation}
\mathcal{Q}^{L}(\bm{x},\bm{x}')=\sigma^{2}\dot{K}^{L}(\bm{x},\bm{x}')\mathcal{Q}^{L-1}(\bm{x},\bm{x}')+K_{GP}^{L}(\bm{x},\bm{x}')
\end{equation}
with initial condition 
\begin{equation}
\mathcal{Q}^{L=0}(\bm{x},\bm{x}')=N_{0}^{-1}\bm{x}\cdot\bm{x}'
\end{equation}
Therefore $\mathcal{Q}^{L}$ obeys the same iterative relation and
initial condition as $\tilde{K}^{L}\lambda$. So we have 
\begin{equation}
\mathcal{Q}^{L}(\bm{x},\bm{x}')=\tilde{K}^{L}(\bm{x},\bm{x}')\lambda
\end{equation}
We also have 
\begin{equation}
\sum_{l=1}^{L}\langle\partial_{W_{random}^{l}}f(\Theta_{random},\bm{x})\cdot\partial_{W_{random}^{l}}f(\Theta_{random},\bm{x}')\rangle=\mathcal{Q}^{L}(\bm{x},\bm{x}')-K_{GP}^{L}(\bm{x},\bm{x}')
\end{equation}

\end{itemize}

Combining the two contributions above we have
\begin{equation}
\langle\partial_{\Theta_{random}}f(\Theta_{random},\bm{x})\cdot\partial_{\Theta_{random}}f(\Theta_{random},\bm{x}')\rangle=\mathcal{Q}^{L}(\bm{x},\bm{x}')=\lambda\tilde{K}^{L}(\bm{x},\bm{x}')
\end{equation}

The relation is similar to what has been shown in \citep{avidan2023connecting},
the relevant scales of temperature $\beta^{-1}$, $\lambda$ and time
$t$ are different.Thus $\tilde{K}^{L}(\bm{x},\bm{x}')$ is time-independent,
and is equivalent to the NTK. 

\section{Multi-Head Theory}
\label{subsec:multiheadtheory}

\subsection{Summary of Main Theoretical Results}
\label{subsubsec:summarymultihead}
In multi-head CL, we consider both the mean and variance of the network's input-output mappings, for $T=2$ and $L=1$. The variance is not
negligible as in single-head CL, as it causes the divergence of $G_{2,2}$ in the overfitting regime. Since we take $L$ to be 1, in this section the $L$ superscript of the kernels is neglected. Analogous to the kernel $\tilde{K}_{t,t'}^{L}(\bm{x},\bm{x}')$
defined in SI Eq. \ref{eq:ktilde}, we introduce a new ``renormalized'' kernel for multi-head CL, using the same notation 
\begin{equation}
\tilde{K}_{2,2}(\bm{x},\bm{x}')\equiv u_{2,2}^{1}K_{2,2}^{1}(\bm{x},\bm{x}')-u_{2,2}^{0}K_{2,2}^{0}(\bm{x},\bm{x}')
\end{equation}
where the ``renormalization factors'' $u_{2,2}^{1}$ and $u_{2,2}^{0}$
can be solved self-consistently as detailed in SI
\ref{subsec:multiheadtheory}.\ref{paragraph:selfconsistenteqs}. Similarly as in SI \ref{subsec:kernels}, we introduce $\tilde{\bm{k}}_{2,2}(\bm{x})\in\mathbb{R}^{P},\tilde{\bm{K}}_{2,2}\in\mathbb{R}^{P\times P}$
and $\tilde{k}_{2,2}\in\mathbb{R}$ for this kernel function applied on the training and testing data. The expressions for the statistics of the network's input-output mappings are given below.

\subsubsection{The Mean Input-Output Mappings}

The mean input-output mappings are given by
\begin{align}
\langle f_{2}^{1}(\bm{x})\rangle & =u_{1,2}^{1}\Delta\bm{k}_{2,2}(\bm{x})^{\top}\tilde{\bm{K}}_{2,2}^{-1}\left(\bm{Y}_{2}-u_{1,2}^{1}\left(u_{1,1}^{1}\right)^{-1}\bm{K}_{2,1}^{1}\left(\bm{K}_{1,1}^{1}\right)^{-1}\bm{Y}_{1}\right)+\bm{k}_{2,1}^{1}(\bm{x})^{\top}\left(\bm{K}_{1,1}^{1}\right)^{-1}\bm{Y}_{1}\label{eq:f_21}
\end{align}
\begin{align}
\langle f_{2}^{2}(\bm{x})\rangle & =\tilde{\bm{k}}_{2,2}(\bm{x})^{\top}\tilde{\bm{K}}_{2,2}^{-1}\left(\bm{Y}_{2}-u_{1,2}^{1}\left(u_{1,1}^{1}\right)^{-1}\bm{K}_{2,1}^{1}\left(\bm{K}_{1,1}^{1}\right)^{-1}\bm{Y}_{1}\right)+u_{2,1}^{1}\left(u_{1,1}^{1}\right)^{-1}\bm{k}_{2,1}^{1}(\bm{x})^{\top}\left(\bm{K}_{1,1}^{1}\right)^{-1}\bm{Y}_{1}\label{eq:f_22}
\end{align}

\subsubsection{The Variance of the Input-Output Mappings}

The variance of the input-output mappings are 
\begin{align}
\langle\delta f_{2}^{1}(\bm{x})^{2}\rangle & =u_{1,1}^{1}k_{2,2}^{1}(\bm{x},\bm{x})-\left(u_{1,2}^{1}\right)^{2}\Delta\bm{k}_{2,2}(\bm{x})^{\top}\tilde{\bm{K}}_{2,2}^{-1}\Delta\bm{k}_{2,2}(\bm{x})-\left(u_{1,1}^{1}\bm{k}_{2,1}^{1}(\bm{x})-u_{1,2}^{1}\Delta\bm{k}_{2,2}(\bm{x})\tilde{\bm{K}}_{2,2}^{-1}\tilde{\bm{K}}_{2,1}\right)\tilde{\bm{K}}_{1,1}^{-1}\nonumber \\
 & \left(u_{1,1}^{1}\bm{k}_{2,1}^{1}(\bm{x})-u_{1,2}^{1}\Delta\bm{k}_{2,2}(\bm{x})\tilde{\bm{K}}_{2,2}^{-1}\tilde{\bm{K}}_{2,1}\right)^{\top}-2\left(u_{1,2}\right)^{2}\Delta\bm{k}_{2,2}(\bm{x})\tilde{\bm{K}}_{2,2}^{-1}\bm{k}_{2,2}^{0}(\bm{x})^{\top}\nonumber \\
 & +\left(u_{1,2}^{1}\right)^{2}u_{2,2}^{0}\Delta\bm{k}_{2,2}(\bm{x})^{\top}\tilde{\bm{K}}_{2,2}^{-1}\bm{K}_{2,2}^{0}\tilde{\bm{K}}_{2,2}^{-1}\Delta\bm{k}_{2,2}(\bm{x})\label{eq:varf21}
\end{align}
\begin{align}
\langle\delta f_{2}^{2}(\bm{x})^{2}\rangle & =u_{2,2}^{1}k_{2,2}^{1}(\bm{x},\bm{x})-\left(u_{2,2}^{1}\right)^{2}\Delta\bm{k}_{2,2}(\bm{x})^{\top}\tilde{\bm{K}}_{2,2}^{-1}\Delta\bm{k}_{2,2}(\bm{x})-\left(\tilde{\bm{k}}_{2,1}(\bm{x})-\tilde{\bm{k}}_{2,2}(\bm{x})\tilde{\bm{K}}_{2,2}^{-1}\tilde{\bm{K}}_{2,1}\right)\tilde{\bm{K}}_{1,1}^{-1}\nonumber \\
 & \left(\tilde{\bm{k}}_{2,1}(\bm{x})-\tilde{\bm{k}}_{2,2}(\bm{x})\tilde{\bm{K}}_{2,2}^{-1}\tilde{\bm{K}}_{2,1}\right)^{\top}-2u_{2,2}^{0}\tilde{\bm{k}}_{2,2}(\bm{x})\tilde{\bm{K}}_{2,2}^{-1}\bm{k}_{2,2}^{0}(\bm{x})^{\top}\nonumber \\
 & +u_{2,2}^{0}\tilde{\bm{k}}_{2,2}(\bm{x})\tilde{\bm{K}}_{2,2}^{-1}\bm{K}_{2,2}^{0}\tilde{\bm{K}}_{2,2}^{-1}\tilde{\bm{k}}_{2,2}(\bm{x})^{\top}.\label{eq:varf22}
\end{align}
These results hold for arbitrary $\lambda$. The ``renormalization factors'' $\left\{ u_{1,1}^{1},u_{1,2}^{1},u_{2,2}^{1},u_{2,2}^{0}\right\} $ are solved with SI Eqs. \ref{eq:u11}-\ref{eq:u220} in SI \ref{paragraph:selfconsistenteqs}.
Plugging them back into Eqs. \ref{eq:f_21}-\ref{eq:varf22} allows us to evaluate $F_{2,1}=\left\langle \mathcal{L}(f_{2}^{1},D_{1})\right\rangle $,
$G_{2,2}=\left\langle \mathcal{L}(f_{2}^{2},D_{2}^{test})\right\rangle $
and $G_{2,1}=\left\langle \mathcal{L}(f_{2}^{1},D_{1}^{test})\right\rangle $,
which we showed in our multi-head results.

Furthermore, in the $\lambda\rightarrow\infty$ limit, the phase-transition
boundary between the overfitting regime and the generalization regime
and the corresponding $\alpha_{c}$ (shown in Figs. \ref{fig5:ptdistractor}, \ref{fig6:phase-transition-benchmark}) is calculated by solving 
\begin{equation}
\frac{\lVert\bm{Y}_1\rVert^2}{\alpha}\bm{V}_1^\top \bm{P}_2 \bm{V}_1 -\left(\frac{\lVert\bm{Y}_1\rVert\lVert\bm{Y}_2\rVert}{\alpha}\bm{V}_1^\top\bm{V}_2\right)^2\cdot \left(\frac{\lVert\bm{Y}_1\rVert^2}{\alpha}\lVert \bm{V}_1\rVert^2\right) = u_{1,1}^{1}\left(\gamma_{\text{feature}}-\alpha^{-1/2}\right)\label{eq:ptboundary}
\end{equation}
\begin{equation}
\sigma^{-2}\left(u_{1,1}^{1}\right)^{2}-(1-\alpha)u_{1,1}^{1}-\lVert\bm{Y}_1\rVert^2\lVert\bm{V}_1\rVert^2 = 0\label{eq:u11-1}
\end{equation}
These set of equations are polynomial equations in $\alpha$, and can't be analytically solved. However, if we further assume that $\sigma^2=1$, and the data of each task is properly normalized such that $\frac{1}{P}\bm{Y}_1^\top \bm{K}_{1,1}^1 \bm{Y}_1 = 1$, thus $\lVert\bm{V}_1\rVert^2 \equiv \frac{1}{N}\bm{Y}_1^\top \bm{K}_{1,1}^1 \bm{Y}_1/\lVert\bm{Y}_1\rVert^2 = \frac{\alpha}{\lVert\bm{Y}_1\rVert^2}$, and similarly for $\bm{V}_2$. We can simplify SI Eqs. \ref{eq:ptboundary}, \ref{eq:u11-1} to
\begin{equation}
-\frac{\bm{V}_1^\top\bm{P}_2\bm{V}_1}{\lVert\bm{V}_1\rVert^2} + \cos(\bm{V}_1,\bm{V}_2)^2 + \gamma_{\text{feature}}= \alpha^{-1/2}\label{eq:simpleptboundary}
\end{equation}
as given by main text Eq. \ref{eq:gammasim}. 

For details of different solutions of the renormalization factors in the three phases, and the derivations of the phase-transition boundary see SI \ref{subsec:multiheadtheory}.\ref{paragraph:ptboundary}. Theoretical results regarding the hidden representations (Fig. \ref{fig7:optimal-lambda}(c,f)) are shown in SI \ref{subsec:multiheadtheory}.\ref{paragraph:Hidden-layer-kernels}. 

\subsection{Detailed Derivation}
\label{subsubsec:derivationmulti}

\subsubsection{Kernel Renormalization Theory}
\label{paragraph:krtheory}

In this section, we will present the detailed derivation for mapping
statistics of multi-head CL, in the thermodynamic finite-width limit
($P\rightarrow\infty,N\rightarrow\infty$). We start from the MGF
for multi-head CL, given by 
\begin{equation}
\mathcal{M}(\left\{ \ell_{T}^{task\ t}\right\} _{t=1,\cdots,T})\equiv\left[Z\left(D_{1}\right)\prod_{t=2}^{T}Z\left(\Theta_{t-1},D_{t}\right)\right]^{-1}\exp\left(-\beta E\left(\Theta_{1}|D_{1}\right)-\sum_{t=1}^{T}\beta E\left(\Theta_{t}|\Theta_{t-1},D_{t}\right)+\sum_{t=1}^{T}\ell_{t}f_{T}^{t}({\bf x})\right)\label{eq:mgfmultihead}
\end{equation}
where
\begin{equation}
E\left(\Theta_{t}|\Theta_{t-1},D_{t}\right)=\frac{1}{2}\sum_{\mu=1}^{P}\left(f_{t}^{t}\left(\bm{x}_{t}^{\mu}\right)-y_{t}^{\mu}\right)^{2}\label{eq:energy=000020fn-1} +\frac{1}{2}\beta^{-1}\sigma^{-2}\left\Vert \Theta_{t}\right\Vert ^{2}+\frac{1}{2}\beta^{-1}\lambda\left\Vert \mathcal{W}_{t}-\mathcal{W}_{t-1}\right\Vert ^{2}.
\end{equation}
\begin{equation}
E\left(\Theta_{1}|D_{1}\right)=\frac{1}{2}\sum_{\mu=1}^{P}\left(f_{1}^{1}\left(\bm{x}_{1}^{\mu}\right)-y_{1}^{\mu}\right)^{2}+\frac{1}{2}\beta^{-1}\sigma^{-2}\left\Vert \Theta_{1}\right\Vert ^{2}
\end{equation}
and 
\begin{equation}
Z\left(\Theta_{t-1},D_{t}\right)\equiv\int d\Theta_{t}\exp(-\beta E\left(\Theta_{t}|\Theta_{t-1},D_{t}\right))
\end{equation}
\begin{equation}
Z\left(D_{1}\right)\equiv\int d\Theta_{1}\exp(-\beta E\left(\Theta_{1}|D_{1}\right))
\end{equation}
Here we introduce fields $\ell_{t}$ coupled to each mapping $f_{T}^{t}({\bf x})$
after learning the $T$-th task, the statistics of $f_{T}^{t}({\bf x})$
can therefore be calculated by 
\begin{equation}
\langle f_{T}^{t}({\bf x})\rangle=\frac{\partial\mathcal{M}(\left\{ \ell_{t}\right\} _{t=1,\cdots,T})}{\partial\ell_{t}}\lvert_{\left\{ \ell_{t}\right\} _{t=1,\cdots,T}=0}\label{eq:meanpredictor}
\end{equation}
\begin{equation}
\langle\delta^{2}f_{T}^{t}({\bf x})\rangle=\frac{\partial^{2}\mathcal{M}(\left\{ \ell_{t}\right\} _{t=1,\cdots,T})}{\partial(\ell_{t})^{2}}\lvert_{\left\{ \ell_{t}\right\} _{t=1,\cdots,T}=0}\label{eq:variancepredictor}
\end{equation}
Similarly as in single-head CL, we use the replica method for the
denominator in SI Eq. $\ref{eq:mgfmultihead}$,
and introduce auxilliary integration variable $\left\{ \bm{v}_{t}^{\alpha}\right\} _{t=2,\cdots,T;\alpha=1,\cdots n},\bm{v}_{1}^{n}\in\mathbb{R}^{P}$
using the H-S transform, and arrive at
\begin{align}
\mathcal{M}\left(\left\{ \ell_{T}^{task\ t}\right\} _{t=1,\cdots,T}\right) &=\lim_{n\rightarrow0}\intop\prod_{\alpha=1}^{n}\prod_{t=2}^{T}d\bm{v}_{t}^{\alpha}\int d\bm{v}_{1}^{n}\int\prod_{\alpha=1}^{n}\prod_{t=2}^{T}d\Theta_{t}^{\alpha}\int d\Theta_{1}^{n}\nonumber \\
 & \exp\Bigg[-i\sum_{t,\alpha}\sum_{\mu=1}^{P}v_{t}^{\alpha,\mu}\left(f_{t}^{t}\left(\Theta_{t}^{\alpha},\bm{x}_{t}^{\mu}\right)-y_{t}^{\mu}\right) -\frac{1}{2}\sigma^{-2}\sum_{t,\alpha}\left\Vert \Theta_{t}^{\alpha}\right\Vert ^{2}-\frac{1}{2}\lambda\sum_{t=2}^{T}\sum_{\alpha=1}^{n}\left\Vert \mathcal{W}_{t}^{\alpha}-\mathcal{W}_{t-1}^{n}\right\Vert ^{2}\nonumber \\
 & -\frac{1}{2}\beta^{-1}\sum_{t,\alpha}\bm{v}_{t}^{\alpha\top}\bm{v}_{t}^{\alpha}+\sum_{t=1}^{T}\ell_{t}f_{T}^{t}(\bm{a}_{t}^{n},\mathcal{W}_{T}^{n},{\bf x})\Bigg] 
\end{align}
where we use $f_{t}^{t}\left(\Theta_{t}^{\alpha},\bm{x}_{t}^{\mu}\right)$
to denote the mapping with the replicated $\Theta_{t}^{\alpha},$$f_{t}^{t}(\Theta_{t}^{\alpha},{\bf x}_{t}^{\mu})\equiv\frac{1}{\sqrt{N}}\bm{a}_{t}^{\alpha}\Phi(\mathcal{W}_{t}^{\alpha},{\bf x}_{t}^{\mu})$;
and we use $f_{T}^{t}(\bm{a}_{t}^{n},\mathcal{W}_{T}^{n},{\bf x})$
to denote the $t$-th mapping after learning all $T$ tasks with physical
parameters $\left\{ \bm{a}_{t}^{n},\mathcal{W}_{T}^{n}\right\} $,
on an arbitrary test input ${\bf x}$.Similar to single-head CL, we
define $\sum_{t,\alpha}(\cdot)_{t,\alpha}\equiv(\cdot)_{1,n}+\sum_{t=2}^{T}\sum_{\alpha=1}^{n}(\cdot)_{t,\alpha}.$

We then integrate the readout weights $\left\{ \bm{a}_{t}^{\alpha}\right\} _{t=1,\cdots T;\alpha=1,\cdots n}$,
and obtain in the $\beta\rightarrow\infty$ limit
\begin{equation}
 \mathcal{M}\left(\left\{ \ell_{t}\right\} _{t=1,\cdots,T}\right) =  \lim_{n\rightarrow0}\intop\prod_{\alpha=1}^{n}\prod_{t=1}^{T}d\bm{v}_{t}^{\alpha}\exp\Bigg(-\frac{1}{2}\beta^{-1}\sum_{t,\alpha}\bm{v}_{t}^{\alpha\top}\bm{v}_{t}^{\alpha}+i\sum_{t,\alpha}\bm{v}_{t}^{\alpha\top}\bm{Y}_{t} +G(\left\{ \tilde{\bm{v}}_{t}^{\alpha}\right\} _{\alpha=1,\cdots,n;t=2,\cdots,T},\tilde{\bm{v}}_{1}^{n})\Bigg)\label{eq:multiheadmgfwithG}
\end{equation}
where 
\begin{equation}
G(\left\{ \tilde{\bm{v}}_{t}^{\alpha}\right\} _{\alpha=1,\cdots,n;t=2,\cdots,T},\tilde{\bm{v}}_{1}^{n}) \equiv \log\left[\intop\prod_{\alpha=1}^{n}\prod_{t=1}^{T}d\mathcal{W}_{t}^{\alpha}\exp\left(S_{0}\left(\left\{ \mathcal{W}^{\alpha}\right\} _{\alpha=1,\cdots,n}\right)-\frac{\sigma^{2}}{2}\sum_{t,\alpha}\tilde{\bm{v}}_{t}^{\alpha\top}\bm{M}_{t,t}^{\alpha\alpha}\tilde{\bm{v}}_{t}^{\alpha}\right)\right]
\end{equation}
and $S_{0}(\left\{ \mathcal{W}^{\alpha}\right\} _{\alpha=1,\cdots,n})$
and $\bm{M}$ are defined in the same way as in SI Eqs. \ref{eq:prior}, \ref{eq:M}. The only difference is that only
diagonal elements of $\bm{M}$ appear in $G$.For simplicity, here
we denote $\tilde{\bm{v}}_{t}^{\alpha}\equiv[\bm{v}_{t}^{\alpha},\delta_{\alpha n}\ell_{t}]\in\mathbb{R}^{P+1}$,
absorbing the fields coupled to the mappings on arbitrary ${\bf x}$
into $\bm{v}_{t}^{\alpha}$. It is still highly nontrivial to integrate
the hidden-layer weights and compute $G(\left\{ \tilde{\bm{v}}_{t}^{\alpha}\right\} _{\alpha=1,\cdots,n;t=2,\cdots,T},\tilde{\bm{v}}_{1}^{n})$
in general. 

\paragraph{Infinite-width Limit:}
In the infinite-width limit, the distribution of $\left\{ \mathcal{W}^{\alpha}\right\} _{\alpha=1,\cdots,n}$
is dominated by the prior determined by $S_{0}(\left\{ \mathcal{W}^{\alpha}\right\} _{\alpha=1,\cdots,n})$.
Therefore, $G$ can be calculated by integrating over Gaussian $\left\{ \mathcal{W}^{\alpha}\right\} _{\alpha=1,\cdots,n}$,
resulting in 
\begin{equation}
G(\left\{ \tilde{\bm{v}}_{t}^{\alpha}\right\} _{\alpha=1,\cdots,n;t=2,\cdots,T},\tilde{\bm{v}}_{1}^{n})=-\frac{\sigma^{2}}{2}\sum_{t,\alpha}\tilde{\bm{v}}_{t}^{\alpha\top}\langle\bm{M}_{t,t}^{\alpha\alpha}\rangle_{\mathcal{W}}\tilde{\bm{v}}_{t}^{\alpha}
\end{equation}
Compared to SI Eq. $\text{\ref{eq:Gsinglehead}}$, there
is no coupling between different replica indices, and no coupling between different time indices, which allows us to get rid of the
replica easily. Using SI Eqs. \ref{eq:averagedM}-\ref{eq:testdatascalar}, we have, in the $\beta\rightarrow\infty$ limit,

\begin{align}
\mathcal{M}\left(\left\{ \ell_{t}\right\} _{t=1,\cdots,T}\right) &= \int\prod_{t=1}^{T}d\bm{v}_{t}\exp\Bigg(i\sum_{t=1}^{T}\bm{v}_{t}^{\top}\bm{Y}_{t}-\frac{\sigma^{2}}{2}\sum_{t=1}^{T}\bm{v}_{t}^{\top}\bm{K}_{t,t}^{L,1}\bm{v}_{t} -\frac{\sigma^{2}}{2}\left(\ell_{t}\right)^{2}k_{T,T}^{L,1}(\bm{x},\bm{x})-\sigma^{2}\sum_{t=1}^{T}\ell_{t}k_{T,t}^{L,1}(\bm{x})\bm{v}_{t}\Bigg)\\
&=\exp\Bigg(-\frac{\sigma^{2}}{2}\left(\ell_{t}\right)^{2}k_{T,T}^{L,1}(\bm{x},\bm{x})-\frac{\sigma^{-2}}{2}\sum_{t=1}^{T}(\bm{Y}_{t}+\sigma^{2}i\ell_{t}\bm{k}_{T,t}^{L,1}(\bm{x}))\left(\bm{K}_{t,t}^{L,1}\right)^{-1}(\bm{Y}_{t}+i\ell_{t}\sigma^{2}\bm{k}_{T,t}^{L,1}(\bm{x}))^{\top}\Bigg)
\end{align}

The mapping statistics are then simply given by 
\begin{equation}
\langle f_{T}^{t}(\bm{x})\rangle=\bm{k}_{T,t}^{L,1}(\bm{x})\left(\bm{K}_{t,t}^{L,1}\right)^{-1}\bm{Y}_{t}
\end{equation}
and 
\begin{equation}
\langle\delta f_{T}^{t}(\bm{x})^{2}\rangle=\sigma^{2}\left(k_{T,T}^{L,1}(\bm{x},\bm{x})-\bm{k}_{T,t}^{L,1}(\bm{x})\left(\bm{K}_{t,t}^{L,1}\right)^{-1}\bm{k}_{T,t}^{L,1}(\bm{x})^{\top}\right)
\end{equation}

Therefore, the infinite-width limit in multi-head CL is trivial. The
mapping statistics are identical as learning a single task, where
the readout weights are learned with hidden-layer weights $\mathcal{W}_{t}$.
There is no coupling between different tasks induced by the learning
of hidden-layer weights. As a result, we focus on the thermodynamic
finite-width limit, where the hidden-layer weights become task-relevant,
and induce interactions between different tasks during CL. 

\paragraph{Thermodynamic Finite-Width Limit:}
We focus on the thermodynamic finite-Width limit, and use the kernel
renormalization approach as in \citep{li2021statistical} to derive
the mapping statistics in this regime for networks with a \textit{single
hidden-layer.} First, for $L=1$, using $\bm{w}_{t}^{\alpha}$ to
denote a single row of $W_{t}^{1,\alpha}$, we have 
\begin{equation}
G(\left\{ \tilde{\bm{v}}_{t}^{\alpha}\right\} _{\alpha=1,\cdots,n;t=2,\cdots,T},\tilde{\bm{v}}_{1}^{n}) = N\log\left[\intop\prod_{\alpha=1}^{n}\prod_{t=1}^{T}d\bm{w}_{t}^{\alpha}\exp\left(S_{0}\left(\bm{w}\right)-\frac{\sigma^{2}}{2}\sum_{t,\alpha}\tilde{\bm{v}}_{t}^{\alpha\top}\bm{M}_{t,t}^{\alpha\alpha,w}\tilde{\bm{v}}_{t}^{\alpha}\right)\right]
\end{equation}
where $\bm{M}_{t,t}^{\alpha\alpha,w}$ is defined similarly as SI
Eq. \ref{eq:M}, but replacing the $P\times N$ matrix $\Phi(\mathcal{W}_{t}^{\alpha},{\bf x}_{t})$
and $N$ dimensional vector $\Phi(\bm{w}_{t}^{\alpha},{\bf x})$ ,
and replacing the $P$ dimensional vector $\Phi(\mathcal{W}_{t}^{\alpha},{\bf x})$
with a scalar $\Phi(\bm{w}_{t}^{\alpha},{\bf x})$, and scaled by
$N$ to keep the elements $\mathcal{O}(1)$. $S_{0}(\bm{w})$ is also
defined similarly as in SI Eq. \ref{eq:prior},
by replacing $\left\{ \mathcal{W}_{t}^{\alpha}\right\} _{t=1,\cdots,T;\alpha=1,\cdots,n}$
with the row vectors $\left\{ \bm{w}_{t}^{\alpha}\right\} _{t=1,\cdots,T;\alpha=1,\cdots,n}$.
Furthermore, we adopt the Gaussian approximation equivalent to \citep{li2021statistical},
such that 
\begin{equation}
z_{t}^{\alpha}=N^{-1/2}\sigma\bm{v}_{t}^{\alpha\top}\Phi(\bm{w}_{t}^{\alpha},{\bf x}_{t})+N^{-1/2}\sigma\delta_{\alpha n}\ell^{task\ t}\Phi(\bm{w}_{T}^{\alpha},{\bf x})
\end{equation}
is Gaussian with $\langle z_{t}^{\alpha}\rangle=0$ and $\langle z_{t}^{\alpha}z_{t'}^{\beta}\rangle=\sigma^{2}\frac{1}{N}\tilde{\bm{v}}_{t}^{\alpha\top}\langle\bm{M}_{t,t'}^{\alpha\beta,w}\rangle_{w}\tilde{\bm{v}}_{t'}^{\beta}$where
the average $\langle\cdot\rangle_{w}$ is w.r.t. the prior Gaussian
distribution in $\bm{w}$, whose probability density function is proportional
to $\exp(-S_{0}(\bm{w}))$. Therefore, we replace the integral over
$\bm{w}_{t}^{\alpha}$ with a Gaussian integral over $z_{t}^{\alpha}$,
and introducing $H_{t,t',\alpha,\beta}=\sigma^{2}\frac{1}{N}\tilde{\bm{v}}_{t}^{\alpha\top}\langle\bm{M}_{t,t'}^{\alpha\beta,w}\rangle_{w}\tilde{\bm{v}}_{t'}^{\beta}$,
we have 
\begin{align}
 & G(\left\{ \tilde{v}_{t}^{\alpha}\right\} _{\alpha=1,\cdots,n;t=1,\cdots,T})=G(\bm{H})\nonumber\\
= & N\log\left[\intop\prod_{\alpha=1}^{n}\prod_{t=1}^{T}dz_{t}^{\alpha}\det(\bm{H})^{-1/2}\exp\left(-\frac{1}{2}\sum_{t,\alpha}\sum_{t',\beta}z_{t}^{\alpha\top}[\bm{H}^{-1}]_{t,t',\alpha\beta}z_{t'}^{\beta}-\frac{1}{2}\sum_{t,\alpha}(z_{t}^{\alpha})^{2}\right)\right] \nonumber\\
= & -\frac{N}{2}\log\det(\mathbb{I}+\bm{H})\label{eq:GH}
\end{align}
Plugging into SI Eq.  \ref{eq:multiheadmgfwithG},
and using the Fourier representation of the Dirac delta function to
introduce $\bm{H}$,i.e., 
\[
\delta(\sigma^{-2}H_{t,t',\alpha,\beta}-\frac{1}{N}\tilde{\bm{v}}_{t}^{\alpha\top}\langle\bm{M}_{t,t'}^{\alpha\beta,w}\rangle_{w}\tilde{\bm{v}}_{t'}^{\beta})=\int d\bm{U}\exp(\sum_{t,t',\alpha,\beta}U_{t,t',\alpha,\beta}(\sigma^{-2}H_{t,t',\alpha,\beta}-\frac{1}{N}\tilde{\bm{v}}_{t}^{\alpha\top}\langle\bm{M}_{t,t'}^{\alpha\beta,w}\rangle_{w}\tilde{\bm{v}}_{t'}^{\beta}))
\]
, we have 

\begin{align}
\mathcal{M}\left(\left\{ \ell_{t}\right\} _{t=1,\cdots,T}\right) &= \lim_{n\rightarrow0}\intop\prod_{\alpha=1}^{n}\prod_{t=1}^{T}d\bm{v}_{t}^{\alpha}\int d\bm{U}\int d\bm{H} \exp(-\frac{1}{2}\beta^{-1}\sum_{t,\alpha}\bm{v}_{t}^{\alpha\top}\bm{v}_{t}^{\alpha}+i\sum_{t,\alpha}\bm{v}_{t}^{\alpha\top}\bm{Y}_{t}-\frac{N}{2}\log\det(\mathbb{I}+\bm{H})\nonumber \\
& +\sigma^{-2}\frac{N}{2}\mathrm{Tr}(\bm{U}\bm{H})-\frac{1}{2}\sum\sum_{t',\beta}U_{t,t',\alpha,\beta}\tilde{\bm{v}}_{t}^{\alpha\top}\langle\bm{M}_{t,t'}^{\alpha\beta,w}\rangle_{w}\tilde{\bm{v}}_{t'}^{\beta})\label{eq:mgfUH}
\end{align}
In the limit $N\rightarrow\infty$, $P\rightarrow\infty$, $\alpha=P/N\sim\mathcal{O}(1)$.
The exponent of $\mathcal{M}(\ell)$ scales with $N$, allowing us
to adopt the saddle-point approximation for the integral over $H$
and $U$. Taking derivative of the exponent w.r.t. $H$ and set it
to $0$, we obtain $\sigma^{-2}\bm{U}=(\mathbb{I}+\bm{H})^{-1}$,plugging
back in, we have 

\begin{align}
\mathcal{M}\left(\left\{ \ell_{t}\right\} _{t=1,\cdots,T}\right) &= \lim_{n\rightarrow0}\intop\prod_{\alpha=1}^{n}\prod_{t=1}^{T}d\bm{v}_{t}^{\alpha}\int d\bm{U} \exp\Bigg[-\frac{1}{2}\beta^{-1}\sum_{t,\alpha}\bm{v}_{t}^{\alpha\top}\bm{v}_{t}^{\alpha}+i\sum_{t,\alpha}\bm{v}_{t}^{\alpha\top}\bm{Y}_{t}+\frac{N}{2}\log\det(\bm{U})\nonumber \\
&-\sigma^{-2}\frac{N}{2}\mathrm{Tr}(\bm{U})-\frac{1}{2}\sum_{t,\alpha}\sum_{t',\beta}U_{t,t',\alpha,\beta}\tilde{\bm{v}}_{t}^{\alpha\top}\langle\bm{M}_{t,t'}^{\alpha\beta,w}\rangle_{w}\tilde{\bm{v}}_{t'}^{\beta}\Bigg]\label{eq:mgfreplica}
\end{align}

To proceed from SI Eq. $\text{\ref{eq:mgfreplica}}$, we
take the replica symmetric ansatz, analogous to calculation for the
Franz-Parisi potential\citep{franz1995recipes}, for $t,t'\geq2$
\begin{equation}
U_{t,t',\alpha,\beta}=\begin{cases}
u_{t,t'}^{1} & \left\{ \alpha=\beta,\tau=\tau^{\prime}\right\} \cup\left\{ \alpha=n,\tau<\tau^{\prime}\right\} \cup\left\{ \beta=n,\tau>\tau^{\prime}\right\} \\
u_{t,t'}^{0} & otherwise
\end{cases}
\end{equation}
otherwise denote 
\begin{equation}
U_{t,1,\alpha,n}=u_{t,1}^{1};u_{t,1}^{0}\equiv0
\end{equation}
Furthermore, $\langle\bm{M}_{t,t'}^{\alpha\beta,w}\rangle_{w}$ can
be written in the same way as SI Eqs. \ref{eq:averagedM}-\ref{eq:testdatascalar},
thus getting rid of the replica indices in $\bm{M}$. Analogous to
SI Eq. $\text{\ref{eq:generalized-ntk}}$, we introduce
\begin{equation}
\tilde{K}_{t,t'}(\bm{x},\bm{x}')=u_{t,t'}^{1}K_{t,t'}^{L,1}(\bm{x},\bm{x}')-u_{t,t'}^{0}K_{t,t'}^{L,0}(\bm{x},\bm{x}')
\end{equation}
and correspondingly, $\tilde{\bm{K}}_{t,t'}^{L}=u_{t,t'}^{1}\bm{K}_{t,t'}^{L,1}-u_{t,t'}^{0}\bm{K}_{t,t'}^{L,0}$.
$\tilde{\bm{k}}_{t,t'}^{L}(\bm{x})=u_{t,t'}^{1}\bm{k}_{t,t'}^{L,1}(\bm{x})-u_{t,t'}^{0}\bm{k}_{t,t'}^{L,0}(\bm{x})$.We
also introduce $\Delta\bm{k}_{t,t'}(\bm{x})=\bm{k}_{t,t'}^{L,1}(\bm{x})-\bm{k}_{t,t'}^{L,0}(\bm{x})$.
Note that for multi-head CL, the theoretical results are for $L=1$,
we thus neglect the $L$ index hereafter. 

$\mathcal{M}(\ell)$ is quadratic in $\bm{v}_{t}^{\alpha}$, in principle,
it allows us to perform integration over $\bm{v}_{t}^{\alpha}$ for
arbitrary $T$ in a similar way as in SI \ref{subsec:singleheadtheory}.\ref{subsec:predictor-statistics}.
By introducing $\bm{p}_{t}=\sum_{\alpha=1}^{n}\bm{v}_{t}^{\alpha}$
and integrating over $\left\{ \bm{v}_{t}^{\alpha}\right\} _{\alpha=1,\cdots,n-1;t=1,\cdots,T}$, letting $\bm{v}_{t}\equiv\bm{v}_{t}^{n}$, we arrive at 

\begin{align}
\mathcal{M}\left(\left\{ \ell_{t}\right\} _{t=1,\cdots,T}\right) &= \int d\bm{U}\int\prod_{t=2}^{T}d\bm{p}_{t}\int\prod_{t=1}^{T}d\bm{v}_{t}\exp\Bigg[\frac{N}{2}\log\det(\bm{U})\nonumber \\
& -\sigma^{-2}\frac{N}{2}\mathrm{Tr}(\bm{U})-\frac{1}{2}\sum_{t=2}^{T}\bm{p}_{t}^{\top}(u_{t,t}^{0}\bm{K}_{t,t}^{0}-\tilde{\bm{K}}_{t,t})\bm{p}_{t}-\frac{1}{2}u_{1,1}^{1}\bm{v}_{1}^{\top}\bm{K}_{1,1}^{1}\bm{v}_{1}\nonumber \\
& -\sum_{t=t'}^{T}\sum_{t'=2}^{T}\bm{p}_{t}^{\top}\tilde{\bm{K}}_{t,t'}\bm{v}_{t'}-\sum_{t=t'+1}^{T}\sum_{t'=2}^{T}u_{t,t'}^{0}\bm{p}_{t}^{\top}\bm{K}_{t,t'}^{0}\bm{p}_{t'}-\sum_{t=2}^{T}u_{t,1}^{1}\bm{v}_{1}^{\top}\bm{K}_{1,t}^{1}\bm{p}_{t}+i\sum_{t=2}^{T}\bm{p}_{t}^{\top}\bm{Y}_{t}+i\bm{v}_{1}^{\top}\bm{Y}_{1}\nonumber \\
& -(n-2)\sum_{t=2}^{T}\log\det\tilde{\bm{K}}_{t,t}+\frac{n}{2}\sum_{t=2}^{T}\bm{p}_{t}^{\top}\tilde{\bm{K}}_{t,t}\bm{p}_{t}-n\sum_{t=2}^{T}\bm{p}_{t}^{\top}\tilde{\bm{K}}_{t,t}\bm{v}_{t}+\frac{n}{2}\sum_{t=1}^{T}\bm{v}_{t}^{\top}\tilde{\bm{K}}_{t,t}\bm{v}_{t}\nonumber \\
& -\sum_{t=1}^{T}\sum_{t'=t+1}^{T}u_{t,t'}^{1}\ell_{t}\Delta\bm{k}_{T,t'}(\bm{x})\bm{v}_{t'}-\sum_{t=1}^{T}\sum_{t'=1}^{t}\ell_{t}[u_{t,t'}^{1}\bm{k}_{T,t'}^{1}(\bm{x})-u_{t,t'}^{0}k_{T,t'}^{0}(\bm{x})]\bm{v}_{t'}\nonumber \\
& -\sum_{t=1}^{T}\sum_{t'=t+1}^{T}u_{t,t'}^{1}\ell_{t}\bm{k}_{T,t'}^{0}(\bm{x})\bm{p}_{t'}-\sum_{t=1}^{T}\sum_{t'=2}^{t}u_{t,t'}^{0}\ell_{t}\bm{k}_{T,t'}^{0}(\bm{x})\bm{p}_{t'} -\sum_{t,t'=1}^{T}k_{T,T}^{1}(\bm{x},\bm{x})u_{t,t'}^{1}\ell_{t}\ell_{t'}\Bigg]\label{eq:mgfmulti}
\end{align}

First, we note that the structure of SI Eq. \ref{eq:mgfmulti}
is quite similar to SI Eq. \ref{eq:mgfsingle}.
Comparing the two equations, there are several major differences.
One is that the factors $\left\{ m_{t,t'}^{1}\right\} _{t,t'=1,\cdots,T}$
and $\left\{ m_{t,t'}^{0}\right\} _{t,t'=1,\cdots,T}$ in SI
Eq. \ref{eq:mgfsingle} are determined by the prior distribution
in the readout weights $\bm{a}_{t}$, the renormalization factors
$\left\{ u_{t,t'}^{1}\right\} _{t,t'=1,\cdots,T}$ and $\left\{ u_{t,t'}^{0}\right\} _{t,t'=1,\cdots,T}$
incorporate the effect of learning the data, and need to be solved
self-consistently. The other difference is in the terms containing
the external field $\left\{ \ell_{t}\right\} _{t=1,\cdots,T}$, while
in single-head CL we only introduce one external field $\ell$, in
multi-head CL we introduce $T$ external fields each coupled to one
mapping. Furthermore, in multi-head CL, the mappings utilize $\bm{a}_{t}$
and $\mathcal{W}_{T}$, at different times, some kernels are renormalized
by renormalization factors with different time indices. As an example,
we see kernels with time indices $t,T$ renormalized by renormalization
factors with time indices $t$ and $t'$. In single-head CL, however,
kernels with time indices $t$ and $t'$ always appear with $m_{t,t'}^{1}$
or $m_{t,t'}^{0}$.Finally, we stress that unlike in SI Eq. \ref{eq:mgfsingle} in SI \ref{subsec:singleheadtheory}.\ref{paragraph:predictor-statistics},
we keep also the $\mathcal{O}(n)$ terms, as we will see later, these
terms will determine the self-consistent equations in the renormalization
factors. 

\subsubsection{Statistics of the Input-Output Mappings for Arbitrary Number of Tasks $T$}

\label{paragraph:multiheadpredictor}

The statistics of input-output mappings can be evaluated for arbitrary
$T$, by taking derivative of $\mathcal{M}(\ell)$ w.r.t. $\ell_{t}$.
Using SI Eqs. $\text{\ref{eq:meanpredictor}, \ref{eq:variancepredictor}}$,
we have 

\paragraph{Mean:}
The mean mappings are given by
\begin{equation}
\langle f_{T}^{t}(\bm{x})\rangle=\sum_{t'=t+1}^{T}u_{t,t'}^{1}\Delta\bm{k}_{T,t'}(\bm{x})^{\top}\langle-i\bm{v}_{t'}\rangle+\sum_{t'=1}^{t}[u_{t,t'}^{1}\bm{k}_{T,t'}^{L,1}(\bm{x})-u_{t,t'}^{0}k_{T,t'}^{L,0}(\bm{x})]^{\top}\langle-i\bm{v}_{t'}\rangle\label{eq:meanpred}
\end{equation}

\paragraph{Variance:}
\begin{align}
\langle\delta^{2}f_{T}^{t}(\bm{x})\rangle &=u_{t,t}^{1}k_{T,T}^{1}(\bm{x},\bm{x})-\sum_{\tau,\tau'=t+1}^{T}\Delta\bm{k}_{T,\tau}(\bm{x})u_{t,\tau}^{1}u_{t,\tau'}\langle\delta\bm{v}_{\tau}\delta\bm{v}_{\tau'}^{\top}\rangle\Delta\bm{k}_{T,\tau'}(\bm{x})^{\top}\nonumber \\
& -\sum_{\tau,\tau'=1}^{t}[u_{t,\tau}^{1}\bm{k}_{T,\tau}^{1}(\bm{x})-u_{t,\tau}^{0}\bm{k}_{T,\tau}^{0}(\bm{x})]\langle\delta\bm{v}_{\tau}\delta\bm{v}_{\tau'}^{\top}\rangle[u_{t,\tau'}^{1}\bm{k}_{T,\tau'}^{1}(\bm{x})-u_{t,\tau'}^{0}\bm{k}_{T,\tau'}^{0}(\bm{x})]^{\top}\nonumber \\
& -2\sum_{\tau=1}^{t}\sum_{\tau'=t+1}^{T}u_{t,\tau'}^{1}[u_{t,\tau}^{1}\bm{k}_{T,\tau}^{1}(\bm{x})-u_{t,\tau}^{0}\bm{k}_{T,\tau}^{0}(\bm{x})]\langle\delta\bm{v}_{\tau}\delta\bm{v}_{\tau'}^{\top}\rangle\Delta\bm{k}_{T,\tau'}(\bm{x})^{\top}\nonumber \\
&-2\sum_{\tau=t+1}^{T}\sum_{\tau'=t+1}^{\tau}u_{t,\tau}^{1}u_{t,\tau'}^{1}\Delta\bm{k}_{T,\tau}(\bm{x})\langle\delta\bm{v}_{\tau}\delta\bm{p}_{\tau'}^{\top}\rangle\bm{k}_{T,\tau'}^{0}(\bm{x})^{\top}\nonumber \\
& -2\sum_{\tau=t+1}^{T}\sum_{\tau'=2}^{t}u_{t,\tau}^{1}u_{t,\tau'}^{0}\Delta\bm{k}_{T,\tau}(\bm{x})\langle\delta\bm{v}_{\tau}\delta\bm{p}_{\tau'}^{\top}\rangle\bm{k}_{T,\tau'}^{0}(\bm{x})^{\top}\nonumber \\
 & -2\sum_{\tau=1}^{t}\sum_{\tau'=1}^{\tau}u_{t,\tau'}^{0}[u_{t,\tau}^{1}\bm{k}_{T,\tau}^{1}(\bm{x})-u_{t,\tau}^{0}k_{T,\tau}^{0}(\bm{x})]\langle\delta\bm{v}_{\tau}\delta\bm{p}_{\tau'}^{\top}\rangle\bm{k}_{T,\tau'}^{0}(\bm{x})^{\top}\label{eq:varpred}
\end{align}
The first and second moments of $\left\{ \bm{v}_{t},\bm{p}_{t}\right\} _{t=2}^{T}$
and $\bm{v}_{1}$ are the same as in SI \ref{subsec:multiheadtheory}.\ref{paragraph:predictor-statistics},
given by SI Eqs. \ref{eq:meanv}, \ref{eq:covpv}, \ref{eq:covv}, but replacing the factors $\left\{ m_{t,t'}^{1}\right\} _{t,t'=1,\cdots,T}$
, $\left\{ m_{t,t'}^{0}\right\} _{t,t'=1,\cdots,T}$ with $\left\{ u_{t,t'}^{1}\right\} _{t,t'=1,\cdots,T}$
and $\left\{ u_{t,t'}^{0}\right\} _{t,t'=1,\cdots,T}$. Taking $T=2$
and $t=1,2$, we can obtain the results presented in SI \ref{subsec:multiheadtheory}.\ref{subsubsec:summarymultihead}. 

To complete the calculation, we need to compute the renormalization
factors $\left\{ u_{t,t'}^{1}\right\} _{t,t'=1,\cdots,T}$ and $\left\{ u_{t,t'}^{0}\right\} _{t,t'=1,\cdots,T}$.
They obey self-consistent equations derived from saddle-point approximation
of the integrals in SI Eq. \ref{eq:mgfmulti} w.r.t. $\bm{U}$. The self-consistent equations for general $T$ are complicated
for the following reasons. First, using the replica symmetry ansatz,
explicitly writing down $\log\det\bm{U}$ as a function of $\left\{ u_{t,t'}^{1}\right\} _{t,t'=1,\cdots,T}$
and $\left\{ u_{t,t'}^{0}\right\} _{t,t'=1,\cdots,T}$ for arbitrary
$T$ is complicated. Second, the total number of renormalization factors
grow quadratically with the number of tasks. For $T$ tasks, there
are a total of $T^{2}$ renormalization factors. In the scope of this
paper, we will only present the self-consistent equations for up to
$T=3$ in SI \ref{subsec:multi-head}.\ref{paragraph:selfconsistenteqs},
which we use in our main results.

\subsubsection{Self-Consistent Equations for the Renormalization Factors Up to $T=3$}

\label{paragraph:selfconsistenteqs}

In this section, we start from SI Eq. \ref{eq:mgfmulti} and present the detailed derivation of the self-consistent equations
for the renormalization factors for $T=2$. For $T=3$, we present the final result of the effective Hamiltonian used to derive the self-consistent equations on the renormalization factors, the detailed derivation is similar as for $T=2$.

\paragraph{Self-Consistent Equations for $T=2:$}

The terms coupled to the external field $\left\{ \ell_{t}\right\} _{t=1,\cdots,T}$
do not impact the self-consistent equation, and we may neglect them
for simplicity, and for $T=2$, SI Eq. $\text{\ref{eq:mgfmulti}}$
simplifies to

\begin{align}
\mathcal{M}\left(\left\{ \ell_{t}\right\} _{t=1,2}=0\right)&=\int d\bm{U}\int d\bm{p}_{2}\int\prod_{t=1}^{2}d\bm{v}_{t}\exp\Bigg[\frac{N}{2}\log\det(\bm{U})-\sigma^{-2}\frac{N}{2}\mathrm{Tr}(\bm{U})\nonumber -\frac{1}{2}\bm{p}_{2}^{\top}(u_{2,2}^{0}\bm{K}_{2,2}^{0}-\tilde{\bm{K}}_{2,2})\bm{p}_{2}-\frac{1}{2}\bm{v}_{1}^{\top}\tilde{\bm{K}}_{1,1}\bm{v}_{1}\nonumber \\
& -\bm{p}_{2}^{\top}\tilde{\bm{K}}_{2,2}\bm{v}_{2}-\bm{v}_{1}^{\top}\tilde{\bm{K}}_{1,2}\bm{p}_{2}+i\bm{p}_{2}^{\top}\bm{Y}_{2}+i\bm{v}_{1}^{\top}\bm{Y}_{1} -\frac{1}{2}(n-2)\log\det\tilde{\bm{K}}_{2,2} \nonumber \\
& +\frac{n}{2}\bm{p}_{2}^{\top}\tilde{\bm{K}}_{2,2}\bm{p}_{2}-n\bm{p}_{2}^{\top}\tilde{\bm{K}}_{2,2}\bm{v}_{2}+\frac{n}{2}\sum_{t=1}^{2}\bm{v}_{t}^{\top}\tilde{\bm{K}}_{t,t}\bm{v}_{t}\Bigg]\label{eq:mgfmulti-1}
\end{align}
First, note that for $T=2$, we can derive explicit form of $\log\det(\bm{U})$
in terms of $\left\{ u_{t,t'}^{1}\right\} _{t,t'=1,\cdots,T}$ and
$\left\{ u_{t,t'}^{0}\right\} _{t,t'=1,\cdots,T}$, as 
\begin{equation}
\log\det(\bm{U})=\log u_{1,1}^{1}+n\log\left(u_{2,2}^{1}-u_{2,2}^{0}\right)+n\left(u_{2,2}^{1}-u_{2,2}^{0}\right)^{-1}\left(u_{2,2}^{0}-(u_{1,2}^{1})^{2}(u_{1,1}^{1})^{-1}\right)
\end{equation}
Plugging in $\log\det(\bm{U})$ and $\mathrm{Tr}(\bm{U})$, and integrating
over $\bm{p}_{2}$ and $\bm{v}_{2}$, we have 

\begin{align}
 & \mathcal{M}\left(\left\{ \ell_{t}\right\} _{t=1,2}=0\right)\nonumber \\
= & \int du_{1,1}^{1}\int du_{1,2}^{1}\int du_{2,2}^{1}\int du_{2,2}^{0}\int d\bm{v}_{1}\nonumber \\
 & \exp\Bigg[\frac{N}{2}\log u_{1,1}^{1}+\frac{N}{2}n\log\left(u_{2,2}^{1}-u_{2,2}^{0}\right)+\frac{N}{2}n\left(u_{2,2}^{1}-u_{2,2}^{0}\right)^{-1}\left(u_{2,2}^{0}-(u_{1,2}^{1})^{2}(u_{1,1}^{1})^{-1}\right)-\sigma^{-2}\frac{N}{2}u_{1,1}^{1}-\sigma^{-2}n\frac{N}{2}u_{2,2}^{1}\nonumber \\
 & +i\bm{v}_{1}^{\top}\bm{Y}_{1}-\frac{1}{2}\bm{v}_{1}^{\top}\tilde{\bm{K}}_{1,1}\bm{v}_{1}-\frac{n}{2}\log\det\tilde{\bm{K}}_{2,2}-\frac{n}{2}u_{2,2}^{0}\mathrm{Tr}(\left(\tilde{\bm{K}}_{2,2}\right)^{-1}\bm{K}_{2,2}^{0})\nonumber \\
 & +\frac{N}{2}n\left(i\bm{Y}_{2}+\tilde{\bm{K}}_{2,1}\bm{v}_{1}\right)^{\top}\left(\tilde{\bm{K}}_{2,2}\right)^{-1}\left(i\bm{Y}_{2}+\tilde{\bm{K}}_{2,1}\bm{v}_{1}\right)\Bigg]\label{eq:mgfmulti-1-1}
\end{align}
Further integrating $\bm{v}_{1}$,
we have 
\begin{equation}
\mathcal{M}(\left\{ \ell_{t}\right\} _{t=1,2}=0)=\int du_{1,1}^{1}\int du_{1,2}^{1}\int du_{2,2}^{1}\int du_{2,2}^{0}\exp(-\frac{N}{2}\mathcal{H}_{\text{eff}}(u_{1,1}^{1},u_{1,2}^{1},u_{2,2}^{1},u_{2,2}^{0}))
\end{equation}
where
\begin{align}
\mathcal{H}_{\text{eff}}(u_{1,1}^{1},u_{1,2}^{1},u_{2,2}^{1},u_{2,2}^{0})&=\sigma^{-2}u_{1,1}^{1}-(1-\alpha)\log u_{1,1}^{1}+\frac{1}{N}\bm{Y}_{1}^{\top}\tilde{\bm{K}}_{1,1}^{-1}\bm{Y}_{1}+\log\det\bm{K}_{1,1}^{1} \nonumber \\
&-n\log\left(u_{2,2}^{1}-u_{2,2}^{0}\right)-n\left(u_{2,2}^{1}-u_{2,2}^{0}\right)^{-1}\left(u_{2,2}^{0}-(u_{1,2}^{1})^{2}(u_{1,1}^{1})^{-1}\right)+\sigma^{-2}nu_{2,2}^{1}\nonumber \\
&+n\frac{1}{N}\log\det\tilde{\bm{K}}_{2,2}+nu_{2,2}^{0}\frac{1}{N}\mathrm{Tr}(\tilde{\bm{K}}_{2,2}^{-1}\bm{K}_{2,2}^{0})-n\frac{1}{N}\mathrm{Tr}(\tilde{\bm{K}}_{2,1}\tilde{\bm{K}}_{1,1}^{-1}\tilde{\bm{K}}_{2,1}^{\top}\tilde{\bm{K}}_{2,2}^{-1})\nonumber \\
&+n\frac{1}{N}\left(\bm{Y}_{2}-\tilde{\bm{K}}_{2,1}\tilde{\bm{K}}_{1,1}^{-1}\bm{Y}_{1}\right)^{\top}\tilde{\bm{K}}_{2,2}^{-1}\left(\bm{Y}_{2}-\tilde{\bm{K}}_{2,1}\tilde{\bm{K}}_{1,1}^{-1}\bm{Y}_{1}\right)\label{eq:effectiveH}
\end{align}
We see that the leading terms in $u_{1,1}^{1}$ are $\mathcal{O}(1)$,
while the leading terms in $u_{2,2}^{1}$,$u_{2,2}^{0}$ and $u_{1,2}^{1}$
are all $\mathcal{O}(n)$, this indicates that the solution of $u_{1,1}^{1}$
will not be affected by the other renormalization factors, reflecting
the sequential nature of the learning.The self-consistent equations
for these renormalization factors are derived by taking derivative
of SI Eq. $\text{\ref{eq:effectiveH}}$ w.r.t. them and setting to 0, resulting in 
\begin{equation}
\sigma^{-2}\left(u_{1,1}^{1}\right)^{2}-(1-\alpha)u_{1,1}^{1}-\frac{1}{N}\bm{Y}_{1}^{\top}\left(\bm{K}_{1,1}^{1}\right)^{-1}\bm{Y}_{1}=0\label{eq:u11}
\end{equation}
\begin{equation}
u_{1,2}^{1}\left(u_{1,1}^{1}(u_{2,2}^{1}-u_{2,2}^{0})\right)^{-1}+\frac{1}{N}\mathrm{Tr}\left(\tilde{\bm{K}}_{2,1}\tilde{\bm{K}}_{1,1}^{-1}\bm{K}_{1,2}^{1}\tilde{\bm{K}}_{2,2}^{-1}\right) +\frac{1}{N}\bm{Y}_{1}^{\top}\tilde{\bm{K}}_{1,1}^{-1}\bm{K}_{1,2}^{1}\tilde{\bm{K}}_{2,2}^{-1}\left(\bm{Y}_{2}-\tilde{\bm{K}}_{2,1}\tilde{\bm{K}}_{1,1}^{-1}\bm{Y}_{1}\right)=0\label{eq:u12}
\end{equation}
\begin{align}
&\left(u_{2,2}^{1}-u_{2,2}^{0}\right)^{-1}-\sigma^{-2}+u_{2,2}^{0}\frac{1}{N}\mathrm{Tr}\left(\tilde{\bm{K}}_{2,2}^{-1}\Delta\bm{K}_{2,2}\tilde{\bm{K}}_{2,2}^{-1}\bm{K}_{2,2}^{0}\right) -\frac{1}{N}\mathrm{Tr}\left(\tilde{\bm{K}}_{2,2}^{-1}\bm{K}_{2,2}^{1}\right) \nonumber\\
&-\frac{1}{N}\mathrm{Tr}\left(\tilde{\bm{K}}_{2,1}\tilde{\bm{K}}_{1,1}^{-1}\tilde{\bm{K}}_{2,1}^{\top}\tilde{\bm{K}}_{2,2}^{-1}\Delta\bm{K}_{2,2}\tilde{\bm{K}}_{2,2}^{-1}\right)+\frac{1}{N}\left(\bm{Y}_{2}-\tilde{\bm{K}}_{2,1}\tilde{\bm{K}}_{1,1}^{-1}\bm{Y}_{1}\right)^{\top}\tilde{\bm{K}}_{2,2}^{-1}\Delta\bm{K}_{2,2}\tilde{\bm{K}}_{2,2}^{-1}\left(\bm{Y}_{2}-\tilde{\bm{K}}_{2,1}\tilde{\bm{K}}_{1,1}^{-1}\bm{Y}_{1}\right)=0\label{eq:u221}
\end{align}
\begin{align}
&\left(u_{2,2}^{0}-\left(u_{1,2}^{1}\right)^{2}\left(u_{1,1}^{1}\right)^{-1}\right)\left(u_{2,2}^{1}-u_{2,2}^{0}\right)^{-2}-u_{2,2}^{0}\frac{1}{N}\mathrm{Tr}\left(\tilde{\bm{K}}_{2,2}^{-1}\bm{K}_{2,2}^{0}\tilde{\bm{K}}_{2,2}^{-1}\bm{K}_{2,2}^{0}\right) \nonumber \\
&+\frac{1}{N}\mathrm{Tr}\left(\tilde{\bm{K}}_{2,1}\tilde{\bm{K}}_{1,1}^{-1}\tilde{\bm{K}}_{2,1}^{\top}\tilde{\bm{K}}_{2,2}^{-1}\bm{K}_{2,2}^{0}\tilde{\bm{K}}_{2,2}^{-1}\right) -\frac{1}{N}\left(\bm{Y}_{2}-\tilde{\bm{K}}_{2,1}\tilde{\bm{K}}_{1,1}^{-1}\bm{Y}_{1}\right)^{\top}\tilde{\bm{K}}_{2,2}^{-1}\bm{K}_{2,2}^{0}\tilde{\bm{K}}_{2,2}^{-1}\left(\bm{Y}_{2}-\tilde{\bm{K}}_{2,1}\tilde{\bm{K}}_{1,1}^{-1}\bm{Y}_{1}\right)=0\label{eq:u220}
\end{align}

\paragraph{Effective Hamiltonian for $T=3:$}

The effective Hamiltonian for $T=3$ can be derived similarly as for
$T=2$. For $T=3$, we have 9 renormalization factors, among which
$u_{1,1}^{1},u_{1,2}^{1},u_{2,2}^{1}$ and $u_{2,2}^{0}$ satisfy
SI Eqs.  \ref{eq:u11}-\ref{eq:u12}.
The other renormalization factors including $u_{1,3}^{1},u_{2,3}^{1},u_{2,3}^{0},u_{3,3}^{1}$
and $u_{3,3}^{0}$ can be solved by taking derivative of the following
effective Hamiltonian (SI Eq. \ref{eq:hamiltonian}) w.r.t. each of them and setting to 0.

\begin{align}
&\mathcal{H}_{\text{eff}}(u_{1,3}^{1},u_{2,3}^{1},u_{2,3}^{0},u_{3,3}^{1},u_{3,3}^{0}) = n\sigma^{-2}u_{3,3}^{1}+n\log\left(u_{3,3}^{1}-u_{3,3}^{0}\right) + n\left(u_{3,3}^{1}-u_{3,3}^{0}\right)^{-1}\left(u_{3,3}^{0}-\left(u_{1,3}^{1}\right)^{2}\left(u_{1,1}^{1}\right)^{-1}\right)\nonumber \\
& -2n\left(u_{3,3}^{1}-u_{3,3}^{0}\right)^{-1}\left(u_{2,3}^{1}-u_{2,3}^{0}\right)\left(u_{2,2}^{1}-u_{2,2}^{0}\right)^{-1}\left(u_{2,3}^{0}-u_{1,3}^{1}u_{1,2}^{1}\left(u_{1,1}^{1}\right)^{-1}\right)\nonumber \\
& -n\left(u_{3,3}^{1}-u_{3,3}^{0}\right)^{-1}\left(u_{2,3}^{1}-u_{2,3}^{0}\right)^{2}\left(u_{2,2}^{1}-u_{2,2}^{0}\right)^{-1}\left(1-\left(u_{2,2}^{1}-u_{2,2}^{0}\right)^{-1}\left(u_{2,2}^{0}-\left(u_{1,2}^{1}\right)^{2}u_{1,1}^{-1}\right)\right)\nonumber \\
& +n\log\det\tilde{\bm{K}}_{3,3}+nu_{3,3}^{0}\mathrm{Tr}\left(\tilde{\bm{K}}_{3,3}^{-1}\bm{K}_{3,3}^{0}\right)-n\mathrm{Tr}\left(\tilde{\bm{K}}_{3,3}^{-1}\tilde{\bm{K}}_{3,1}\tilde{\bm{K}}_{1,1}^{-1}\tilde{\bm{K}}_{1,3}\right)\nonumber \\
& +2n\mathrm{Tr}\left(\tilde{\bm{K}}_{3,3}^{-1}\tilde{\bm{K}}_{3,2}\tilde{\bm{K}}_{2,2}^{-1}\left(\tilde{\bm{K}}_{2,1}\tilde{\bm{K}}_{1,1}^{-1}\tilde{\bm{K}}_{1,3}-u_{2,3}^{0}\bm{K}_{2,3}^{0}\right)\right) - \\
&
n\mathrm{Tr}\left(\tilde{\bm{K}}_{3,3}^{-1}\tilde{\bm{K}}_{3,2}\left(\tilde{\bm{K}}_{2,2}^{-1}-\tilde{\bm{K}}_{2,2}^{-1}\left(u_{2,2}^{0}\bm{K}_{2,2}^{0}-\tilde{\bm{K}}_{2,1}\tilde{\bm{K}}_{1,1}^{-1}\tilde{\bm{K}}_{1,2}\right)\tilde{\bm{K}}_{2,2}^{-1}\right)\tilde{\bm{K}}_{2,3}\right)\nonumber \\
& +n\left(\bm{Y}_{3}-\tilde{\bm{K}}_{3,2}\tilde{\bm{K}}_{2,2}^{-1}\left(\bm{Y}_{2}-\tilde{\bm{K}}_{2,1}\tilde{\bm{K}}_{1,1}^{-1}\bm{Y}_{1}\right)-\tilde{\bm{K}}_{3,1}\tilde{\bm{K}}_{1,1}^{-1}\bm{Y}_{1}\right)^{\top}\tilde{\bm{K}}_{3,3}^{-1}\left(\bm{Y}_{3}-\tilde{\bm{K}}_{3,2}\tilde{\bm{K}}_{2,2}^{-1}\left(\bm{Y}_{2}-\tilde{\bm{K}}_{2,1}\tilde{\bm{K}}_{1,1}^{-1}\bm{Y}_{1}\right)-\tilde{\bm{K}}_{3,1}\tilde{\bm{K}}_{1,1}^{-1}\bm{Y}_{1}\right)\label{eq:hamiltonian}
\end{align}

\subsubsection{Interpretation of Renormalization Factors}

\label{paragraph:relationtoanorm}

In this section we show that the renormalization factors introduced
in SI \ref{subsec:multiheadtheory}.\ref{paragraph:krtheory}
are directly linked to the norm and inner product of the readout weights.
To show this, we can compute the readout norm by introducing moment
generating term coupled to them. Specifically we introduce $L_{t,t'}$
coupled to $\bm{a}_{t}\cdot\bm{a}_{t'}$ in the replicated partition
function, resulting in 
\begin{align}
\mathcal{M}\left(\bm{L}\right)= & \lim_{n\rightarrow0}\intop\prod_{\alpha=1}^{n}\prod_{t=2}^{T}d\bm{v}_{t}^{\alpha}\int d\bm{v}_{1}^{n}\int\prod_{\alpha=1}^{n}\prod_{t=2}^{T}d\Theta_{t}^{\alpha} \int d\Theta_{1}^{n}\exp(-i\sum_{t,\alpha}\sum_{\mu=1}^{P}v_{t}^{\alpha,\mu}\left(f_{t}^{t}\left(\Theta_{t}^{\alpha},\bm{x}_{t}^{\mu}\right)-y_{t}^{\mu}\right)\nonumber \\
& -\frac{1}{2}\sigma^{-2}\sum_{t,\alpha}\left\Vert \Theta_{t}^{\alpha}\right\Vert ^{2}-\frac{1}{2}\lambda\sum_{t=2}^{T}\sum_{\alpha=1}^{n}\left\Vert \mathcal{W}_{t}^{\alpha}-\mathcal{W}_{t-1}^{n}\right\Vert ^{2} -\frac{1}{2}\beta^{-1}\sum_{t,\alpha}\bm{v}_{t}^{\alpha\top}\bm{v}_{t}^{\alpha}-\sum_{t,t'}L_{t,t'}\bm{a}_{t}^{n}\cdot\bm{a}_{t'}^{n})
\end{align}
Thus by taking derivative of the MGF w.r.t. $\bm{L}$, we can derive
\begin{equation}
\langle\bm{a}_{t}\cdot\bm{a}_{t'}\rangle=-\frac{\partial\mathcal{M}(\bm{L})}{\partial L_{t,t'}}\lvert_{\bm{L}=0}
\end{equation}

We can perform the calculation in the same way as in SI \ref{subsec:multiheadtheory}.\ref{paragraph:krtheory}, define 
\begin{equation}
z_{t}^{\alpha}=N^{-1/2}\sigma\bm{v}_{t}^{\alpha\top}\Phi(\bm{w}_{t}^{\alpha},\bm{X}_{t})
\end{equation}
and we obtain 
\begin{align}
 & G(\left\{ \bm{v}_{t}^{\alpha}\right\} _{\alpha=1,\cdots,n;t=1,\cdots,T})=G(\bm{H})\nonumber \\
 & =N\log\Bigg[\intop\prod_{\alpha=1}^{n}\prod_{t=1}^{T}dz_{t}^{\alpha}\det(\bm{H})^{-1/2}\det(\mathbb{I}+\sigma^{2}\bm{L})^{-1/2}\nonumber \\
 & \exp\left(-\frac{1}{2}\sum_{t,\alpha}\sum_{t',\beta}z_{t}^{\alpha\top}[\bm{H}^{-1}]_{t,t',\alpha\beta}z_{t'}^{\beta}-\frac{1}{2}\sum_{t,\alpha}z_{t}^{\alpha}\left(\mathbb{I}+\sigma^{2}\bm{L}\right)^{-1}{}_{t,t',\alpha,\beta}z_{t'}^{\beta}\right)\Bigg]\\
 & =-\frac{N}{2}\log\det(\bm{H}+\mathbb{I}+\sigma^{2}\bm{L})
\end{align}
in place of SI Eq. \ref{eq:GH}, and 
\begin{align}
\mathcal{M}\left(\bm{L}\right)&=\lim_{n\rightarrow0}\intop\prod_{\alpha=1}^{n}\prod_{t=1}^{T}d\bm{v}_{t}^{\alpha}\int d\bm{U}\int d\bm{H}\exp(-\frac{1}{2}\beta^{-1}\sum_{t,\alpha}\bm{v}_{t}^{\alpha\top}\bm{v}_{t}^{\alpha}+i\sum_{t,\alpha}\bm{v}_{t}^{\alpha\top}\bm{Y}_{t}-\frac{N}{2}\log\det(\mathbb{I}+\bm{H}+\sigma^{2}\bm{L})\nonumber\\
&+\sigma^{-2}\frac{N}{2}\mathrm{Tr}(\bm{U}\bm{H})-\frac{1}{2}\sum\sum_{t',\beta}U_{t,t',\alpha,\beta}\bm{v}_{t}^{\alpha\top}\Phi(\bm{w}_{t}^{\alpha},\bm{X}_{t})\cdot\Phi(\bm{w}_{t'}^{\beta},\bm{X}_{t'})\bm{v}_{t'}^{\beta})\label{eq:mgfUHL}
\end{align}
in place of SI Eq. $\text{\ref{eq:mgfUH}}$.

Taking derivative of $\bm{H}$ and setting it to 0, we obtain,
\begin{equation}
\sigma^{-2}\bm{U}=\left(\mathbb{I}+\bm{H}+\sigma^{2}\bm{L}\right)^{-1}
\end{equation}
Plugging into SI Eq. \ref{eq:mgfUHL} yields
\begin{multline}
\mathcal{M}\left(\bm{L}\right)=\lim_{n\rightarrow0}\intop\prod_{\alpha=1}^{n}\prod_{t=1}^{T}d\bm{v}_{t}^{\alpha}\int d\bm{U}\int d\bm{H}\exp(-\frac{1}{2}\beta^{-1}\sum_{t,\alpha}\bm{v}_{t}^{\alpha\top}\bm{v}_{t}^{\alpha}+i\sum_{t,\alpha}\bm{v}_{t}^{\alpha\top}\bm{Y}_{t}+\frac{N}{2}\log\det(\bm{U})\\
-\sigma^{-2}\frac{N}{2}\mathrm{Tr}(\bm{U})-\frac{N}{2}\mathrm{Tr}(\bm{U}\bm{L})-\frac{1}{2}\sum\sum_{t',\beta}U_{t,t',\alpha,\beta}\bm{v}_{t}^{\alpha\top}\Phi(\bm{w}_{t}^{\alpha},\bm{X}_{t})\cdot\Phi(\bm{w}_{t'}^{\beta},\bm{X}_{t'})\bm{v}_{t'}^{\beta})
\end{multline}
in place of Supplementary Eq. $\text{\ref{eq:mgfreplica}}$.

Taking derivative w.r.t. $\bm{L}$, we have 
\begin{equation}
\langle\bm{a}_{t},\bm{a}_{t}\rangle=NU_{t,t',n,n}
\end{equation}
, and thus
\begin{equation}
N^{-1}\langle\bm{a}_{t}\cdot\bm{a}_{t'}\rangle=u_{t,t'}^{1}
\end{equation}
The renormalization factors $\left\{ u_{t,t'}^{1}\right\} _{t,t'=1,\cdots,T}$
are directly related the the covariances of the readout weights. In
the specific case of $T=2$, 
\begin{equation}
u_{2,2}^{1}=N^{-1}\langle\lVert\bm{a}_{2}\rVert^{2}\rangle
\end{equation}
\begin{equation}
u_{2,1}^{1}=N^{-1}\langle\bm{a}_{2}\cdot\bm{a}_{1}\rangle
\end{equation}
The renormalization factors $\left\{ u_{t,t'}^{0}\right\} _{t,t'=1,\cdots,T}$
are less interpretable, the difference between $\left\{ u_{t,t'}^{1}\right\} _{t,t'=1,\cdots,T}$
and $\left\{ u_{t,t'}^{0}\right\} _{t,t'=1,\cdots,T}$ can be related
to the susceptibility of the readout weights, but we leave out the
detailed calculation for simplicity.

\subsubsection{Predicting Phase Transition Boundaries with the Task Simialrity OP}
\label{paragraph:ptboundary}

Even when there are only two tasks, SI Eqs. \ref{eq:u11}-\ref{eq:u220} need to be solved numerically in general. In the $\lambda\rightarrow\infty$ limit, we can analyze the scaling of the solutions with $\lambda$.
Interestingly, we find 3 different consistent scalings of the solutions with $\lambda$. $u_{1,1}^{1}$ only depends on the first task, and
is always $\mathcal{O}(1)$ w.r.t. $\lambda$. In one regime, we have
$u_{2,2}^{1},u_{2,2}^{0},u_{2,2}^{1}-u_{2,2}^{0}\sim\mathcal{O}(1)$, SI Eqs. \ref{eq:u220}-\ref{eq:u12} can be simplified as 
\begin{equation}
u_{1,2}^{1}=0\label{eq:u12-1}
\end{equation}
\begin{equation}
u_{2,2}^{0}=\frac{1}{N(1-\alpha)}\bm{Y}_{2}^{\top}\left(\bm{K}_{2,2}^{1}\right)^{-1}\bm{Y}_{2}
\end{equation}
\begin{equation}
u_{2,2}^{1}-u_{2,2}^{0}=\sigma^{2}(1-\alpha)
\end{equation}
Note that since $u_{2,2}^{1}>u_{2,2}^{0}$ and $u_{2,2}^{0}>0$ (see SI \ref{subsec:multiheadtheory}.\ref{paragraph:relationtoanorm}), this solution can only hold when $\alpha<1$. In particular, in the
infinite-width limit, $\alpha=0$, and we have $u_{2,2}^{0}=0,u_{2,2}^{1}=\sigma^{2}$.
This result is equivalent to learning each task individually, with
Gaussian random hidden-layer weights.

When $\alpha>1$, there are two consistent scalings of the solutions
with $\lambda$. In one regime, we have $u_{2,2}^{1},u_{2,2}^{0}\sim\mathcal{O}(\lambda^{1/2}),u_{2,2}^{1}-u_{2,2}^{0}\sim\mathcal{O}(\lambda^{-1/2})$,
$u_{1,2}^{1}\sim\mathcal{O}(1)$. This corresponds to the overfitting
regime in. To the leading order, SI Eq. \ref{eq:u12} remains the same, SI Eqs. \ref{eq:u220}, \ref{eq:u221} are simplified as 
\begin{equation}
1-\frac{1}{N}\left(u_{2,2}^{1}-u_{2,2}^{0}\right)^{2}\mathrm{Tr}\left(\tilde{\bm{K}}_{2,2}^{-1}\bm{K}_{2,2}^{1}\tilde{\bm{K}}_{2,2}^{-1}\bm{K}_{2,2}^{1}\right)=0\label{eq:impededregime}
\end{equation}
\begin{align}
&\sigma^{-2}u_{2,2}^{0}-\left(u_{1,2}^{1}\right)^{2}\left(u_{1,1}^{1}\right)^{-1}\left(u_{2,2}^{1}-u_{2,2}^{0}\right)^{-1}+\frac{1}{N}\mathrm{Tr}\left(\tilde{\bm{K}}_{2,1}\tilde{\bm{K}}_{1,1}^{-1}\tilde{\bm{K}}_{1,2}\tilde{\bm{K}}_{2,2}^{-1}\right) \nonumber \\
&-\frac{1}{N}\left(\bm{Y}_{2}-\tilde{\bm{K}}_{2,1}\tilde{\bm{K}}_{1,1}^{-1}\bm{Y}_{1}\right)^{\top}\tilde{\bm{K}}_{2,2}^{-1}\left(\bm{Y}_{2}-\tilde{\bm{K}}_{2,1}\tilde{\bm{K}}_{1,1}^{-1}\bm{Y}_{1}\right)=0\label{eq:impededregime-1}
\end{align}

In the other regime, we have $u_{2,2}^{1},u_{2,2}^{0},u_{1,2}^{1}\sim\mathcal{O}(1),u_{2,2}^{1}-u_{2,2}^{0}\sim\mathcal{O}(\lambda^{-1})$, this corresponds to the generalization regime. To the leading order,
SI Eqs. \ref{eq:u12}, \ref{eq:u220} remain the same, SI Eq. \ref{eq:u221} is simplified as 
\begin{align}
&\left(u_{2,2}^{1}-u_{2,2}^{0}\right)^{-1}+u_{2,2}^{0}\frac{1}{N}\mathrm{Tr}\left(\tilde{\bm{K}}_{2,2}^{-1}\Delta\bm{K}_{2,2}\tilde{\bm{K}}_{2,2}^{-1}\bm{K}_{2,2}^{L,0}\right)-\frac{1}{N}\mathrm{Tr}\left(\tilde{\bm{K}}_{2,2}^{-1}\bm{K}_{2,2}^{L,1}\right) \nonumber\\
&-\frac{1}{N}\mathrm{Tr}\left(\tilde{\bm{K}}_{2,1}\tilde{\bm{K}}_{1,1}^{-1}\tilde{\bm{K}}_{2,1}^{\top}\tilde{\bm{K}}_{2,2}^{-1}\Delta\bm{K}_{2,2}\tilde{\bm{K}}_{2,2}^{-1}\right) +\frac{1}{N}\left(\bm{Y}_{2}-\tilde{\bm{K}}_{2,1}\tilde{\bm{K}}_{1,1}^{-1}\bm{Y}_{1}\right)^{\top}\tilde{\bm{K}}_{2,2}^{-1}\Delta\bm{K}_{2,2}\tilde{\bm{K}}_{2,2}^{-1}\left(\bm{Y}_{2}-\tilde{\bm{K}}_{2,1}\tilde{\bm{K}}_{1,1}^{-1}\bm{Y}_{1}\right)=0\label{eq:u221-1}
\end{align}

It is difficult to see where the transition between the two scalings
occur in general. Therefore, we make a further simplification assuming
that the kernels $\bm{K}_{2,2}^{1}$,$\bm{K}_{2,2}^{0},\tilde{\bm{K}}_{2,2}$
and $\Delta\bm{K}_{2,2}$ are different only in their magnitudes.
To the leading order $\bm{K}_{2,2}^{1}=\bm{K}_{2,2}^{0}$
\begin{equation}
\Delta\bm{K}_{2,2}\approx\sigma^{-2}\lambda^{-1}\bm{K}_{2,2}^{1}\sim\mathcal{O}(\lambda^{-1})
\end{equation}
\begin{equation}
\tilde{\bm{K}}_{2,2}\approx\left((u_{2,2}^{1}-u_{2,2}^{0})+\sigma^{-2}\lambda^{-1}u_{2,2}^{0}\right)\bm{K}_{2,2}^{1}
\end{equation}
This approximation is exact for linear networks, for nonlinear networks $\bm{K}_{2,2}^{0}$ and $\bm{K}_{2,2}^{1}$ are not only
different in their scales but also their magnitude, and the approximation
is only heuristic. In Figs. \ref{fig5:ptdistractor} in the main text, we see that the approximation is very accurate. With this approximation SI Eqs. \ref{eq:impededregime} , \ref{eq:impededregime-1} further simplify to 
\begin{equation}
\left(1+\sigma^{-2}\lambda^{-1}\frac{u_{2,2}^{0}}{u_{2,2}^{1}-u_{2,2}^{0}}\right)^{2}=\alpha\label{eq:impededregime-2}
\end{equation}
\begin{align}
&\sigma^{-2}\alpha^{1/2}u_{2,2}^{0}(u_{2,2}^{1}-u_{2,2}^{0})u_{1,1}^{1}-\alpha^{1/2}\left(u_{1,2}^{1}\right)^{2}+\left(u_{1,2}^{1}\right)^{2}\frac{1}{N}\mathrm{Tr}\left(\bm{K}_{2,1}^{1}\left(\bm{K}_{1,1}^{1}\right)^{-1}\bm{K}_{2,1}^{1}\left(\bm{K}_{2,2}^{1}\right)^{-1}\right)\nonumber\\
&-\frac{1}{N}u_{1,1}^{1}\bm{Y}_{2}^{\top}\left(\bm{K}_{2,2}^{1}\right)^{-1}\bm{Y}_{2}+2u_{1,2}^{1}\frac{1}{N}\bm{Y}_{2}^{\top}\left(\bm{K}_{2,2}^{1}\right)^{-1}\bm{K}_{2,1}^{1}\left(\bm{K}_{1,1}^{1}\right)^{-1}\bm{Y}_{1}\nonumber\\
&-\frac{1}{N}\left(u_{1,2}^{1}\right)^{2}\left(u_{1,1}^{1}\right)^{-1}\bm{Y}_{1}^{\top}\left(\bm{K}_{1,1}^{1}\right)^{-1}\bm{K}_{1,2}^{1}\left(\bm{K}_{2,2}^{1}\right)^{-1}\bm{K}_{2,1}^{1}\left(\bm{K}_{1,1}^{1}\right)^{-1}\bm{Y}_{1}=0\label{eq:impededregime-1-1}
\end{align}
For this solutions to be valid, it is evident that we require $\alpha>1$,
confirming the phase-transition boundary at $\alpha=1$. Furthermore,
since $u_{2,2}^{0},u_{2,2}^{1},u_{2,2}^{1}-u_{2,2}^{0}>0$, we have
\begin{align}
&\alpha^{-1/2}u_{1,2}^{1}-u_{1,2}^{1}\frac{1}{P}\mathrm{Tr}\left(\bm{K}_{2,1}^{1}\left(\bm{K}_{1,1}^{1}\right)^{-1}\bm{K}_{2,1}^{1}\left(\bm{K}_{2,2}^{1}\right)^{-1}\right) +\frac{1}{P}u_{1,1}^{1}\left(u_{1,2}^{1}\right)^{-1}\bm{Y}_{2}^{\top}\left(\bm{K}_{2,2}^{1}\right)^{-1}\bm{Y}_{2}-2\frac{1}{P}\bm{Y}_{2}^{\top}\left(\bm{K}_{2,2}^{1}\right)^{-1}\bm{K}_{2,1}^{1}\left(\bm{K}_{1,1}^{1}\right)^{-1}\bm{Y}_{1} \nonumber \\
&+\frac{1}{P}u_{1,2}^{1}\left(u_{1,1}^{1}\right)^{-1}\bm{Y}_{1}^{\top}\left(\bm{K}_{1,1}^{1}\right)^{-1}\bm{K}_{1,2}^{1}\left(\bm{K}_{2,2}^{1}\right)^{-1}\bm{K}_{2,1}^{1}\left(\bm{K}_{1,1}^{1}\right)^{-1}\bm{Y}_{1}>0\label{eq:impededregime-1-1-1}
\end{align}
This condition determines the phase-transition boundary between the overfitting regime ($u_{2,2}^{1},u_{2,2}^{0}\sim\mathcal{O}(\lambda^{1/2}),u_{2,2}^{1}-u_{2,2}^{0}\sim\mathcal{O}(\lambda^{-1/2})$)
and the generalization regime ($u_{2,2}^{1},u_{2,2}^{0}\sim\mathcal{O}(1),u_{2,2}^{1}-u_{2,2}^{0}\sim\mathcal{O}(\lambda^{-1})$). By
simplifying SI Eq. \ref{eq:u12} , solving for
$u_{1,2}^{1}$, and plugging in SI Eq. \ref{eq:impededregime-1-1-1},
we obtain

\begin{align}
&\frac{1}{P}\bm{Y}_1^\top \left(\bm{K}_{1,1}^1\right)^{-1} \bm{K}_{1,2}^1 \left(\bm{K}_{2,2}^1\right)^{-1} \bm{K}_{2,1}^1 \left(\bm{K}_{1,1}^1\right)^{-1} \bm{Y}_1 -\left(\frac{1}{P}\bm{Y}_1^\top \left(\bm{K}_{1,1}^1\right)^{-1}\bm{K}_{1,2}^1\left(\bm{K}_{2,2}^1\right)^{-1}\bm{Y}_2\right)^2\cdot \frac{1}{P}\bm{Y}_2^\top \left(\bm{K}_{2,2}^1\right)^{-1}\bm{Y}_2\nonumber\\
&>u_{1,1}^{1}\left(\frac{1}{P}\mathrm{Tr}\left(\bm{P}_1\bm{P}_2\right)-\alpha^{-1/2}\right) \label{eq:pt}
\end{align}
where $u_{1,1}^1$ is given by SI Eq. \ref{eq:u11}. Since we are in the $\lambda\rightarrow\infty$ limit, the kernel functions $K_{t,t'}^{1}$ are time invariant, and therefore the kernels in SI Eq. \ref{eq:pt} are given by $\bm{K}_{t,t'}^{1}=K_{GP}(\bm{X}_{t},\bm{X}_{t'})$ $t,t'\in\left\{ 1,2\right\} $. We have thus derived the phase-transition boundary and the relevant OPs as given in SI Eqs. \ref{eq:ptboundary}, \ref{eq:u11-1} in SI \ref{subsec:multiheadtheory}.\ref{subsubsec:summarymultihead}.

By properly normalizing the data, we can further simplify SI Eq. \ref{eq:pt}. Assuming that $\sigma^2=1$ and the data is normalized such that $\frac{1}{P} = 1$, and $\frac{1}{P} = 1$, SI Eq. \ref{eq:pt} can be simplified as 
\begin{equation}
-\frac{1}{P}\bm{Y}_1^\top \left(\bm{K}_{1,1}^1\right)^{-1} \bm{K}_{1,2}^1 \left(\bm{K}_{2,2}^1\right)^{-1} \bm{K}_{2,1}^1 \left(\bm{K}_{1,1}^1\right)^{-1} \bm{Y}_1 +\left(\frac{1}{P}\bm{Y}_1^\top \left(\bm{K}_{1,1}^1\right)^{-1}\bm{K}_{1,2}^1\left(\bm{K}_{2,2}^1\right)^{-1}\bm{Y}_2\right)^2 + \frac{1}{P}\mathrm{Tr}\left(\bm{P}_1\bm{P}_2\right) < \alpha^{-1/2} \label{eq:simpleptineq}
\end{equation}
If this inequality holds, the network in the overfitting regime, otherwise the network is in the generalization regime, thus setting the inequality to equality gives rise to the phase transition boundary, as given in SI Eq.\ref{eq:simpleptboundary}. We define the r.h.s. of Eq. \ref{eq:simpleptineq} as $\gamma_{\text{sim}}$, which ranges from -1 to 1.

\subsubsection{Hidden-Layer Kernels and Representational Changes}

\label{paragraph:Hidden-layer-kernels}

\paragraph{Hidden-Layer Kernels}
To estimate the hidden-layer kernels, we make a heuristic approximation. We assume that the probability distributions of $\Phi(\mathcal{W}_{2}^{\alpha},\bm{x})$ and
$\Phi(\mathcal{W}_{1},\bm{x})$ induced by the Gaussian prior proportional
to $\exp(-S_{0}(\mathcal{W}))$ (SI Eq.\ref{eq:prior}), can be approximated as Gaussian distributions with mean 0 and covariances
\begin{equation}
N^{-1}\langle\Phi(\mathcal{W}_{1},\bm{x})\cdot\Phi(\mathcal{W}_{1},\bm{x}')\rangle_{\mathcal{W}}=K_{1,1}^{1}(\bm{x},\bm{x}')
\end{equation}
\begin{equation}
N^{-1}\langle\Phi(\mathcal{W}_{2}^{\alpha},\bm{x})\cdot\Phi(\mathcal{W}_{2}^{\beta},\bm{x}')\rangle_{\mathcal{W}}=\begin{cases}
K_{2,2}^{1}(\bm{x},\bm{x}') & \alpha=\beta\\
K_{2,2}^{0}(\bm{x},\bm{x}') & \alpha\neq\beta
\end{cases}
\end{equation}
\begin{equation}
N^{-1}\langle\Phi(\mathcal{W}_{1},\bm{x})\cdot\Phi(\mathcal{W}_{2}^{\alpha},\bm{x}')\rangle=K_{1,2}^{1}(\bm{x},\bm{x}')
\end{equation}

This approximation is exact for linear networks, and allows us to evaluate second moments of the representations over the
posterior distribution in $\Theta$ by evaluating Gaussian integrals. In
particular, we define the similarity between representations (i.e. the hidden-layer kernels) as
\begin{equation}
K_{\text{sim}}(\bm{x},\bm{x}')\equiv N^{-1}\langle\Phi(\mathcal{W}_{2},\bm{x})\cdot\Phi(\mathcal{W}_{2},\bm{x}')\rangle\label{eq:defhiddenkernel}
\end{equation}
where the average is w.r.t. the posterior distribution of $\Theta$ for $T=2$.We have 
\begin{align}
K_{\text{sim}}(\bm{x},\bm{x}') &=K_{2,2}^{1}(\bm{x},\bm{x}')-\frac{1}{N}\Big(u_{1,1}^{1}\bm{k}_{2,1}^{1}(\bm{x})^{\top}\langle\bm{v}_{1}\bm{v}_{1}^{\top}\rangle\bm{k}_{1,2}^{1}(\bm{x}')+u_{2,2}^{0}\Delta\bm{k}_{2,2}(\bm{x})^{\top}\langle\bm{v}_{2}\bm{v}_{2}^{\top}\rangle\Delta\bm{k}_{2,2}(\bm{x}') \nonumber \\
&+u_{1,2}^{1}\bm{k}_{2,1}^{1}(\bm{x})^{\top}\langle\bm{v}_{1}\bm{v}_{2}^{\top}\rangle\Delta\bm{k}_{2,2}(\bm{x}')+u_{1,2}^{1}\Delta\bm{k}_{2,2}(\bm{x})^{\top}\langle\bm{v}_{2}\bm{v}_{1}^{\top}\rangle\bm{k}_{1,2}^{1}(\bm{x}')\nonumber\\
&+u_{2,2}^{0}\bm{k}_{2,2}^{0}(\bm{x})^{\top}\langle\delta\bm{v}_{2}\delta\bm{p}_{2}^{\top}\rangle\Delta\bm{k}_{2,2}(\bm{x}')+u_{2,2}^{0}\Delta\bm{k}_{2,2}(\bm{x})^{\top}\langle\delta\bm{v}_{2}\delta\bm{p}_{2}^{\top}\rangle\bm{k}_{2,2}^{0}(\bm{x}')\nonumber\\
&+\left(u_{2,2}^{1}-u_{2,2}^{0}\right)\left(\bm{k}_{2,2}^{1}(\bm{x})^{\top}\langle\bm{v}_{2}\bm{v}_{2}^{\top}\rangle\bm{k}_{2,2}^{1}(\bm{x}')-\bm{k}_{2,2}^{0}(\bm{x})^{\top}\langle\bm{v}_{2}\bm{v}_{2}^{\top}\rangle\bm{k}_{2,2}^{0}(\bm{x}')\right)\Big)\label{eq:hiddenkernel}
\end{align}
where the statistics of $\bm{v}_{2},\bm{p}_{2}$ and $\bm{v}_{1}$
are given in SI Eqs. \ref{eq:meanv}, \ref{eq:covpv}, \ref{eq:covv},
but with $\left\{ m_{t,t'}^{1}\right\} _{t,t'=1,\cdots,T},\left\{ m_{t,t'}^{0}\right\} _{t,t'=1,\cdots,T}$
replaced by $\left\{ u_{t,t'}^{1}\right\} _{t,t'=1,\cdots,T},\left\{ u_{t,t'}^{0}\right\} _{t,t'=1,\cdots,T}$.
$K_{\text{sim}}(\bm{x},\bm{x}')$ has two contributions, the first term in SI
Eq. \ref{eq:hiddenkernel} corresponds to averaging SI Eq. \ref{eq:defhiddenkernel} w.r.t. the Gaussian prior in
$\mathcal{W}_{2}$, and the rest of the terms are induced by learning, which we denote as $K(\bm{x},\bm{x}')$. $K({\bf x},{\bf x}')$ thus
captures the changes in the similarity matrix before learning (with Gaussian prior weights) and after learning. 

Although the learning induced terms are sub-leading and of $\mathcal{O}(1/N)$, we see in Fig. \ref{fig7:optimal-lambda} in the main text that the structure of these terms can affect the generalization performance. We note that, in the overfitting regime, because $u_{2,2}^{1},u_{2,2}^{0}\sim\mathcal{O}(\lambda^{1/2})$ and $u_{2,2}^{1}-u_{2,2}^{0}\sim\mathcal{O}(\lambda^{-1/2})$, several terms vanish in the large $\lambda$ limit, and we have 
\begin{align}
K_{\text{sim}}(\bm{x},\bm{x}')&=K_{2,2}^{1}(\bm{x},\bm{x}')-\frac{1}{N}\Big(u_{1,1}^{1}\bm{k}_{2,1}^{1}(\bm{x})^{\top}\langle\bm{v}_{1}\bm{v}_{1}^{\top}\rangle\bm{k}_{1,2}^{1}(\bm{x}') -\left(u_{2,2}^{0}\right)^{2}\Delta\bm{k}_{2,2}(\bm{x})^{\top}\tilde{\bm{K}}_{2,2}^{-1}\bm{K}_{2,2}^{0}\tilde{\bm{K}}_{2,2}^{-1}\Delta\bm{k}_{2,2}(\bm{x}') \nonumber\\
&+u_{2,2}^{0}\bm{k}_{2,2}^{0}(\bm{x})^{\top}\tilde{\bm{K}}_{2,2}^{-1}\Delta\bm{k}_{2,2}(\bm{x}')+u_{2,2}^{0}\Delta\bm{k}_{2,2}(\bm{x})^{\top}\tilde{\bm{K}}_{2,2}^{-1}\bm{k}_{2,2}^{0}(\bm{x}') \nonumber\\
&-\left(u_{2,2}^{1}-u_{2,2}^{0}\right)u_{2,2}^{0}\left(\bm{k}_{2,2}^{1}(\bm{x})^{\top}\tilde{\bm{K}}_{2,2}^{-1}\bm{K}_{2,2}^{0}\tilde{\bm{K}}_{2,2}^{-1}\bm{k}_{2,2}^{1}(\bm{x}')-\bm{k}_{2,2}^{0}(\bm{x})^{\top}\tilde{\bm{K}}_{2,2}^{-1}\bm{K}_{2,2}^{0}\tilde{\bm{K}}_{2,2}^{-1}\bm{k}_{2,2}^{0}(\bm{x}')\right)\Big)
\end{align}
The structure of $K_{\text{sim}}(\bm{x},\bm{x}')$ is not affected by the
labels of the second task $\bm{Y}_{2}$, confirming our observation
that the network fails to learn task 2 relevant representations in
this regime. In the main text Fig. \ref{fig7:optimal-lambda},
we evaluate the kernel on the training data, $K_{\text{sim}}(\bm{X}_{1})\equiv K_{\text{sim}}(\bm{X}_{1},\bm{X}_{1})\in\mathbb{R}^{P\times P}, K_{\text{sim}}(\bm{X}_{2})\equiv K_{\text{sim}}(\bm{X}_{2},\bm{X}_{2})\in\mathbb{R}^{P\times P}$. Similarly, $K(\bm{X}_{1})\equiv K(\bm{X}_{1},\bm{X}_{1})\in\mathbb{R}^{P\times P},K(\bm{X}_{2})\equiv K(\bm{X}_{2},\bm{X}_{2})\in\mathbb{R}^{P\times P}$ correspond to the learning-induced terms in $K_{\text{sim}}(\bm{X}_1)$ and $K_{\text{sim}}(\bm{X}_2)$.

\paragraph{Representational Changes}
\label{paragraph:representationchange}
For the changes in the representation of $\bm{x}$ defined as $\Delta\Phi(\bm{x})=\Phi(\mathcal{W}_{2},\bm{x})-\Phi(\mathcal{W}_{1},\bm{x})$,
we can derive the second moments in $\Delta\Phi(\bm{x})$ using the
same Gaussian approximation for calculating the hidden-layer kernels.
In the $\lambda\rightarrow\infty$ limit, we obtain
\begin{equation}
\Delta\Phi(\bm{x})\cdot\Delta\Phi(\bm{x}') =-u_{2,2}^{0}\Delta\bm{k}_{2,2}(\bm{x})^{\top}\langle\bm{v}_{2}\bm{v}_{2}^{\top}\rangle\Delta\bm{k}_{2,2}(\bm{x}')
\end{equation}
By analyzing the scalings of $u_{2,2}^{1},u_{2,2}^{0}$,$u_{2,2}^{1}-u_{2,2}^{0}$
and $u_{1,2}^{1}$ in the different regimes in the $\lambda\rightarrow\infty$
limit, we can analyze the behavior of $\Delta\Phi(\bm{x})\cdot\Delta\Phi(\bm{x}')$.
In the regime $\alpha<1$, we have $\Delta\Phi(\bm{x})\cdot\Delta\Phi(\bm{x}')=0$, confirming
that the hidden-layer weights do not change from task 1 to task 2.
In the overfitting regime, by evaluating $\langle\bm{v}_{2}\bm{v}_{2}^{\top}\rangle$
with the corresponding scaling of the renormalization factors, we
have
\begin{equation}
\Delta\Phi(\bm{x})\cdot\Delta\Phi(\bm{x}') =\left(u_{2,2}^{0}\right)^{2}\Delta\bm{k}_{2,2}(\bm{x})^{\top}\tilde{\bm{K}}_{2,2}^{-1}\bm{K}_{2,2}^{0}\tilde{\bm{K}}_{2,2}^{-1}\Delta\bm{k}_{2,2}(\bm{x}')\label{eq:changefeatureof}
\end{equation}
In the generalization regime, we have 
\begin{align}
\Delta\Phi(\bm{x})\cdot\Delta\Phi(\bm{x}')\nonumber &= \left(u_{2,2}^{0}\right)^{2}\Delta\bm{k}_{2,2}(\bm{x})^{\top}\tilde{\bm{K}}_{2,2}^{-1}\bm{K}_{2,2}^{0}\tilde{\bm{K}}_{2,2}^{-1}\Delta\bm{k}_{2,2}(\bm{x}')\nonumber \\
&+u_{2,2}^{0}\Delta\bm{k}_{2,2}(\bm{x})^{\top}\tilde{\bm{K}}_{2,2}^{-1}\left(\bm{Y}_{2}-\tilde{\bm{K}}_{2,1}\tilde{\bm{K}}_{1,1}^{-1}\bm{Y}_{1}\right)\left(\bm{Y}_{2}-\tilde{\bm{K}}_{2,1}\tilde{\bm{K}}_{1,1}^{-1}\bm{Y}_{1}\right)^{\top}\tilde{\bm{K}}_{2,2}^{-1}\Delta\bm{k}_{2,2}(\bm{x}')\label{eq:changefeature}
\end{align}

As long as $\alpha>1$, $\lVert\Delta\Phi(\bm{x})\rVert^{2} > 0$. Therefore, $F_{2,1} = 0$ in the overfitting regime suggests that $\Delta\Phi(\bm{x})$ has to be in the null space of $\bm{a}_1$. Furthermore, Eqs. \ref{eq:changefeatureof}, \ref{eq:changefeature} shows that the overfitting regime the structure of $\Delta\Phi(\bm{x})\cdot\Delta\Phi(\bm{x}')$
is not aligned with either $\bm{Y}_{1}$ nor $\bm{Y}_{2}$, and in
the generalization regime the structure is affected by both $\bm{Y}_{1}$
and $\bm{Y}_{2}$. 

\part{Details and Parameters of Numerical Experiments}
\label{section:numericsdetails}
\section{The Target-Distractor Task Sequence}
\label{subsec:targetdistractor}
To construct the target-distractor task sequence, we used the CIFAR-100 source dataset  \citep{krizhevsky2009learning}, converted to grayscale. As a preprocessing step, all images are centered (zero-meaned), whitened, and normalized (such that the
squared norm of every image is the input dimension of each source
dataset). Each image is randomly labeled as $\pm 1$ or $0$. In all tasks, $x\%$ of the images are drawn from the ones with $\pm 1$ labels, and the rest are drawn from the ones labeled $0$. For each task sequence, among the images with $\pm 1$ labels, a fraction of $\rho_{\text{target}}$ images are shared across all tasks within the sequence, and the rest are unique for each task. Among the images with $0$ labels, a fraction of $\rho_{\text{shared}}-x\%\cdot\rho_{\text{target}}$ images are shared across all tasks within the sequence, and the rest are unique for each task. Therefore a total fraction of $\rho_{\text{shared}}$ images are the same across all tasks in the sequence, and the rest of the images are unique for each task. Among the images with $\pm 1$ labels and are shared across all tasks within the sequence, the signs of the labels are flipped with probability $\rho_{\text{flipped}}$. $\rho_{\text{flipped}}$ can be varied within the range of $[0, 0.5]$ independently of the other two parameters. For $\rho_{\text{shared}}$ in the range of $[x\%, 1-x\%]$, $\rho_{\text{target}}$ can assume any values in $[0, 1]$, such that $\rho_{\text{target}}$ and $\rho_{\text{shared}}$ can be varied independently. However, for $\rho_{\text{shared}} \in [0, x\%]$, $\rho_{\text{target}}$ can only be in the range of $[0, \rho_{\text{shared}}]$. Similarly if $\rho_{\text{shared}} \in [1-x\%, 1]$, $\rho_{\text{target}}$ can only be in the range of $[\rho_{\text{target}},1]$. Therefore, choosing a small $x\%$ allows $\rho_{\text{target}}$ and $\rho_{\text{shared}}$ to be manipulated more independently. In all our results we chose $x\%=10\%$. We also fixed $P=2000$, in Fig. \ref{fig3:distractortask} $L=10$ and in Fig. \ref{fig5:ptdistractor} $L=1$. 

In Fig. \ref{fig3:distractortask}\textbf{c}, we computed the normalized PVE (proportion of variance explained, \cite{o1982measures}) of $F_{2,1}$ and $\tau_F$. For $F_{2,1}$, we first employed a multiple linear regression model with $F_{2,1}$ as the dependent variable and $\gamma_{\text{RF}}$, $\gamma_{\text{rule}}$, $\gamma_{\text{feature}}$ as the independent variables. We calculated the total $R^2$ value of this regression model to represent the proportion of variance in $F_{2,1}$ collectively explained by all three variables. To evaluate the contribution of each variable individually, we computed the $R^2$ values of the reduced models where each variable was removed one at a time. The PVE is then computed as the difference between the $R^2$ of the full model and the $R^2$ of the reduced model, divided by the $R^2$ of the full model. We then normalized the PVE such that the the PVE's of the three variables sum up to 1. The same procedure is done to calculate the PVE of $\tau_F$. 

In Fig. \ref{fig4:OPsandbenchmark}\textbf{b},\textbf{d}, we showed the generalization performance $G_{2,2}$ using the target-distractor task sequence, to illustrate its divergent behavior. The test data is not defined for the target-distractor task sequence, as the images are randomly labeled, we thus evaluated the 'generalization' using perturbed images of the training data by adding Gaussian noise to the inputs. The standard deviation of the noise is fixed at $0.2$. In Fig. \ref{fig4:OPsandbenchmark}\textbf{f}, we showed the transition boundary $\alpha_c$, estimated using the shape of $F_{2,1}$ as a function of $\alpha$. Specifically, $\alpha_c$ was estimated as the point where $F_{2,1}$ exhibited the most rapid change with respect to $\alpha$. 

\section{Benchmark Task Sequences}
\label{sec:benchmark}
All source datasets used (MNIST \citep{lecun1998gradient}, EMNIST
\citep{cohen2017emnist}, Fashion-MNIST \citep{xiao2017fashion},
and CIFAR-100 \citep{krizhevsky2009learning}) are image classification
datasets. The images are either grayscale (MNIST, EMNIST, Fashion-MNIST)
or converted to grayscale (CIFAR-100). Similarly as for the target-distractor task sequence, as preprocessing, all images
are centered, whitened, and normalized. All of our analysis used subsets of $P=2000$ images in the training set of each task to save computational cost, as commonly done in theoretical studies of deep NNs (e.g., \citep{jacot2018neural,li2021statistical})
-- the subset of images used are redrawn for each random seed used. As shown in Fig. \ref{fig:dependencep}, further increasing the number of images do not significantly improve  generalization performance on previously learned tasks. The specific protocols used for generating task sequences from source datasets are detailed below. In all cases we use the MNIST dataset, which consists of images of digits ``0'' through ``9'', as an example to explain the protocols. 

\subsection{Permutation}
\label{subsec:permutation}
Our permutation protocol largely follows standard practices in the literature \citep{goodfellow2013empirical}. Each source dataset is first turned into a binary classification dataset by randomly dividing the original image classes (e.g., ``0'' through ``9'') into two groups. Images from one group are assigned target label $+1$ and those from the other are assigned $-1$. All training and test images corresponding to the same task undergo the same randomly generated pixel permutation, where the fraction of pixels permuted (relative to the original unpermuted images) is termed the ``permutation ratio''.
The permutation is independently generated for each task. Inputs in each of $D_{1},...,D_{T}$ are permuted versions of the same subset of images. At zero permutation $D_{1}=D_{2}=...=D_{T}$. \\
Note that the protocol above differs from standard practices in that we also permuted images in $D_{1}$ -- this is to ensure such that any pair of tasks in a long sequence have the same statistical relations (Fig. \ref{fig4:OPsandbenchmark}). In all analysis of multi-head CL, as we focused on CL of two tasks, we followed standard practices and did not permuted $D_{1}$ (Figs. \ref{fig6:phase-transition-benchmark}, \ref{fig7:optimal-lambda}).

\subsection{Split}

Our split protocol also largely follows standard practices in the literature \citep{goodfellow2013empirical,zenke2017continual}. Each task contains only images from a disjoint set of classes (e.g.,
task 1 is ``0'' vs. ``1'', task 2 is ``2'' vs. ``3''). Thus, the maximum length of the sequence is limited by the number of classes in the source dataset -- since MNIST and Fashion-MNIST each contains 10 classes, they can produce sequences with at most five tasks. On
the other hand, CIFAR-100 contains 100 classes and EMNIST contains 62 classes, allowing for much longer sequences. Thus, our analysis
of long-term forgetting only applies the split protocol to CIFAR-100 and EMNIST (Fig. \ref{fig4:OPsandbenchmark}). In EMNIST, each task includes a disjoint pair of classes. In CIFAR-100, we are limited by the amount of examples available for each class (500 images per class in the training set), in order to have $P=2000$, each task includes a disjoint set of 4 classes, and the task is to classify 2 of these classes vs. the other 2. Each random seed corresponds to a different subsample of images from each class being used, as well as a random assignment of classes into disjoint sets.

In the special case of having only two tasks (Figs. \ref{fig6:phase-transition-benchmark}, \ref{fig7:optimal-lambda}), we designed a ``partial split'' protocol to parametrically vary relations between $D_{1}$ and $D_{2}$. As an example, suppose the first pair of classes is ``0'' and ``1''. The second pair is ``2'' and ``3''. Under $x\%$ split, the training/test sets of task 1 would have $\left(x/2+50\right)\%$ images from the first pair and $\left(-x/2+50\right)\%$ from the second pair, whereas
task 2 would have $\left(-x/2+50\right)\%$ from the first and $\left(x/2+50\right)\%$
from the second. Images in both tasks are labeled according to the rule ``0'', ``1'' vs. ``2'', ``3''. Under $0\%$ split the two datasets would be identical.



\section{Exponential Fitting of Long-Term Forgetting}
\label{sec:exponential}
\begin{figure}[H]
\centering
\includegraphics[width=0.8\linewidth]
{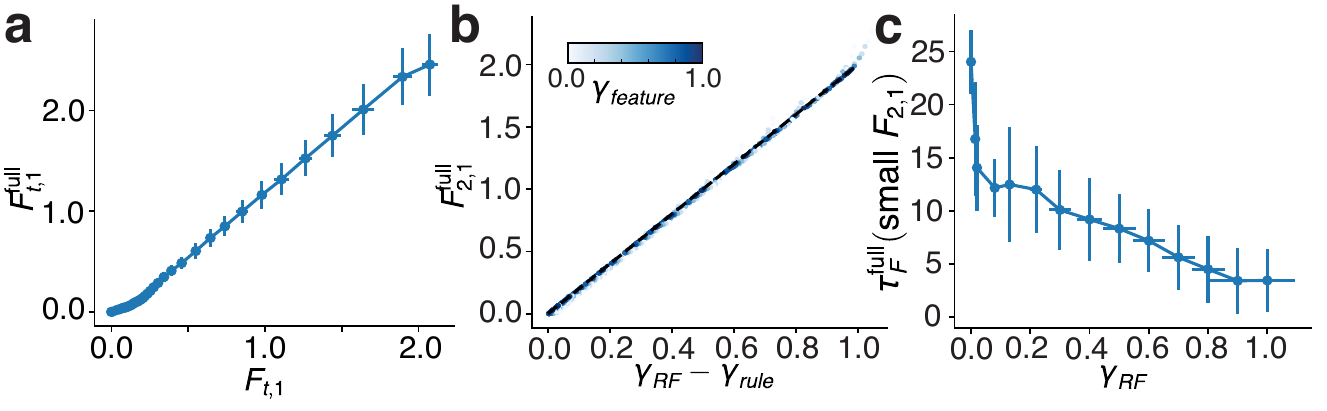}
\caption{\textbf{Comparison between results under the full Gibbs distribution and the random feature model.} \\
\textbf{a} Forgetting evaluated using the full theory ($F_{t,1}^{\text{full}}$, SI Eq. \ref{SIeq:meanpredictor}) vs. the random feature approximation ($F_{t,1}$) for $t$ from $1$ to $T$. The full theory behaves similarly to the random feature approximation across a wide range of $t$ and task relations. \\
\textbf{b} Short-term forgetting evaluated using the full theory ($F_{2,1}^{\text{full}}$) behaves similarly as $F_{2,1}$ evaluated using the random feature approximation. $F_{2,1}^{\text{full}}$ can also be accurately captured by the interference $\gamma_{\text{RF}}-\gamma_{\text{rule}}$, and is independent of $\gamma_{\text{feature}}$. \\
\textbf{c} Long-term forgetting evaluated using the full theory also behaves similarly as the random feature approximation. We focus on the time constant $\tau_F^{\text{full}}$ for task sequences with small $F_{2,1}^{\text{full}}$ as in Fig. \ref{fig3:distractortask} in the main text. $\tau_F^{\text{full}}$ also decreases with $\gamma_{\text{RF}}$. The decrease is fast when $\gamma_{\text{RF}}$ is close to 0 and slower for larger $\gamma_{\text{RF}}$. \\
All results were evaluated on the target-distractor task sequences with the same parameters as in Fig. \ref{fig3:distractortask}, and detailed in SI \ref{subsec:targetdistractor}. Results were averaged over 40 random seeds for task sequence generation.}
\label{fig:comparisonfull}
\end{figure}

\begin{figure}[H]
\centering
\includegraphics[width=0.7\linewidth]
{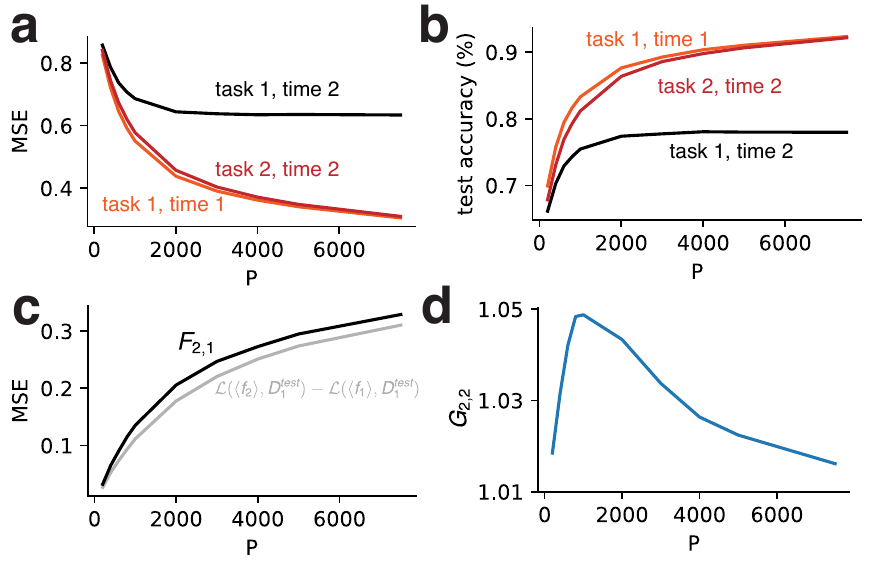}
\caption{\textbf{Generalization in single-head CL and the choice of P.}\\ 
\textbf{a} MSE on the test data of task 1 and task 2, measured after learning task 2. For learning single tasks (task 1, time 1), or learning the current task (task 2, time 2), generalization error improves with $P$. For the previously learned task (task 1, time 2), generalization error plateaus at around $P=2000$, which is the value of $P$ we selected for presenting all results in the main text.\\ 
\textbf{b} Same as $\textbf{a}$, but showing the test accuracy. The accuracy is measured by the percentage of the mean input-output mappings $\langle f_T\rangle$ that have the same sign as the labels. \\
\textbf{c} Short-term forgetting measured on the training data ($F_{2,1}$) increases with $P$. The behavior of $F_{2,1}$ is similar to forgetting measured using the generalization error, i.e., the difference between the generalization error on task 1 after learning task 2 ($\mathcal{L}(\langle f_2\rangle, D_1^{\text{test}})$) and the generalization error of learning task 1 alone ($\mathcal{L}(\langle f_1\rangle, D_1^{\text{test}})$).\\
\textbf{d} The anterograde effect is measured by $G_{2,2} \equiv \mathcal{L}(\langle f_2\rangle , D_2^{\text{test}})/G_2^0$, where $G_2^0$ is the generalization error when learning task 2 alone. $G_{2,2}$ remains close to 1 across all values of $P$, suggesting that the anterograde effect is weak. \\
Results were computed on permuted MNIST with $100\%$ permutation ratio using the random feature approximation, in a network with $L=9$, and averaged over 50 random seeds for task sequence generation.}
\label{fig:dependencep}
\end{figure}

\begin{figure}[H]
\centering
\includegraphics[width=0.7\linewidth]
{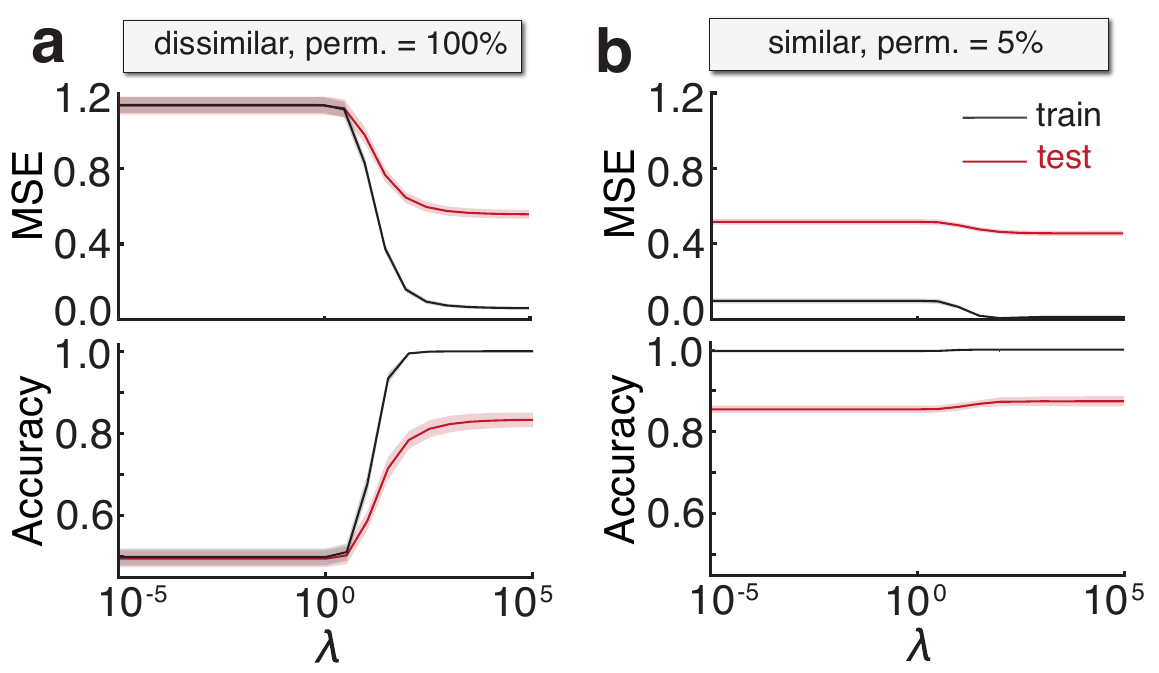}
\caption{\textbf{Dependence of CL performance on $\lambda$.} \\
\textbf{a} MSE and accuracy on both train (black) and test (red) data on the first task, after learning two tasks, for permuted MNIST with $100\%$ permutation ratio. $\lambda$ has a strong effect in mitigating forgetting. MSE on both train and test data decreases significantly with $\lambda$, and accuracy increases.\\ 
\textbf{b} Same as \textbf{a}, but for permuted MNIST with $5\%$ permutation ratio. The effect of $\lambda$ in mitigating forgetting is small, MSE remains small, and accuracy remains high across all values of $\lambda$. \\
The result suggests that $\lambda$ plays a crucial role in mitigating forgetting in dissimilar tasks, but is less effective in similar tasks, where forgetting is already small without any regularizers, consistent with Fig. \ref{fig3:distractortask}\textbf{a} inset in the main text.
Shaded errorbars are standard deviations across 10 random seeds for task sequence generation.  Both MSE and accuracy are evaluated using the mean input-output mappings. The accuracy is measured by the percentage of the mean input-output mappings ($\langle f_T\rangle$) that have the same sign as the labels. Results were computed with $L=9$ and $P=2000$, using the mean input-output mappings over the full Gibbs distribution. }
\label{fig:dependencelambda}
\end{figure}

\begin{figure}[H]
\centering
\includegraphics[width=\linewidth]
{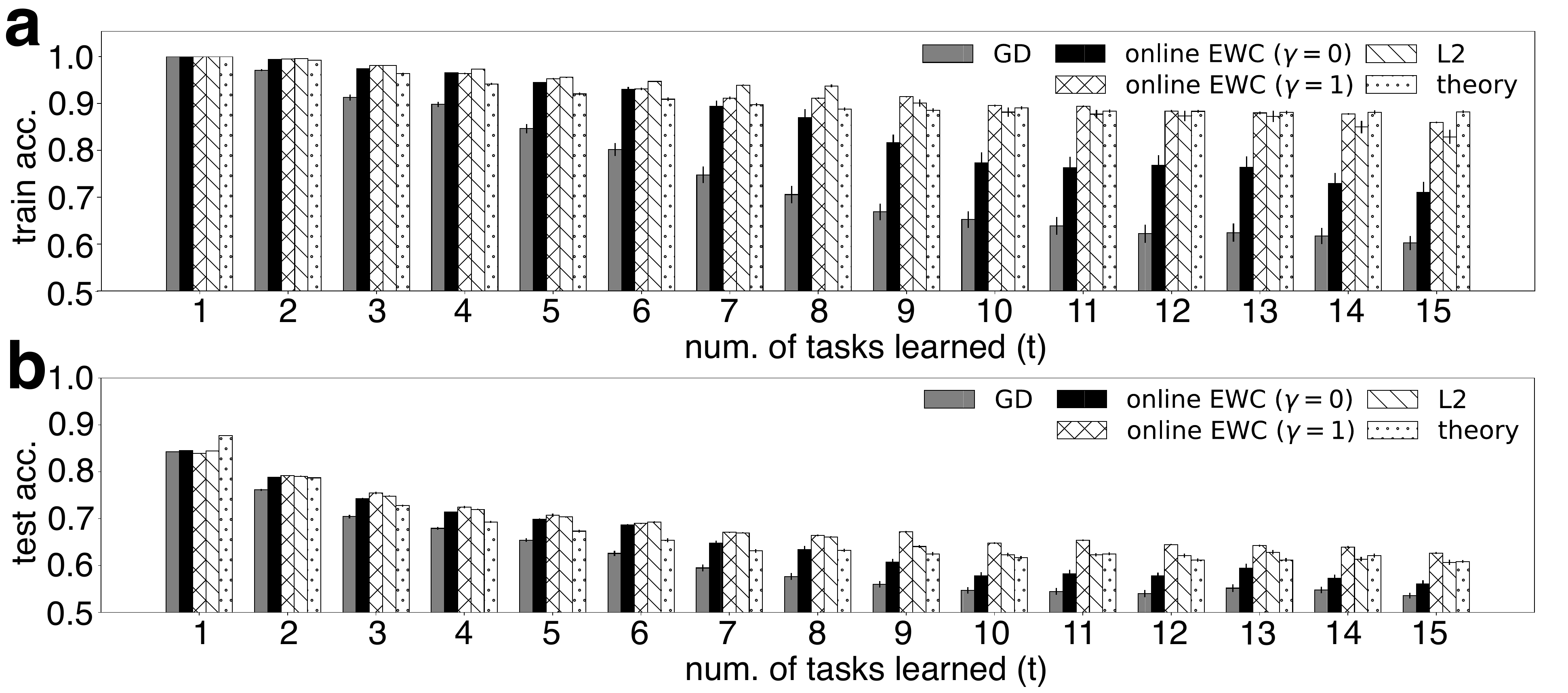}
\caption{\textbf{Forgetting in networks trained with gradient descent with different regularizers in single-head CL.} \\
\textbf{a} Train accuracy on the first task after learning $t$ tasks, for $t$ ranging from 1 to 15. We evaluated the accuracy for networks trained with vanilla gradient descent without explicit regularizers (GD), with online EWC with different decay parameters (online EWC ($\gamma=0$) and online EWC ($\gamma=1$)) or with an $L-2$ regularizer (L2), and compared their performance with accuracy obtained using our theoretical result of $\langle f_T \rangle$. Our theory achieves comparable performance L2 regularizer and online EWC with $\gamma=1$, and outperforms vanilla GD and online EWC with $\gamma=0$. \\ 
\textbf{b} Same as \textbf{a}, but showing the test accuracy. \\
Details of the training process are in SI \ref{sec:gdsimulations}.\ref{subsec:singleheadgd}. For the numerics (GD, online EWC ($\gamma = 0$ and $\gamma=1$), L2), errorbars are standard errors across 100 random seeds of both task sequence generation and initialization. The theory is evaluated using the mean input-output mapping, and the accuracy is measured by the percentage of the mean input-output mappings ($\langle f_T\rangle$) that have the same sign as the labels. The errorbars are standard errors across 50 random seeds of task sequence generation. Results were obtained with $L=9$ and $P=2000$ on the permuted MNIST task sequence with permutation ratio $100\%$, theoretical results were calculated with the random feature approximation, GD simulations with different regularizers were done with $\eta=0.001$, $\kappa=0.1$, $\sigma_0=1$. 
}
\label{fig:gdsingle}
\end{figure}

\begin{figure}[H]
\centering
\includegraphics[width=0.7\linewidth]
{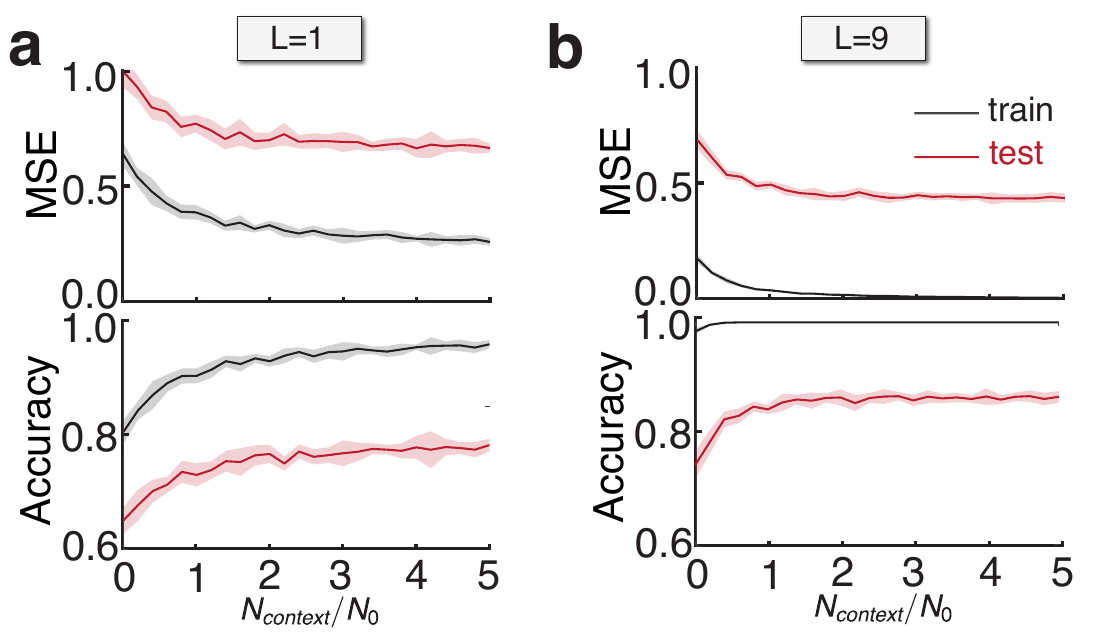}
\caption{\textbf{Incorporating task-identity information in single-head CL.} 
\textbf{a} MSE and accuracy measured on the train and test data on the first task after learning two tasks, as a function of the relative context length $N_{\text{context}}/N_0$, in a network with $L=1$. Forgetting performance measured with both train and test data improves significantly with the context length. $F_{2,1}$ goes to a finite value larger than 0 as the context length increases. The task identity information is incorporated by appending an $N_{\text{context}}$-dimensional task-dependent context vector to each $N_0$-dimensional input. The context vector is drawn from an i.i.d. Gaussian distribution of dimension $N_{\text{context}}$, and is re-sampled for each new task. \\
\textbf{b} Same as \textbf{a}, but for a network with $L=9$.\\
The results show that effectiveness of incorporating task-identity information in mitigating forgetting is stronger in shallower networks, but still less effective than using multi-head CL (where forgetting can be 0). Forgetting in deeper networks is already small without incorporating task-identity information. Results were evaluated on the permuted MNIST task sequence with permutation ratio $100\%$ and $P=2000$, using the random feature approximation of the mean input output mappings. Accuracy is measured by the percentage of the mean input-output mappings that have the same sign as the labels. Errorbars are across 10 random seeds of both task sequence generation and sampling of the context vectors.
}
\label{fig:singlecontext}
\end{figure}

All exponential fitting of forgetting $F_{t,1} $was carried out on the forgetting averaged over randomness of task sequence generation. In the target-distractor task sequence, the randomness
refers to the randomly selected subsets of images and the random labels in each task (see SI \ref{subsec:targetdistractor}). For benchmark task sequences, it refers factors such as different subsamples of the full source dataset (SI \ref{sec:benchmark}). For permutation task sequences with very low permutation ratios, we sometimes observed a non-monotonic relation between averaged $F_{t,1}$ and $t$. In these cases we truncated the averaged $F_{t,1}$ at the maximum before fitting. In Fig. \ref{fig3:distractortask} and Fig. \ref{fig:comparisonfull}, we presented results averaged over 40 random seeds for the target-distractor task sequence, and in Figs. \ref{fig4:OPsandbenchmark}, \ref{fig:dependencep}, \ref{fig:gdsingle} we presented results averaged over 50 random seeds for the benchmark task sequences. 

\section{Gradient Descent Simulations}
\label{sec:gdsimulations}
\subsection{Single-Head CL}
\label{subsec:singleheadgd}
We performed numerical simulations using gradient descent dynamics with different types of regularizers and compared with our theoretical results. 
\subsubsection{Gradient Descent}
\label{subsubsec:gd}
For networks trained with vanilla gradient descent dynamics (GD), the dynamics is simply given by 
\begin{equation}
\Theta_t^\tau = \Theta_t^{\tau-1} - \eta\nabla_{\Theta_t^{\tau-1}}\mathcal{L}(f_t(\Theta_t^{\tau-1}), D_t)
\end{equation}
here we use $t$ to denote the index of the task, and $\tau$ to denote the time steps during training. For the first task the weights are initialized as $\Theta_1^0 \sim \mathcal{N}(0, \sigma_0^2)$, for the following tasks the weights $\Theta_t^0$ are initialized as $\Theta_{t-1}$, i.e., the network weights after learning task $t-1$. The learning process is run until the MSE on the current training data is less than $10^{-3}$. 

\subsubsection{L-2 regularizer}
\label{subsubsec:l2}
For networks trained with an L-2 regularizer, the dynamics is given by 
\begin{equation}
\Theta_t^\tau = \Theta_t^{\tau-1} - \eta\nabla_{\Theta_t^{\tau-1}} \mathcal{L}(f_t(\Theta_t^{\tau-1}), D_t) - \eta\kappa(\Theta_t^{\tau-1} - \Theta_{t-1})
\end{equation}
where $\Theta_{t-1}$ denotes the network weights after finishing learning task $t-1$. Similarly, for the first task the weights are initialized as $\Theta_1^0\sim\mathcal{N}(0,\sigma_0^2)$ and for the following tasks the weights $\Theta_t^0$ are initialized as $\Theta_{t-1}$. For the first task, the learning process was run without the regularizer, and ends when the MSE on the current training data is less than $10^{-3}$. For the following tasks, the learning process is run sufficiently long, in order to approximate the $\lambda\rightarrow \infty$ solution where the closest solution to $\mathcal{W}_{t-1}$ is obtained. 

\subsubsection{Online EWC}
\label{subsubsec:onlineEWC}
For networks trained with online EWC, the dynamics is given by 
\begin{equation}
\Theta_t^\tau = \Theta_t^{\tau-1} - \eta\nabla_{\Theta_t^{\tau-1}} \mathcal{L}(f_t(\Theta_t^{\tau-1}),D_t) - \eta\kappa \bar{F}_t\otimes(\Theta_t^{\tau-1} - \Theta_{t-1})
\end{equation}

\begin{equation}
    \bar{F}_t = \gamma \bar{F}_{t-1} + F_{t}; F_1 = 0
\end{equation}
where $F_t$ denotes the diagonal elements of the normalized fisher information matrix at time $t$,

\begin{equation}
    F_t^{i} = \frac{\left(\frac{\partial \mathcal{L}(f_t(\Theta_t),D_t)}{\partial \Theta_t^i}\right)^2}{\frac{1}{N_{\text{params}}}\sqrt{\sum_{i=1}^{N_{\text{params}}}\left(\frac{\partial \mathcal{L}(f_t(\Theta_t),D_t)}{\partial \Theta_t^i}\right)^4}}
\end{equation}
where $N_{\text{params}}$ denotes the total number of parameters in the network. $F_t$ (and therefore $\bar{F}_t$) has the same number of elements as $\Theta_t$, and $\otimes$ denotes element-wise product between $\bar{F}_t$ and $\Theta_t^{\tau-1}-\Theta_{t-1}$. The decay parameter $\gamma$ controls how much information about the old tasks is preserved, at $\gamma=1$, the fisher information of all previous tasks contributes equally, and at $\gamma=0$, only the fisher information of the last task is retained. The initialization $\Theta_t^0$ and the stopping criteria for each task are the same as in SI \ref{sec:gdsimulations}.\ref{subsubsec:l2}. 

\subsection{Multi-Head CL}
\begin{figure}[H]
\includegraphics[width=1\linewidth]{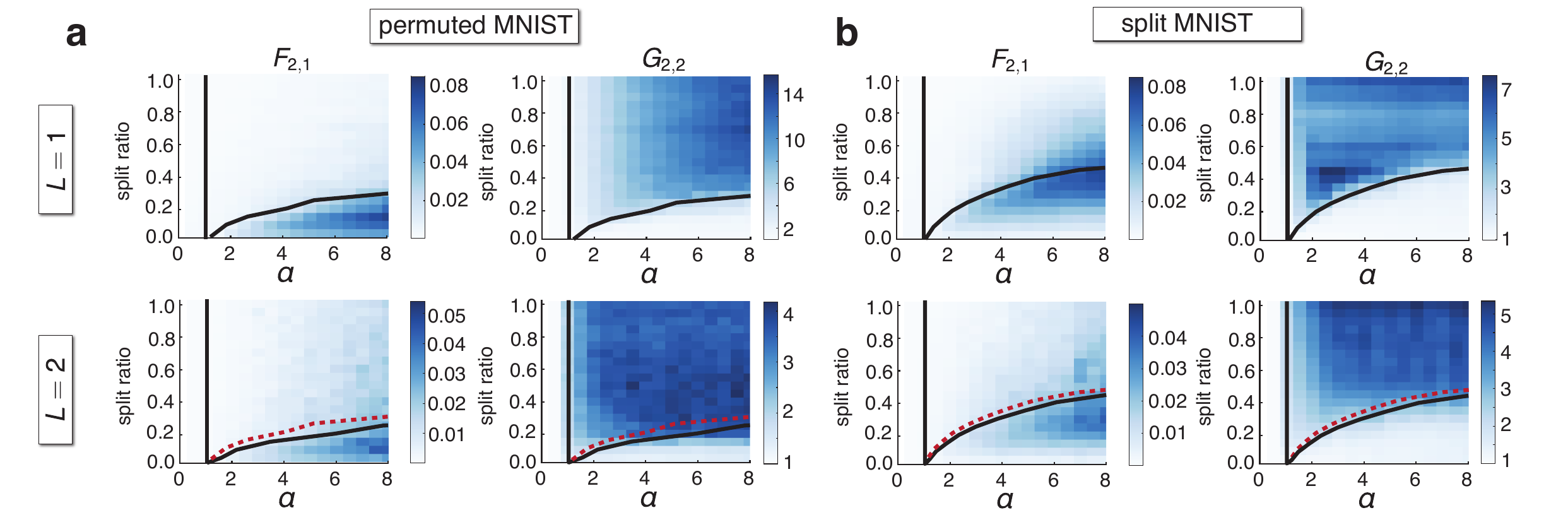}
\caption{\textbf{Phase transitions in networks
trained with gradient descent and explicit L-2 regularizer.}\\
\textbf{a} $F_{2,1}$ (left) and $G_{2,2}$(right) evaluated from networks trained with GD and an explicit L-2 regularizer (SI Eqs. \ref{eq:dynamicsa}, \ref{eq:dynamicsw}), as a function of the permutation ratio and $\alpha$ for permuted MNIST, for $L=1$(top)
and $L=2$(bottom). The same 3 regimes are observed as predicted by
the theory. In networks with $L=1$, theoretical approximation of
the phase-transition boundary (black line) accurately matches the
simulation. In networks with $L=2$, the same phase transitions are observed, and the overfitting regime slightly extends
beyond the theoretical phase-transition boundary of $L=1$ (red dashed line), indicating a slightly stronger anterograde interference effect. Simulations
are done with $\kappa=0.1$, $\sigma_{0}=1$, 
 $\eta=0.01$ for $L=1$
and $\eta=0.001$ for $L=2$ and $P=600$. \\
\textbf{b} Same as a, but on the split MNIST task sequence. Simulations are done
with $\kappa=0.1$, $\sigma_{0}=1$, $\eta=0.005$ for $L=1$ and $\eta=0.001$
for $L=2$ and $P=600$.\\
Results were averaged over 10 random seeds for initialization.}
\label{fig:gdmulti}
\end{figure}

\begin{figure}[H]
\centering 
\includegraphics[width=1\linewidth]{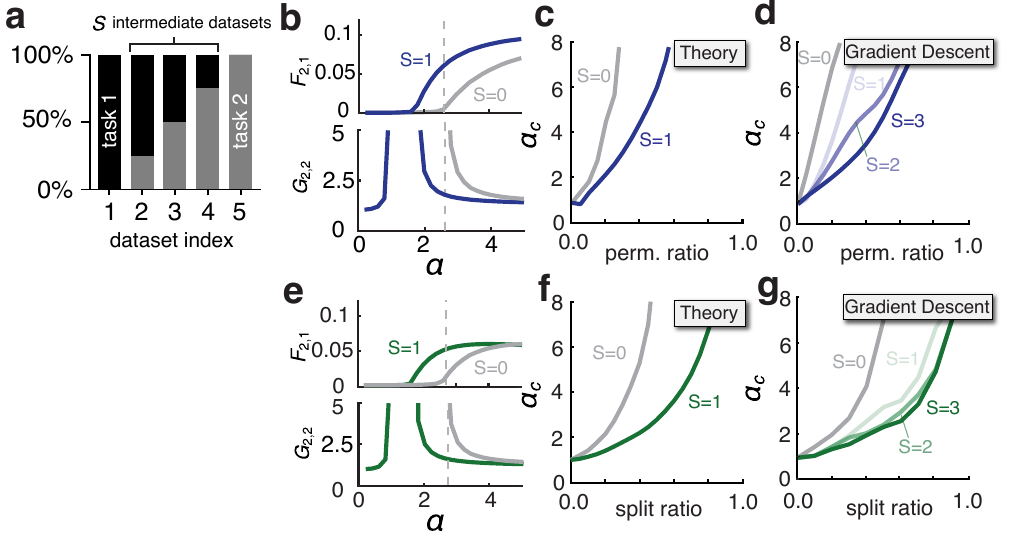}
\caption{\textbf{Intermediate
datasets help mitigate anterograde interference.} \\
\textbf{a} Schematics of a task sequence with intermediate datasets. The first and last datasets solely contain training data from the first and second tasks, respectively, whereas the $S$ intermediate ones contain a mixture of training data from two tasks, with the fraction of examples from task 2 increasing incrementally and equally across these datasets. \\
\textbf{b} Forgetting on the first
task ($F_{2,1}$) and normalized generalization error on the second task ($G_{2,2}$) as a function of the network load $\alpha$ without intermediate datasets ($S=0$, gray) and with one intermediate dataset ($S=1$, blue) for a fixed pair permuted MNIST task sequence with permutation ratio $15\%$. In both cases there exists a critical $\alpha_{c}$, for $\alpha>\alpha_{c}$
the network is in the generalization regime, $F_{2,1}$ is nonzero,
and $G_{2,2}$ is finite; for $1\leq\alpha\leq\alpha_{c}$,
the network is in the overfitting regime, $F_{2,1}=0$ and $G_{2,2}$
diverges. The critical $\alpha_{c}$ becomes smaller with the addition
of an intermediate dataset, indicating a smaller overfitting regime and mitigated anterograde interference. Dashed line: theoretical prediction
of $\alpha_{c}$ for $S=0$. \\
\textbf{c} The critical $\alpha_{c}$ given by the theory, as a function of the permutation ratio in permuted MNIST. For task sequences without any intermediate datasets ($S=0$), $\alpha_{c}$ was calculated by evaluating SI Eq. \ref{eq:simpleptboundary}. For task sequences with 1 intermediate dataset ($S=1$), $\alpha_{c}$ was calculated by estimating the position where $G_{2,2}$ starts to diverge (using $G_{2,2}(\alpha_{c})=\frac{1}{2}(\max_{\alpha>1}G_{2,2}(\alpha)+\min_{\alpha>1}G_{2,2}(\alpha))$). The
difference between the two lines $S=0$ and $S=1$ is larger as the permutation ratio increases, indicating that the benefit of intermediate datasets is more significant when task 1 and task 2 are less similar. \\
\textbf{d} The critical $\alpha_c$ given by gradient descent numerics with an explicit L-2 regularizer (SI Eqs. \ref{eq:dynamicsw}, \ref{eq:dynamicsa}), as a function of the permutation ratio in permuted MNIST, for $S=0-3$. $\alpha_{c}$ was calculated by estimating the position where $G_{2,2}$ starts to diverge in the gradient descent simulations (using $G_{2,2}(\alpha_{c})=\frac{1}{2}(\max_{\alpha>1}G_{2,2}(\alpha)+\min_{\alpha>1}G_{2,2}(\alpha))$ as in \textbf{c}) The effect of mitigating forgetting increases with the number of intermediate datasets $S$. \\
\textbf{e}-\textbf{g} Same as \textbf{b}-\textbf{d}, but for split MNIST. In \textbf{e} the split ratio is $25\%$.\\
Results were computed with $P=600$ and $L=1$. \textbf{d} and \textbf{g} were calculated with $\kappa=0.1$, $\sigma_{0}=1$, $\eta=0.01$, and averaged over 10 random seeds for initialization.}
\label{fig:taskinterpolation}
\end{figure}

\begin{figure}[H]
\centering
\includegraphics[width=0.9\linewidth]{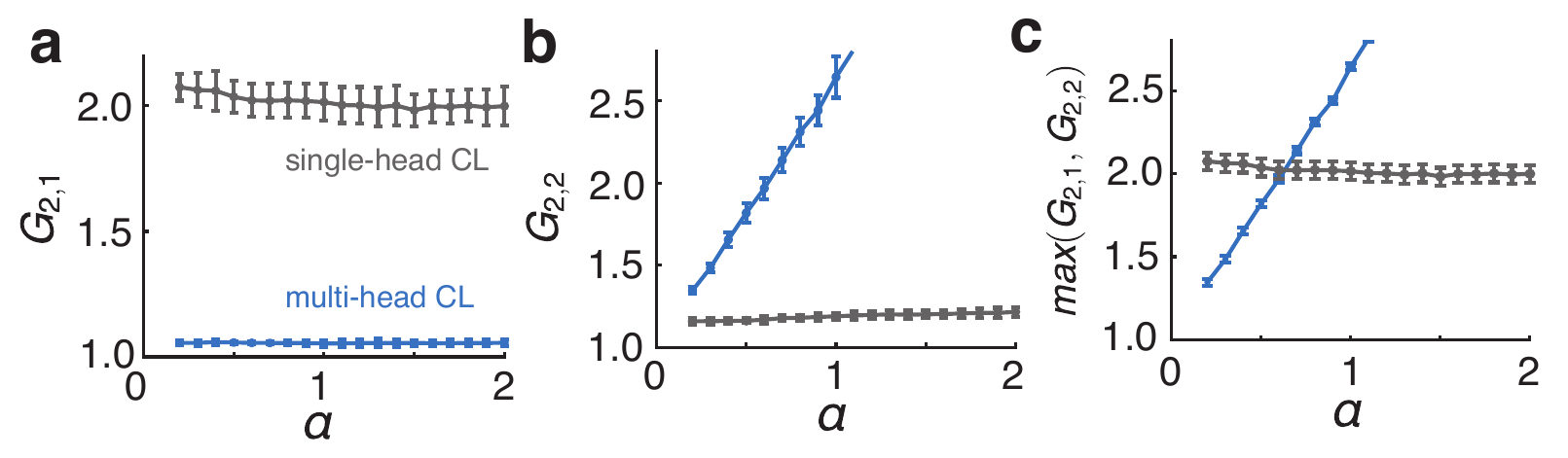}
\caption{\textbf{Empirical comparison between
single-head and multi-Head CL. }\\
\textbf{a} Normalized generalization error on the first task after sequentially learning two tasks for single-head ($G_{2,1}=\left\langle \mathcal{L}(f_{2},D_{1}^{test})\right\rangle/G_1^0 $)
and multi-head ($G_{2,1}=\left\langle \mathcal{L}(f_{2}^{1},D_{1}^{test})\right\rangle/G_2^0$)
CL as a function of the load $\alpha$ ($\alpha$ is changed through modifying $N$ and fixing $P$). Multi-head CL always outperforms single-head CL for the entire range of $\alpha$. \\
\textbf{b} Normalized generalization error on the second task after sequentially
learning two tasks for single ($G_{2,2}=\left\langle \mathcal{L}(f_{2},D_{2}^{test})\right\rangle/G_{2}^0$)
and multi-head ($G_{2,2}=\left\langle \mathcal{L}(f_{2}^{2},D_{2}^{test})\right\rangle/G_2^0$)
CL as a function of $\alpha$. The normalized generalization error remains close to 1 for single-head CL, but increases significantly with $\alpha$ for multi-head CL. \\
\textbf{c} Same as \textbf{a}, \textbf{b}, but taking the maximum over $G_{2,1}$ and
$G_{2,2}$ in order to compare the overall performance on both tasks.\\
Numerics are done using split MNIST dataset with 100\% split ratio, such that multi-head CL lies in the overfitting regime as long as $\alpha>1$. Detailed parameters are $P=600, \gamma=0.1,\eta=0.005$ and $\sigma_{0}=0.5$ for both scenarios.}
\label{fig:singlevsmulti}
\end{figure}

We performed numerical simulations using gradient descent dynamics with an explicit L-2 regularizer to validate the phase transitions predicted by our theory. To capture multi-head CL behaviors in the $\lambda\rightarrow\infty$
limit, we first trained an NN with GD on the first task until the network reaches zero training error. We added an explicit $L-2$ regularizer penalizing the change in hidden-layer weights as we proceed to learn the next tasks. The dynamics is given by 
\begin{align}
\mathcal{W}_{t}^{\tau}&=\mathcal{W}_{t}^{\tau-1}-\eta\nabla_{\mathcal{W}_{t}^{\tau-1}}\mathcal{L}(f_{t}^{t}(\Theta_{t}^{\tau-1}),D_{t})-\eta\kappa(\mathcal{W}_{t}^{\tau-1}-\mathcal{W}_{t-1})\label{eq:dynamicsw}\\
a_{t}^{\tau}&=a_{t}^{\tau-1}-\eta\nabla_{a_{t}^{\tau-1}}\mathcal{L}(f_{t}^{t}(\Theta_{t}^{\tau-1}),D_{t})\label{eq:dynamicsa}
\end{align}
For each task, the readout weights
are always initialized as $a_{t}^{0}\sim\mathcal{N}(0,\sigma_{0}^{2})$,
where as the hidden-layer weights $\mathcal{W}_{t}^{0}$ are initialized
as $\mathcal{W}_{t-1}$, which denotes the weights obtained at the end of training on the previous task. Except for on the first task,
$\mathcal{W}_{1}^{0}$ is initialized as $\mathcal{W}_{1}^{0}\sim\mathcal{N}(0,\sigma_{0}^{2})$. The learning process is run sufficiently long, in order to approximate the $\lambda\rightarrow\infty$ limit where the closest solution to $\mathcal{W}_{t-1}$ is obtained. Results are shown in Figs. \ref{fig:gdmulti}, \ref{fig:taskinterpolation}\textbf{d},\textbf{g}, \ref{fig:singlevsmulti}\textbf{b},\textbf{c}. 

\part{Additional numerical results}
\label{sec:numerics}
\section{Numerical results on single-head CL}
\label{subsec:singleheadnumerics}

\section{Numerical results on multi-head CL} 

\section{Numerical comparison between single-head and multi-head CL}



\end{document}